\documentclass[10pt,twocolumn,letterpaper]{article}

\usepackage{iccv}
\usepackage{times}
\usepackage{epsfig}
\usepackage{graphicx}
\usepackage{amsmath}
\usepackage{amssymb}
\usepackage{xcolor}
\usepackage{caption}
\usepackage{booktabs}

\usepackage[pagebackref=true,breaklinks=true,letterpaper=true,colorlinks,bookmarks=false]{hyperref}

\iccvfinalcopy
\def\iccvPaperID{5326}

\ificcvfinal\fi

\newif\ifdraft
\draftfalse
\ifdraft

\newcommand{\rgc}[1]{{\color{violet}[\textbf{RG:} #1]}}
\newcommand{\dcc}[1]{{\color{orange}[\textbf{DC:} #1]}}

\else
\newcommand{\dcc}[1]{}
\newcommand{\rgc}[1]{}
\newcommand{\opc}[1]{}
\newcommand{\hec}[1]{}
\newcommand{\dgc}[1]{}
\newcommand{\iec}[1]{}

\fi
\begin{document}

\title{Set-the-Scene: Global-Local Training for Generating Controllable NeRF Scenes}

\author{Dana Cohen-Bar
\qquad
Elad Richardson
\qquad
Gal Metzer
\qquad
Raja Giryes
\qquad
Daniel Cohen-Or\\ \\
Tel Aviv University 
}

\twocolumn[{
	\renewcommand\twocolumn[1][]{#1}
	\maketitle
 \vspace{-25pt}
	\begin{center}
        \setlength{\tabcolsep}{2pt}
        \includegraphics[width=1\textwidth] {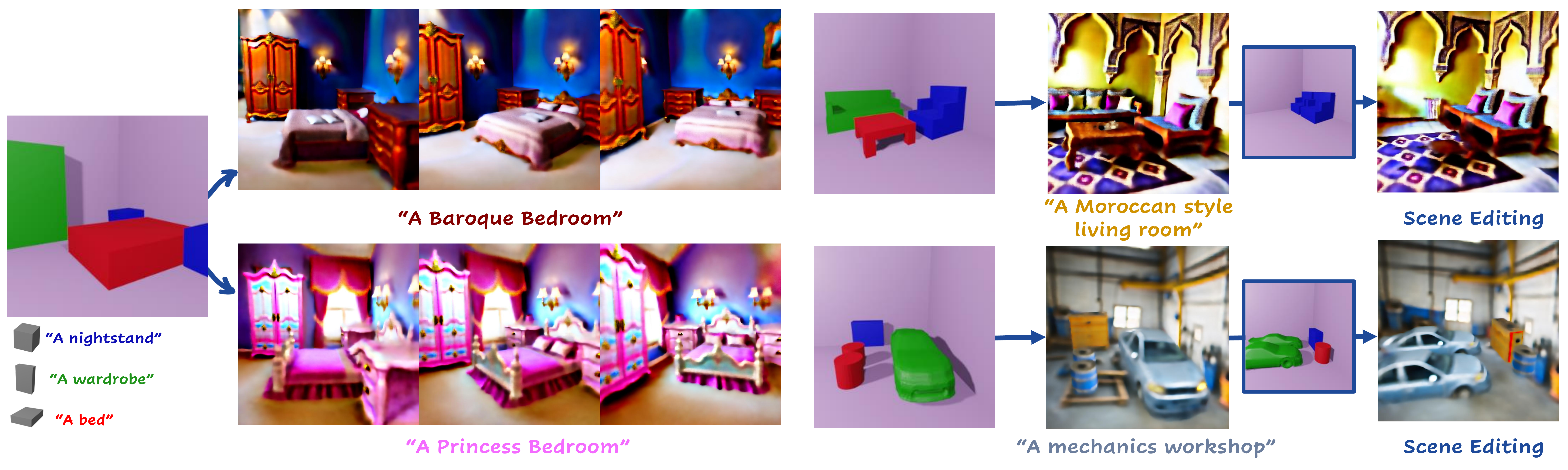} \\
    \captionof{figure}{Set-the-Scene allows generating composable and controllable scenes from text prompts and 3D object proxies. (left) The scene is represented using a set of proxies, defining the location, coarse shape, and text prompt of each target object. A set of NeRFs are then optimized with respect to the object proxies and an additional scene text prompt. (right) By manipulating the proxies, the scene can be edited without additional fine-tuning.}
    \label{fig:teaser}
	\end{center}
}]

\ificcvfinal\thispagestyle{empty}\fi

\begin{abstract}
\vspace{-0.15in}
Recent breakthroughs in text-guided image generation have led to remarkable progress in the field of 3D synthesis from text. By optimizing neural radiance fields (NeRF) directly from text, recent methods are able to produce remarkable results.
Yet, these methods are limited in their control of each object's placement or appearance, as they represent the scene as a whole.
This can be a major issue in scenarios that require refining or manipulating objects in the scene.
To remedy this deficit, we propose a novel Global-Local training framework for synthesizing a 3D scene using object proxies. A proxy represents the object's placement in the generated scene and optionally defines its coarse geometry.
The key to our approach is to represent each object as an independent NeRF. We alternate between optimizing each NeRF on its own and as part of the full scene. Thus, a complete representation of each object can be learned, while also creating a harmonious scene with style and lighting match.
We show that using proxies allows a wide variety of editing options, such as adjusting the placement of each independent object, removing objects from a scene, or refining an object.
Our results show that Set-the-Scene offers a powerful solution for scene synthesis and manipulation, filling a crucial gap in controllable text-to-3D synthesis.
\end{abstract}
\vspace{-0.2in}

\section{Introduction}
Creating high-quality 3D content has traditionally been a time-consuming process, requiring specialized skills and knowledge. However, recent advances in text-to-3D synthesis~\cite{poole2022dreamfusion,lin2022magic3d,metzer2022latent, wang2022score} are revolutionizing the generation of 3D scenes. These methods use pretrained text-to-image diffusion models to optimize a Neural Radiance Field (NeRF) and generate 3D objects that match a given text prompt.

In spite of these exciting advancements, current solutions still lack the capability to create a specific envisioned scene. That is because controlling the generation of text prompts alone is extremely challenging, especially if one wants to describe specific objects with defined dimensions and geometry, and locate them at specific positions. 
Moreover, once generated, modifying specific aspects of the scene, while leaving the others untouched, can prove to be challenging. Various aspects of a scene may need to be modified after generation, such as the position or orientation of certain objects, or the texture or geometry of individual components. 
One may also want to export a single object to be used in another scene. 
However, current methods represent the scene as a whole, and objects are interdependent in their representation, making it impossible to edit a specific scene component or use objects in other scenes.

\begin{figure}[t]
    \centering
    \setlength{\tabcolsep}{0pt}
    {\small
    \begin{tabular}{c c c}
        \includegraphics[width=0.33\linewidth,clip]
        {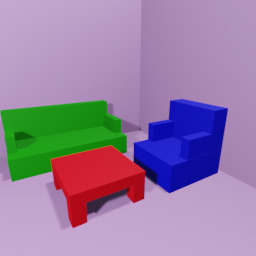} &
        \includegraphics[width=0.33\linewidth,clip]
        {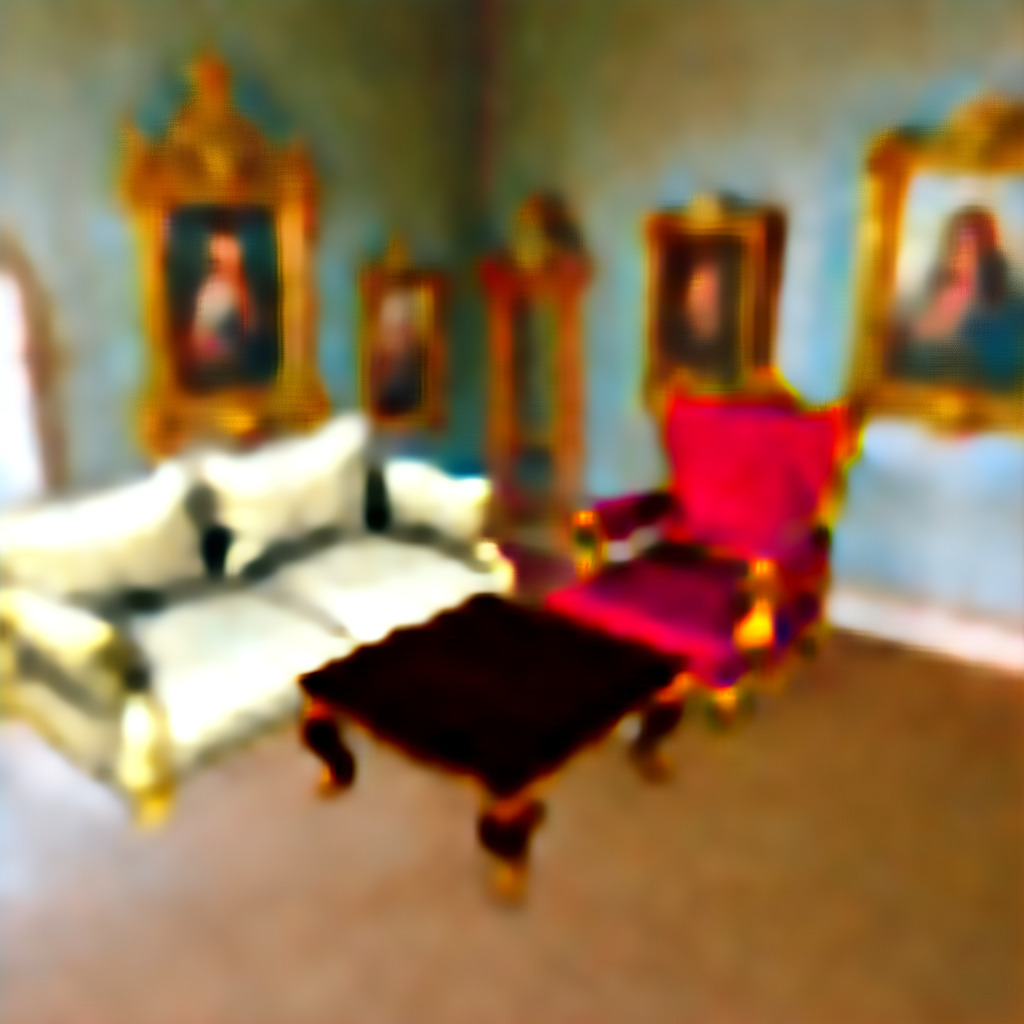} &
        \includegraphics[width=0.33\linewidth,clip]
        {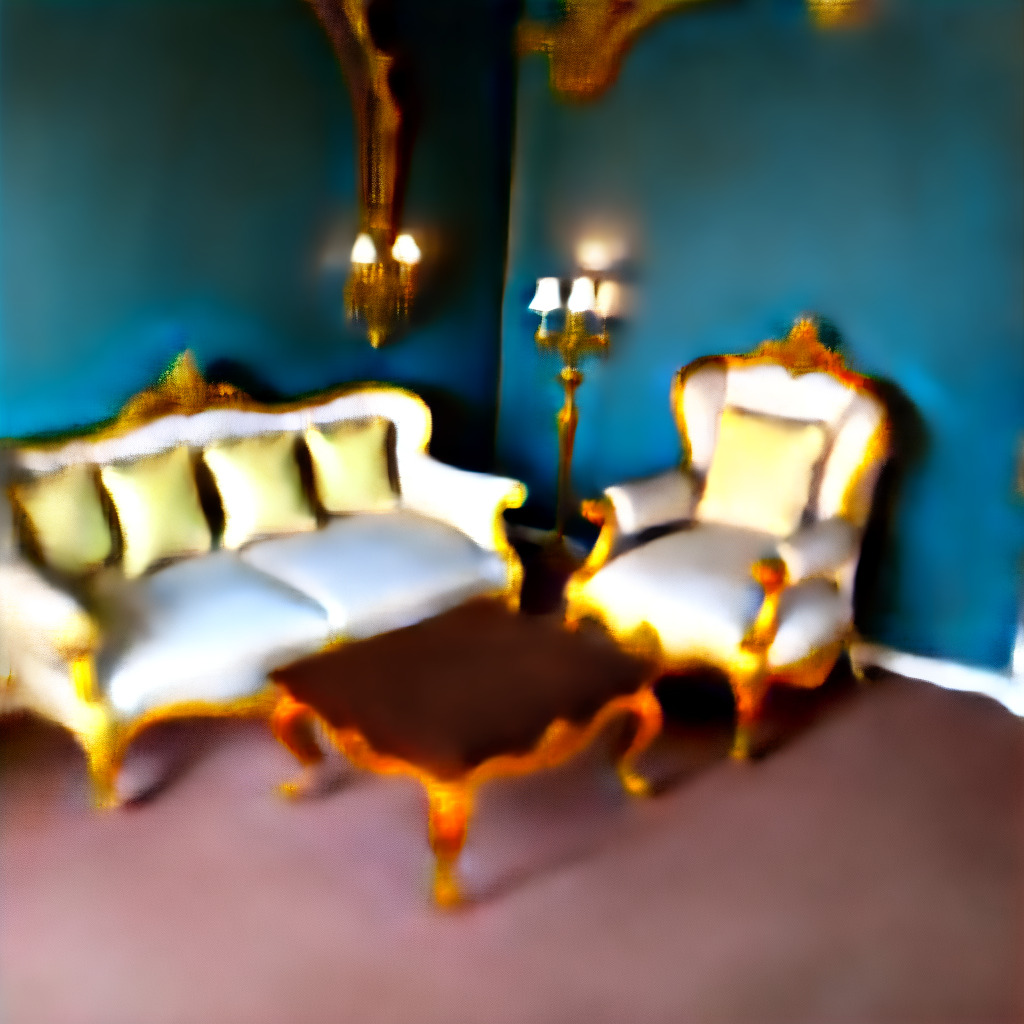} \\
        Input & Local Only & Global-Local 
    \end{tabular}
    }
    \caption{{\bf Importance of Global-Local training.} (Input) The proxy objects used to define the scene. (Local Only) A scene where each object is optimized only for itself; notice how objects look pasted and do not match in terms of color scheme. (Global-Local) Our global-local training, which also interleaves global training steps of the entire scene.}
    \label{fig:ablation_teaser}
\end{figure}

 Given that the generation process of scenes using current text-to-3D methods takes a considerable amount of time, the need for interactive editing capabilities has become even more apparent.
One may potentially save time and gain control over individual objects by generating them separately and then just rendering them together at inference time. However, such an approach has several limitations: it cannot generate objects that interact with each other; it cannot ensure consistency in style between objects; and it cannot model the interaction among objects, like shadows and other global shading effects, see Figure~\ref{fig:ablation_teaser} (A). 

In this paper, we introduce a novel framework for synthesizing a controllable scene using text and object proxies, using a \textbf{Global-Local} approach. The key idea is to represent the scene as a composition of multiple object NeRFs, each built around an object proxy. The models are jointly optimized to ``locally'' represent the required object and ``globally'' to be part of the larger scene. Both the local and the global optimization propagate gradients into the same models, creating a harmonious scene composed of disentangled objects, see Figure~\ref{fig:ablation_teaser} (B).
For optimizing our objects and scenes, we follow~\cite{poole2022dreamfusion} and use the score distillation loss. Our method leverages the composability of our representation and iteratively alternates between localized training of individual objects and optimizing the scene as a whole, where objects are dependent on their representation. When optimizing a single NeRF, we simply render it on its own from a random viewpoint and apply score distillation based on
a text prompt describing the object.  For scene-level optimization, we shift the rays using a rigid transformation to match the desired placement defined by each object proxy and apply score distillation with a ``scene text prompt''.

In many scenarios it is desirable to not only define the placement of an object, but also its dimensions and coarse geometry. Therefore, we also optionally apply a shape loss~\cite{metzer2022latent} on each object proxy to guide it towards a specific shape. 
In addition, we demonstrate that our approach enables the definition of multiple object proxies that can be linked to a single object NeRF. This permits the specification of replicated objects that are intended to be located in multiple positions throughout the scene (e.g., chairs around a table), while aggregating the score distillation from the different placements to optimize a single NeRF

Our proposed Global-Local approach not only provides more control during the training process, but also allows for better editing and fine-tuning of generated scenes. Specifically, using object proxies, we can easily control the placement of an object without the need for further refinement and even remove or duplicate the object as desired. Additionally, we can selectively fine-tune only parts of the scene by defining the set of proxies that are trained and fine-tuning the respective NeRF with modified %
text prompts. Furthermore, we demonstrate that the object proxy can be used to define geometry edits on the coarse shape, which are then applied during the fine-tuning process.

The contributions of our paper are threefold: (i) we propose to represent each object in the scene as a separate NeRF around a proxy, which allows getting a disentangled model for each object, (ii) we introduce a new optimization strategy that interleaves between single-object optimization and scene optimization, resulting in self-contained objects that can be combined to create a plausible scene, and (iii) our strategy provides control over the generated scene, both before and after its creation.

\section{Related Work}

\begin{figure*}
 \begin{center}
        \setlength{\tabcolsep}{2pt}
        \includegraphics[width=1\textwidth] {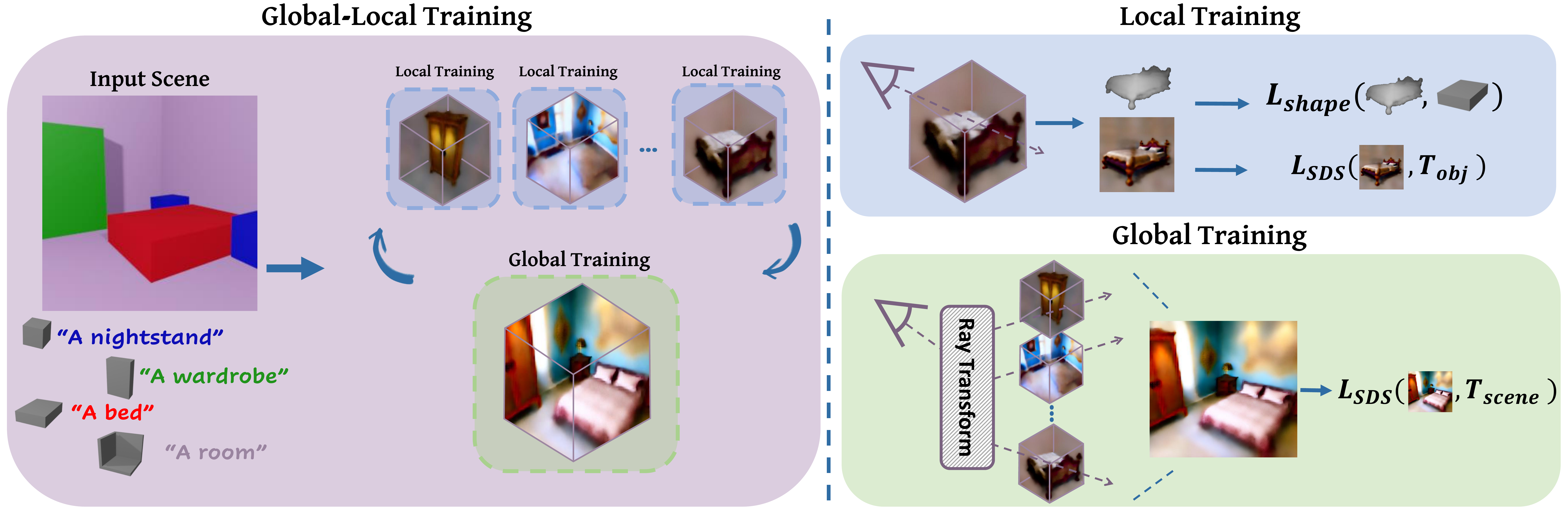} \\
    \vspace{1mm}
    \caption{{\bf Set-the-Scene Training pipeline.}  A scene is first defined using a set of proxies, where each proxy is coupled with a location and a text prompt. Given an input scene, we then apply a global-local training procedure where we alternate between locally optimizing each object on its own and optimizing the entire scene rendered together as a whole.}
    \label{fig:pipeline}
	\end{center}
\end{figure*}        

\paragraph{Text-Driven Shape Generation}
3D shape generation from text has been a well-researched topic for the past couple of years. 
Text2Mesh~\cite{michel2022text2mesh}, ClipMesh~\cite{khalid2022clipmesh} and Tango~\cite{chen2022tango} use Clip~\cite{radford2021learning} to optimize a triangular mesh to match an input text prompt.
ClipForge~\cite{sanghi2021clipforge} trains a point cloud generator conditioned on Clip embeddings, such that novel shapes can be generated solely by a single matching text prompt.
DreamFields~\cite{jain2021dreamfields} uses Clip supervision to guide a 3D NeRF scene to match a target text. 

There has been an extraordinary development in NeRF papers in recent years. 
Originally used for capturing scenes from a collection of images~\cite{mildenhall2021nerf, barron2021mip, mueller2022instant}, NeRFs have recently also been adopted for shape generation~\cite{jain2021dreamfields, poole2022dreamfusion, wang2022score, metzer2022latent, lin2022magic3d, liao2023text}.
DreamFusion~\cite{poole2022dreamfusion} first introduced \textit{Score-Distillation} for generating novel objects conditioned on an input text prompt by leveraging a pretrained 2D diffusion model. 
Latent-NeRF~\cite{metzer2022latent} extended DreamFusion to the latent domain in order to leverage the publicly available Stable-Diffusion~\cite{rombach2021highresolution}. Additionally, Latent-NeRF introduced \textit{Sketch-Shapes} as proxies for specifying a desired approximate target geometry. Yet, their approach is designed to generate only a single object and not multiple ones.

 \paragraph{Scene Generation}
Scene generation has also been tackled before using different approaches. 
Koh \etal \cite{koh2022simple} uses point cloud rendering as a 3D consistent representation and a 2D GAN architecture to synthesize novel realistic and spatially consistent viewpoints of indoor scenes. 
SceneDreamer~\cite{chen2023scenedreamer} is able to generate large-scale 3D landscapes from a collection of images, leveraging a 3D consistent \textit{bird-eye-view} that is trained using a 2D GAN objective.
SceneScape~\cite{fridman2023scenescape} generates novel parts of a scene conditioned on an input text description of the sense and an initial image, by iteratively warping and out-painting the previously generated image according to specified camera movements. 
These methods generate the scene as a whole and do not offer composable control over the generated scenes.

\paragraph{Composable Neural Radiance Fields}
Decomposing neural scenes enables controlling different objects in the scene in a disentangled manner, allowing for explicit editing, removal, and addition of objects.
Compressible-composable NeRF~\cite{tang2022compressible} utilizes a tensor representation of the scene for compression, which also enables rendering multiple objects together as a unified scene.
GIRAFFE~\cite{niemeyer2021giraffe} uses an adversarial loss to train a decomposable generator of feature fields, which when rendered together, form a single scene. The disentangled control over each individual feature field allows controlling the location and appearance of each individual object in the scene.

Object-Centric NeRF~\cite{guo2020object} is able to render multiple NeRF implicit models altogether, where each model was independently optimized to capture a single object.
Similarly, Panoptic Neural Fields~\cite{kundu2022panoptic} also trains separate implicit models for each object, plus a single model for the background. Additionally, each object is associated with a 3D bounding box, which is considered in the merging policy between implicit models at the volumetric rendering stage of the entire scene. 
Nerflets~\cite{Zhang2023Nerflets} propose a set of local NeRFs that represent a scene as a collection of decomposed objects. Each of them maintains its own spatial position, orientation, and extent, allowing for efficient structure-aware 3D scene representation from images. 
Object-NeRF~\cite{yang2021learning} is able to learn individual per-object NeRF models of a cluttered scene, through object-level supervision that is achieved by leveraging a rough segmentation mask of each training image. We adopt the same ray aggregation strategy as Object-NeRF for rendering multiple implicit models in a single image. 

Note that all the above methods are not generative nor text-guided and are limited in their editing capabilities.
Similarly to our global-local paradigm, DisCoScene~\cite{xu2022discoscene} also learns to generate controllable scenes at inference time, 
yet they require a 3D object-level training dataset, while our method does not use any 3D supervision.

\section{Method}
We turn to describe Set-the-Scene. We start by defining object proxies and composable NeRF rendering and then formulate our Global-Local training. Finally, we show how one may easily apply post-training edits using our method.

\subsection{Composable NeRF Rendering}
In order to allow object-level optimization and generation, we represent our scenes as a set of composable NeRFs. 
For explaining our proposed strategy, we first describe general NeRF rendering and the concept of \textit{object proxies} for improved control over the scene.

\vspace{-0.3cm}
\paragraph{NeRF Rendering} A single NeRF~\cite{mildenhall2021nerf} represents a 3D volumetric scene with a 5D function $f_\Theta$ that maps a 3D coordinate $\mathbf{x} = (x,y,z)$ and a 2D viewing direction $\mathbf{d} = (\theta, \phi)$ into a volume density $\sigma$ and an emittance color $\mathbf{c} = (r, g, b)$.
The rendering in Equation~\eqref{eq:rendering} is utilized to render the scene from a defined camera location $o$, by shooting rays from $o$ through each pixel and integrating them for color.
More specifically, given a ray $\mathbf{r}$ originating at $\mathbf{o}$ with direction $\mathbf{d}$, we query $f_\Theta$ at points $\mathbf{x}_i = \mathbf{o} + t_i \mathbf{d}$ that are sequentially sampled along the ray to get densities $\{\sigma_i\}$ and colors $\{\mathbf{c}_i\}$.
Finally, the pixel color is taken to be:
\begin{equation}
    \label{eq:rendering}
    \begin{split}
    \mathbf{\hat C}(\mathbf{r}) = \sum_i  T_i \alpha_i \mathbf{c}_i, 
    T_i = \prod_{j < i} (1 - \alpha_i), \\
    \alpha_i = 1 - \exp(-\sigma_i \delta_i), \delta_i = t_{i+1} - t_i
    \end{split}
\end{equation}
where $\delta_i$ is the step size, $\alpha_i$ is the opacity, and $T_i$ is the transmittance.

\newcommand{\stylesimheight}{0.14\linewidth} 
\begin{figure*}
\centering
    \centering
    \setlength{\tabcolsep}{0pt}
    {\small
    \begin{tabular}{c @{\hspace{0.2cm}} c c @{\hspace{0.1cm}} c c @{\hspace{0.1cm}} c c }
        \includegraphics[height=\stylesimheight]{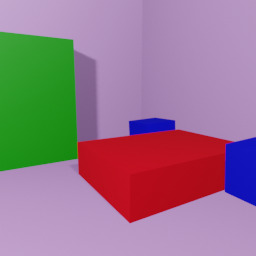} &
        \includegraphics[height=\stylesimheight]{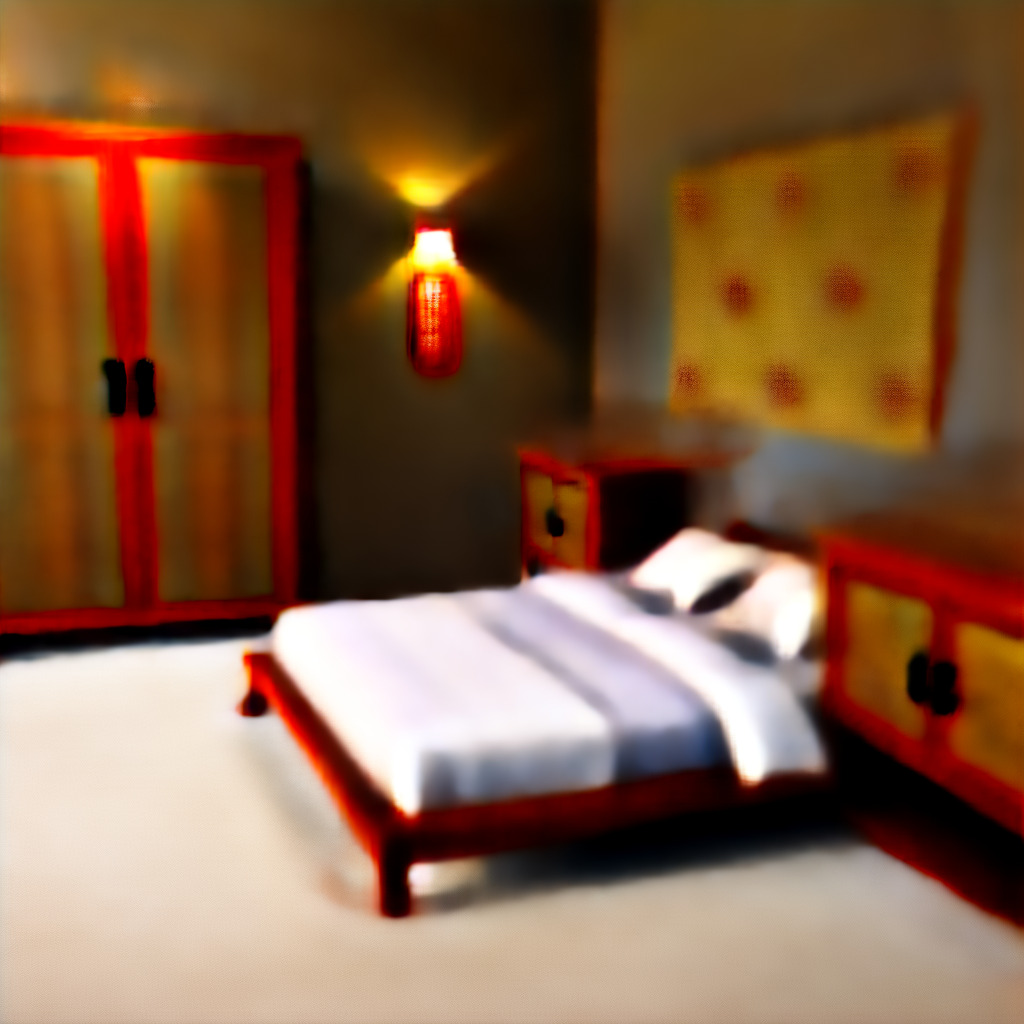} &
        \includegraphics[height=\stylesimheight]{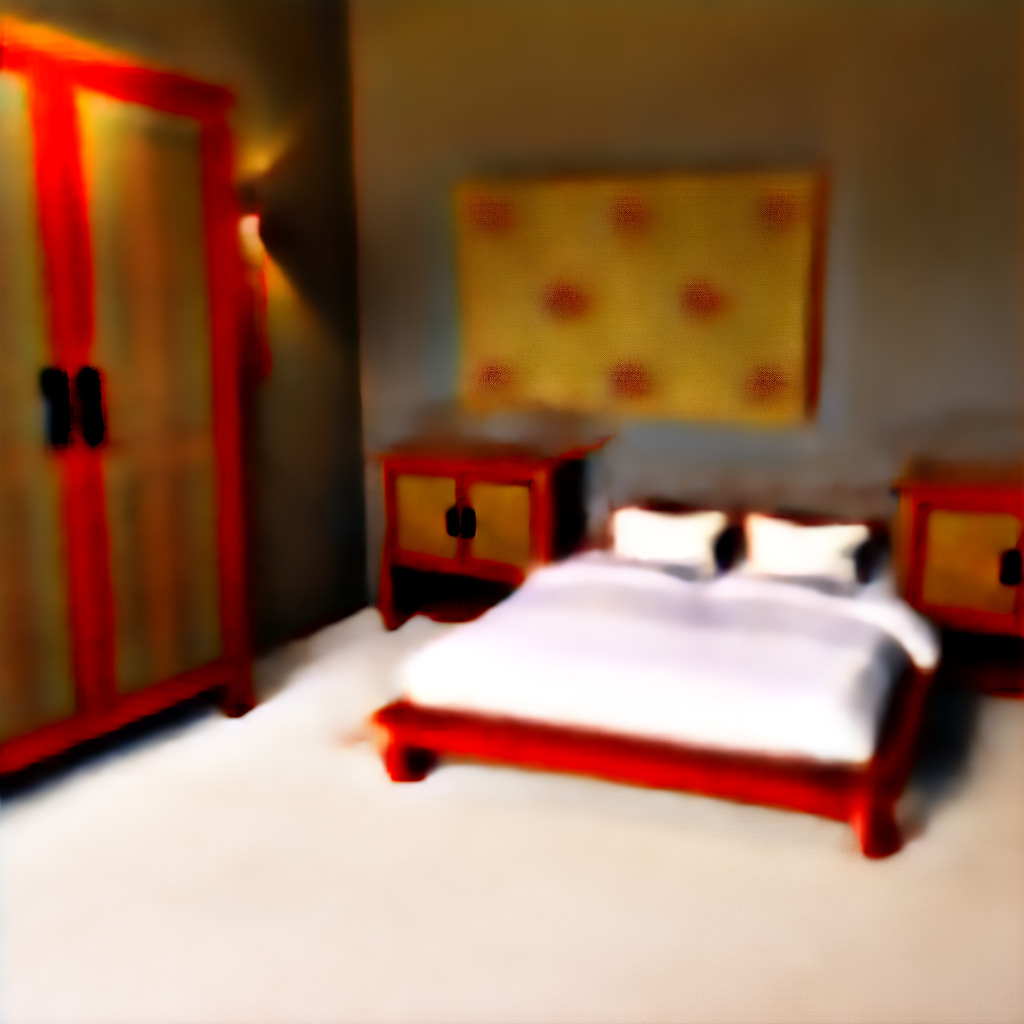} &
        \includegraphics[height=\stylesimheight]{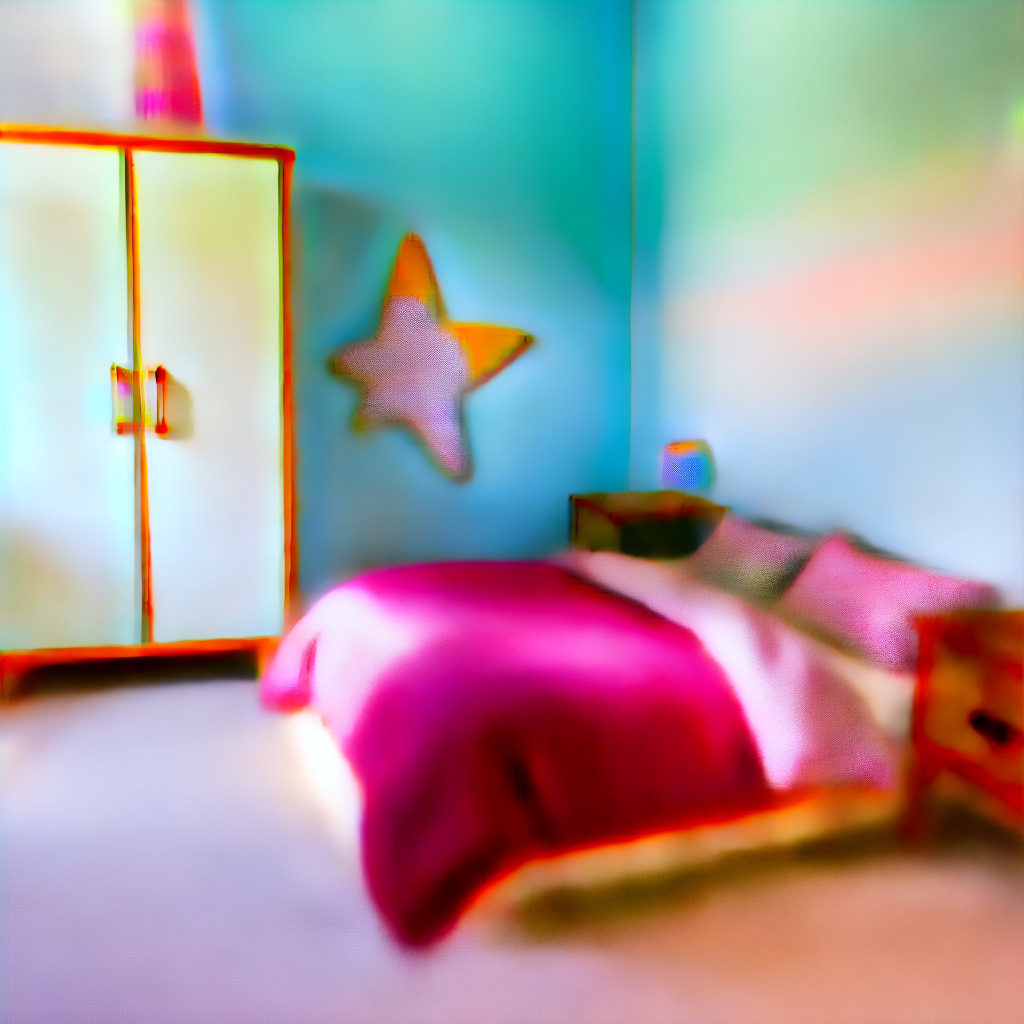} &
        \includegraphics[height=\stylesimheight]{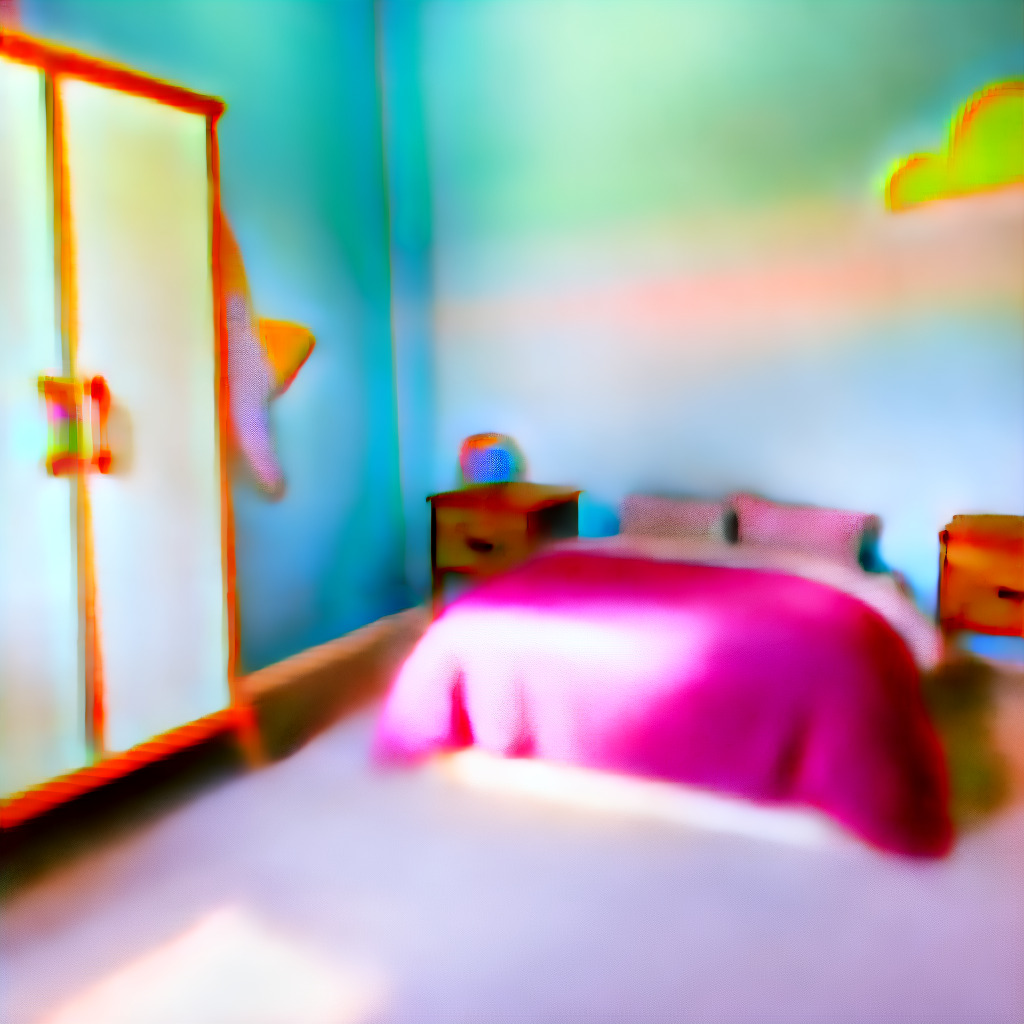} &
        \includegraphics[height=\stylesimheight]{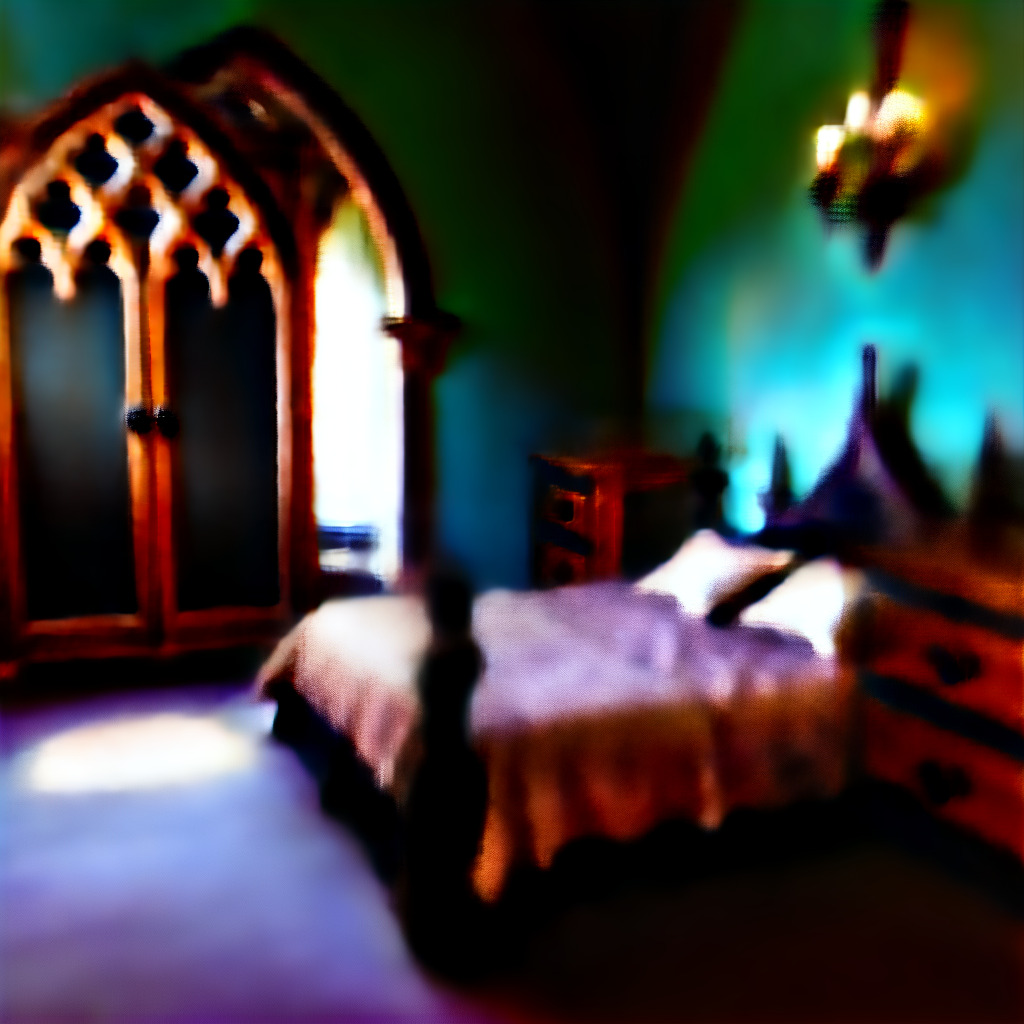} &
        \includegraphics[height=\stylesimheight]{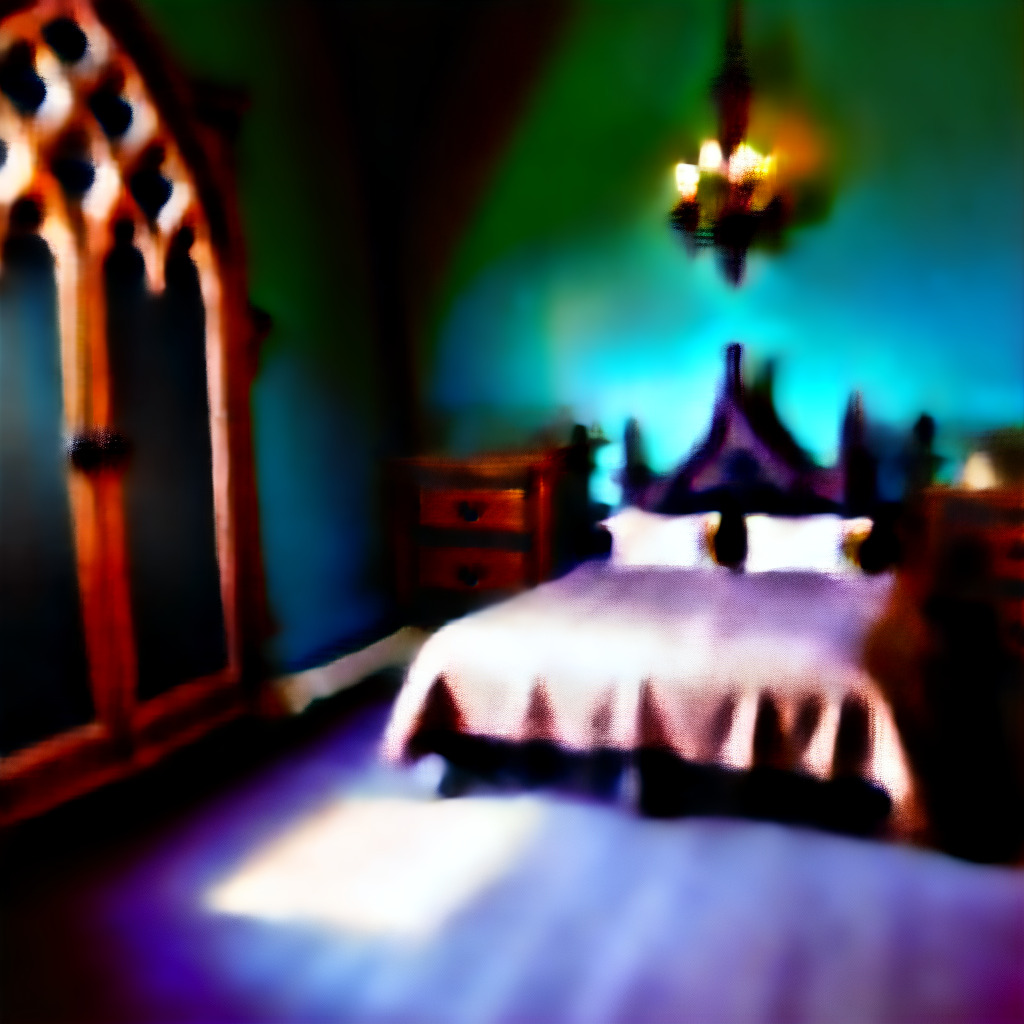}  \\&
          \multicolumn{2}{c}{``An Asian style bedroom''} & \multicolumn{2}{c}{``A kid bedroom''} &
          \multicolumn{2}{c}{``A Gothic style bedroom''}
    \end{tabular}
    }

    \begin{tabular}{c @{\hspace{0.2cm}} c c @{\hspace{0.1cm}} c c @{\hspace{0.1cm}} c c }
    \includegraphics[height=\stylesimheight]{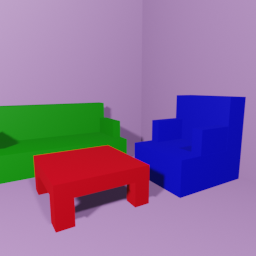} &
    \includegraphics[height=\stylesimheight]{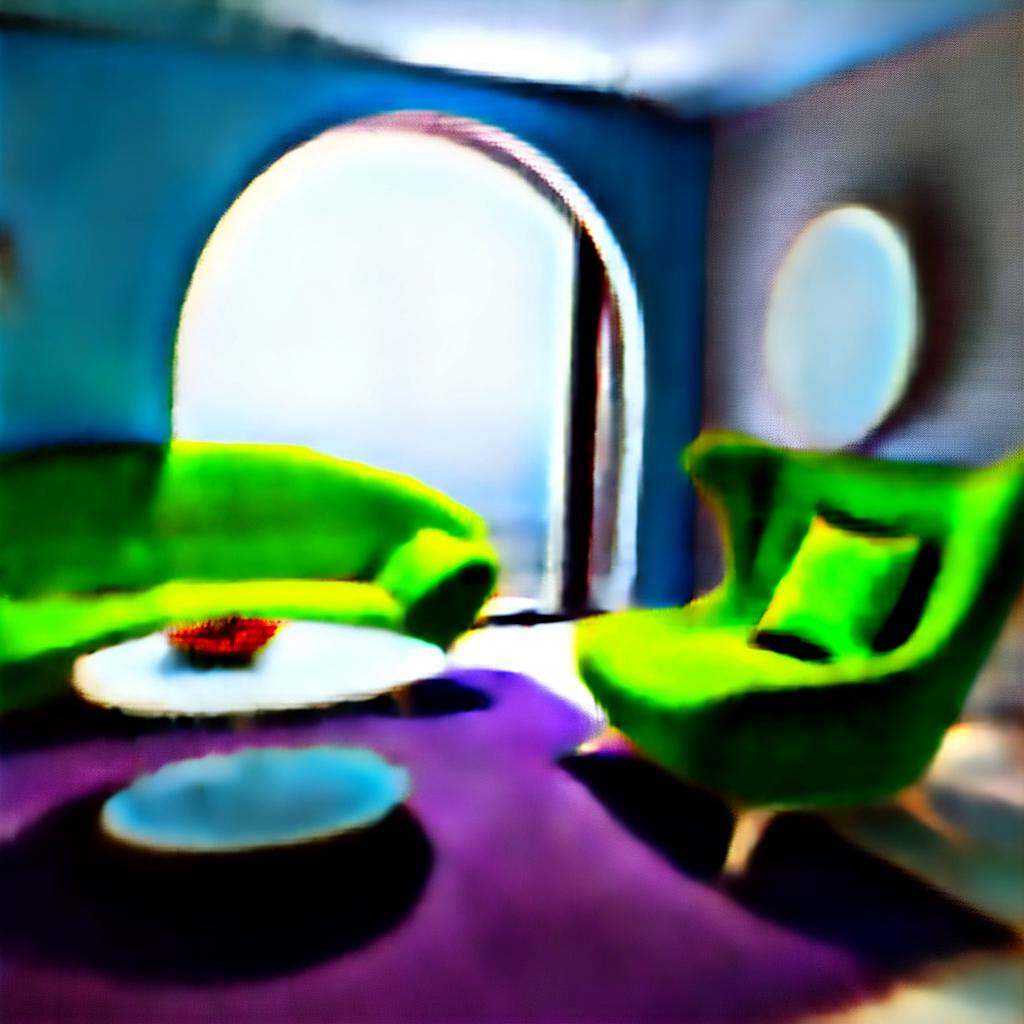} &
    \includegraphics[height=\stylesimheight]{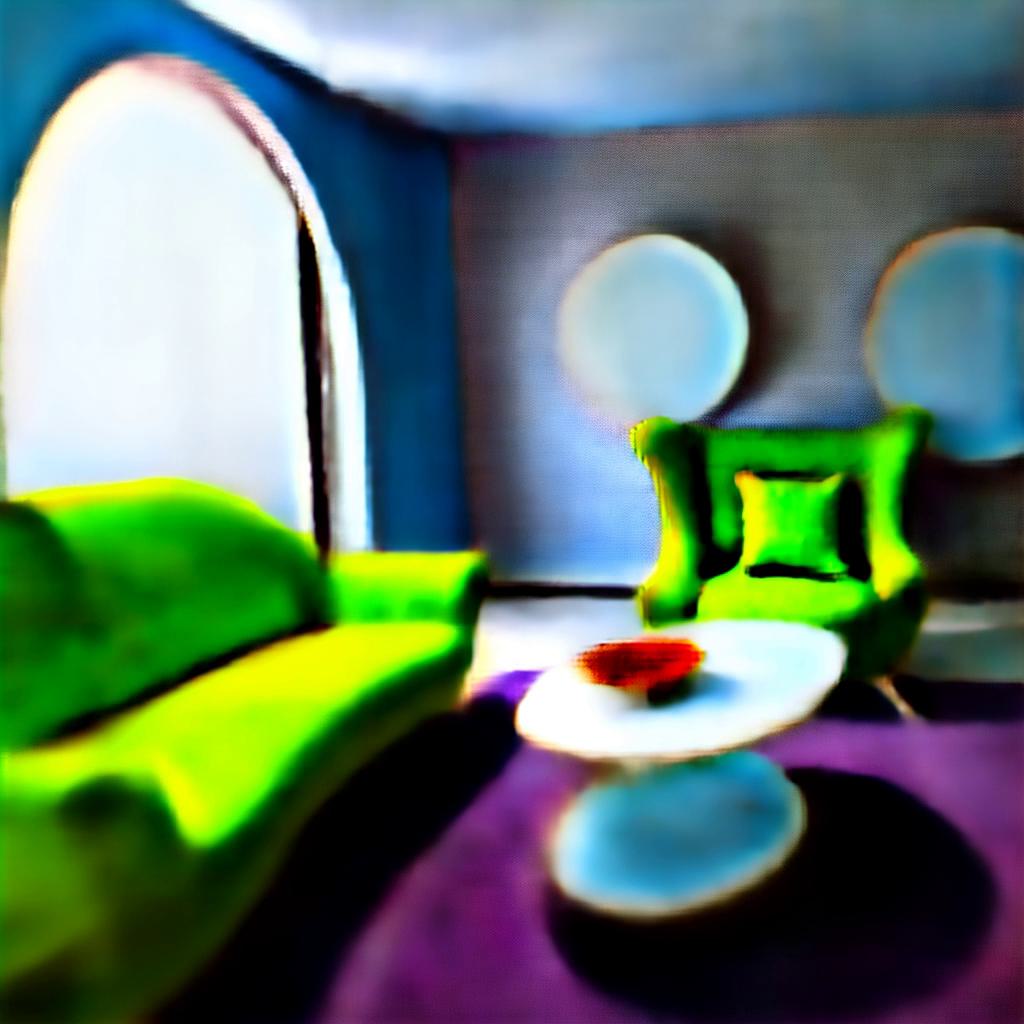} &
    \includegraphics[height=\stylesimheight]{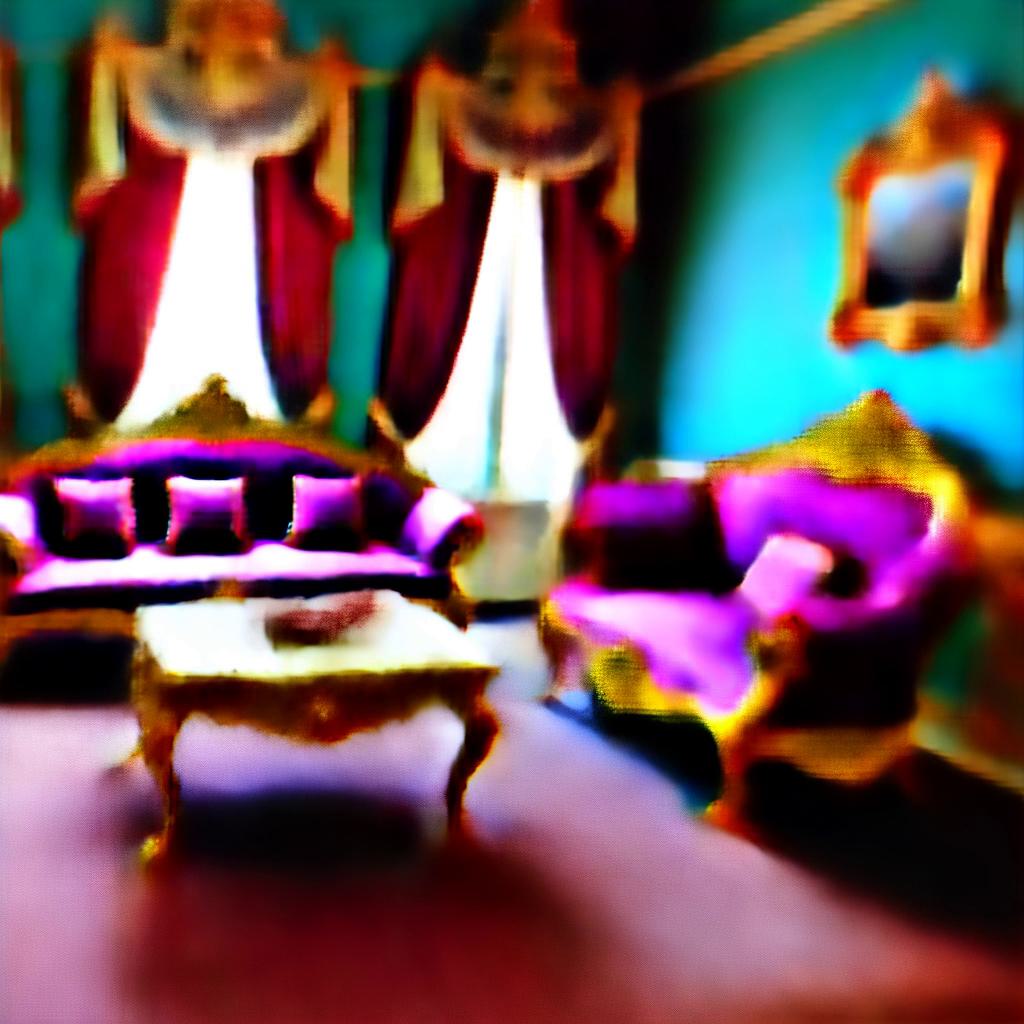} &
    \includegraphics[height=\stylesimheight]
    {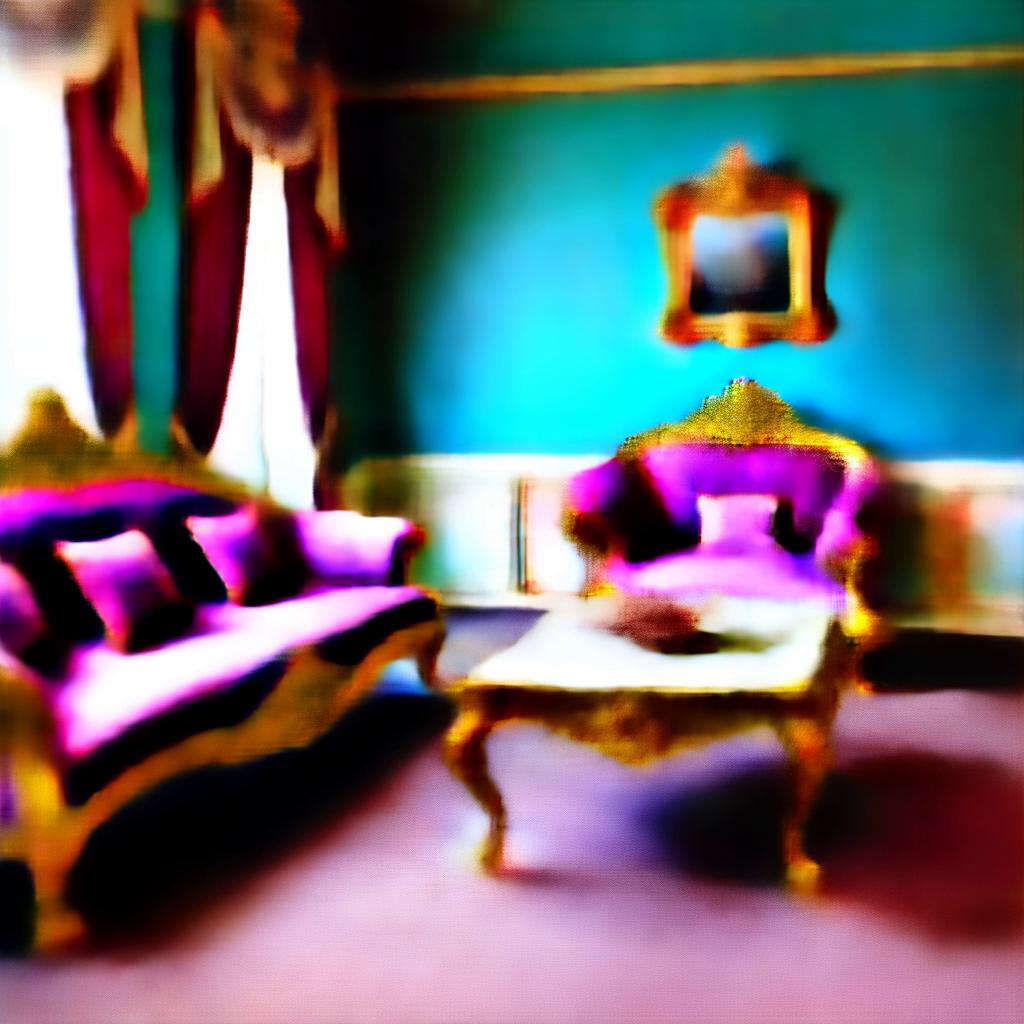} &
    \includegraphics[height=\stylesimheight]{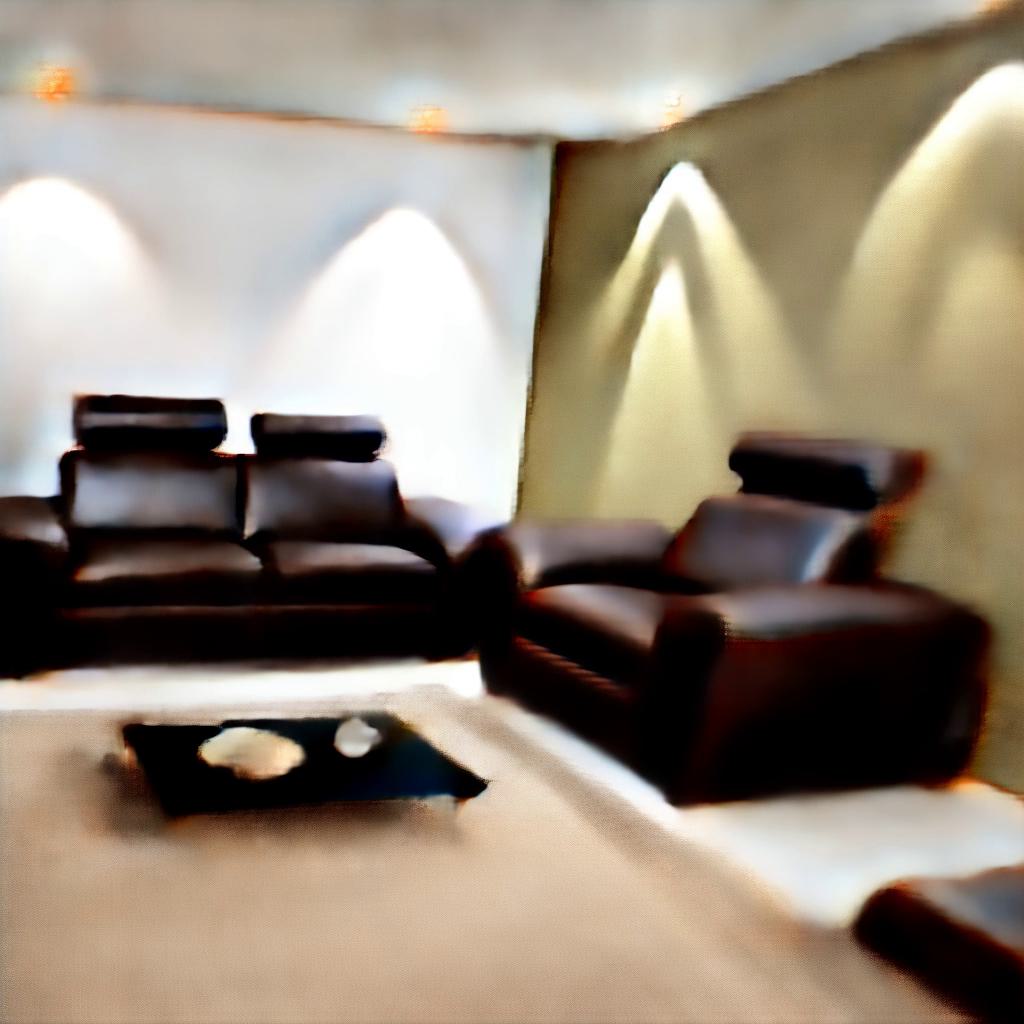} &
    \includegraphics[height=\stylesimheight]
    {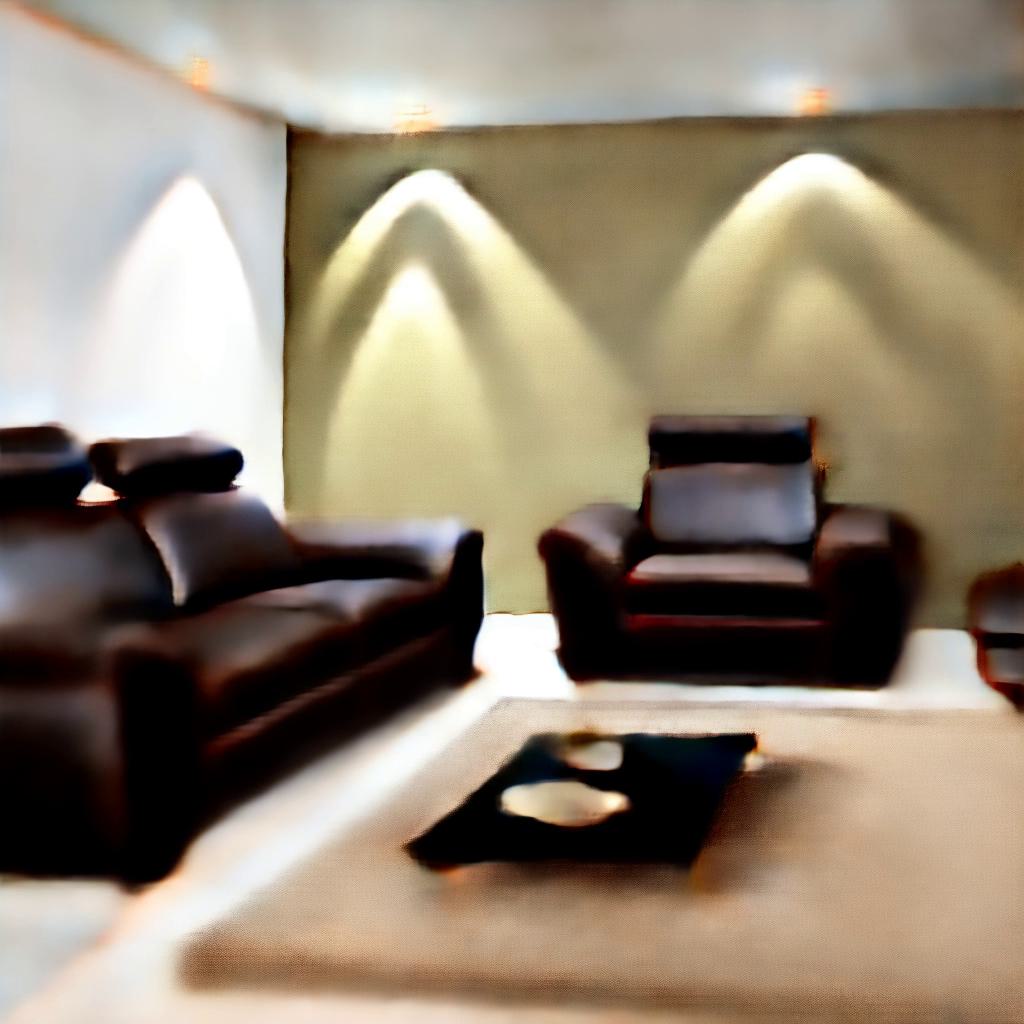} \\&
     \multicolumn{2}{c}{``A futuristic living room''} & \multicolumn{2}{c}{``A Baroque living room''} &
     \multicolumn{2}{c}{``A modern living room''} 
    \end{tabular}
    \vspace{-0.1cm}

    \begin{tabular}{c @{\hspace{0.2cm}} c c @{\hspace{0.1cm}} c c @{\hspace{0.1cm}} c c }
    \includegraphics[height=\stylesimheight]{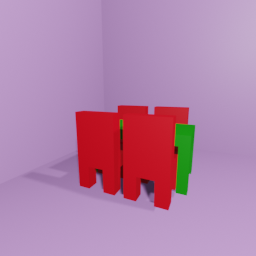} &
    \includegraphics[height=\stylesimheight]{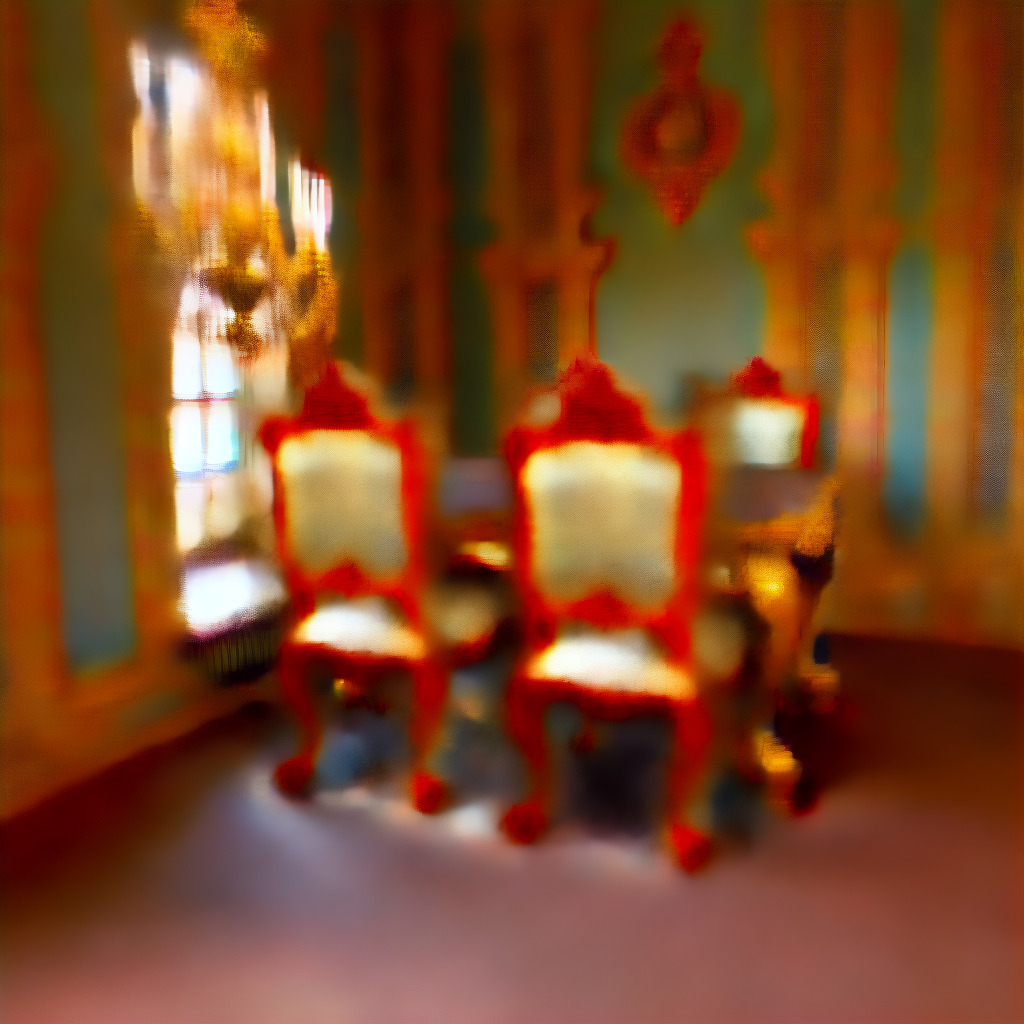} &
    \includegraphics[height=\stylesimheight]{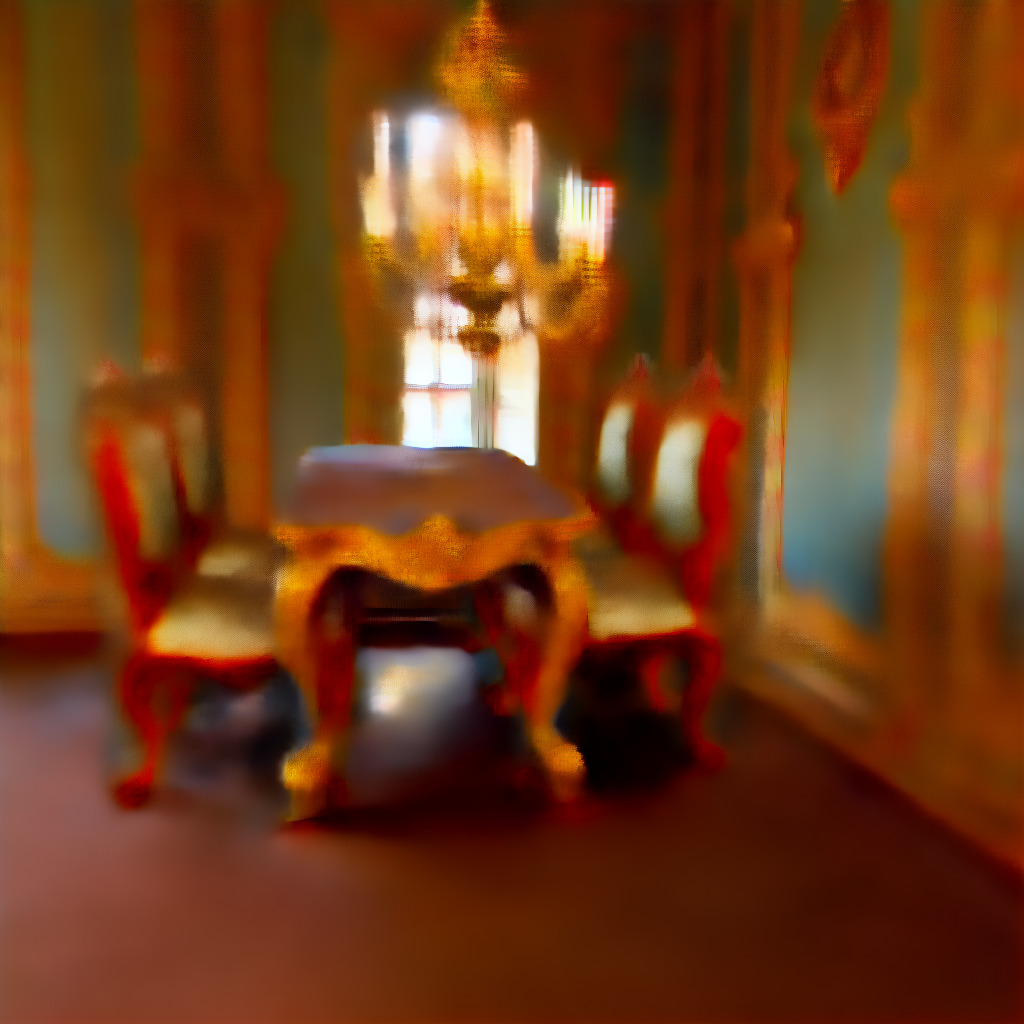} &
    \includegraphics[height=\stylesimheight]{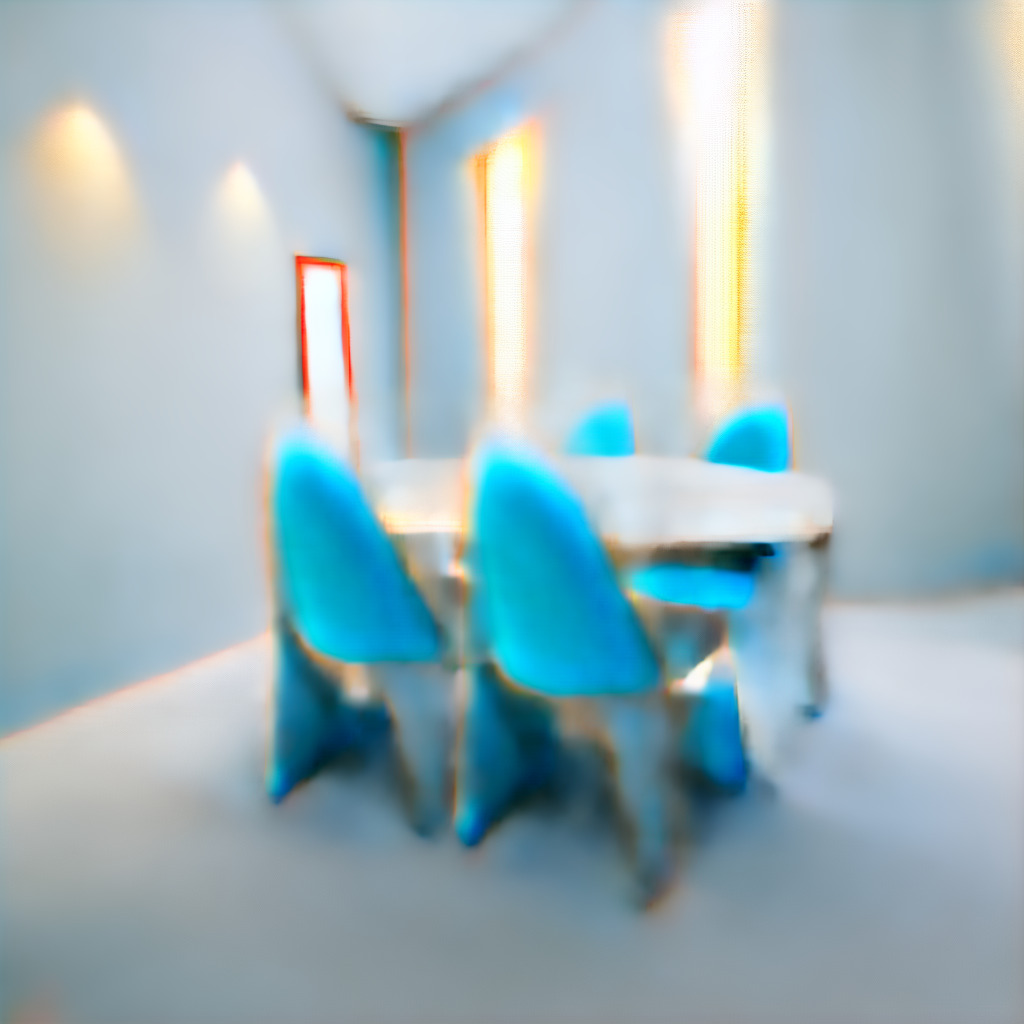} &
    \includegraphics[height=\stylesimheight]
    {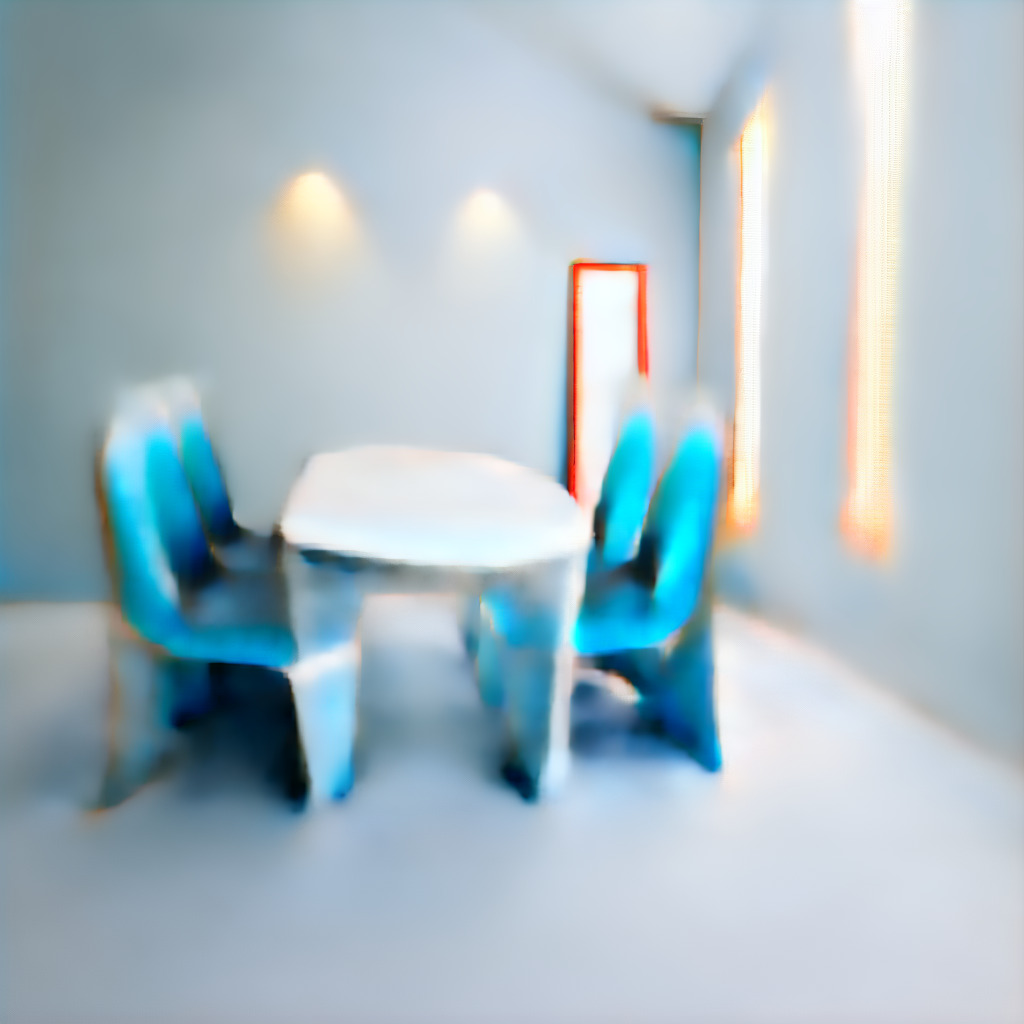} &
    \includegraphics[height=\stylesimheight]{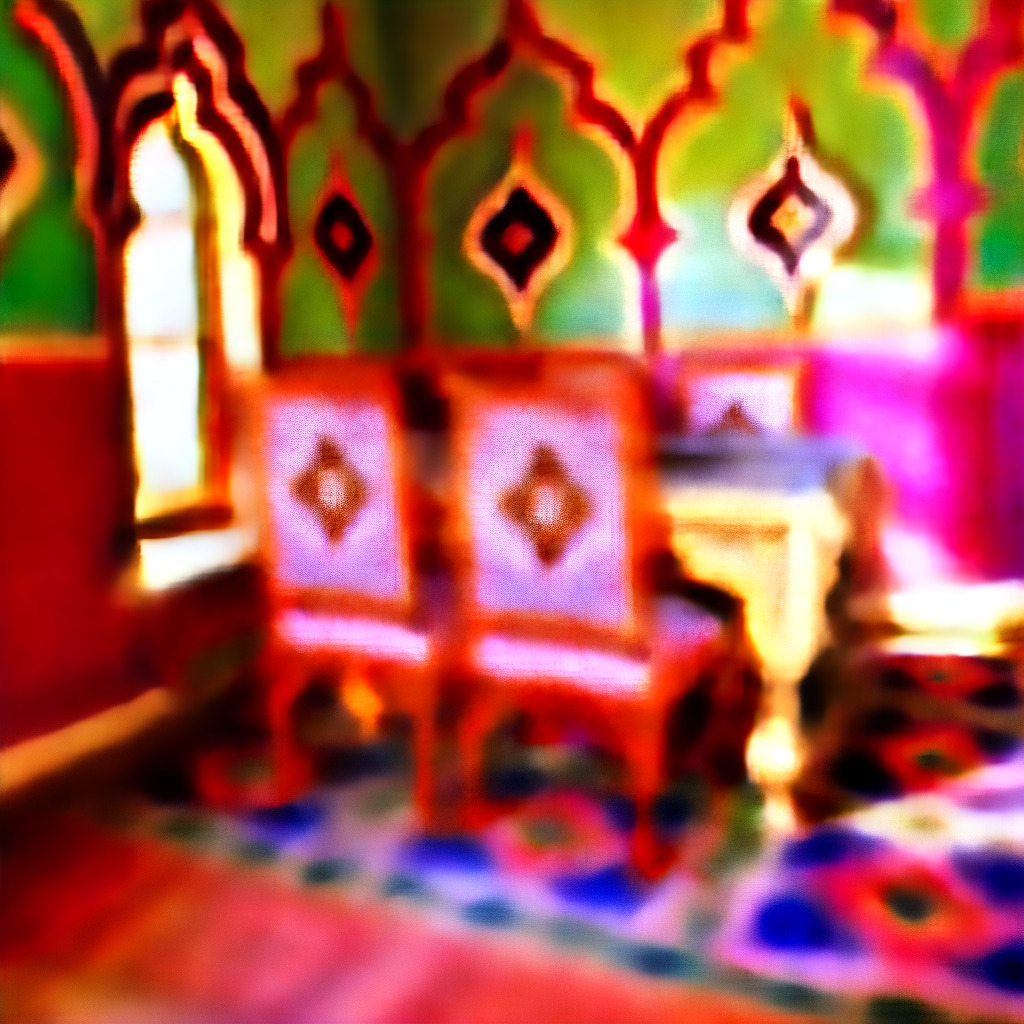} &
    \includegraphics[height=\stylesimheight]
    {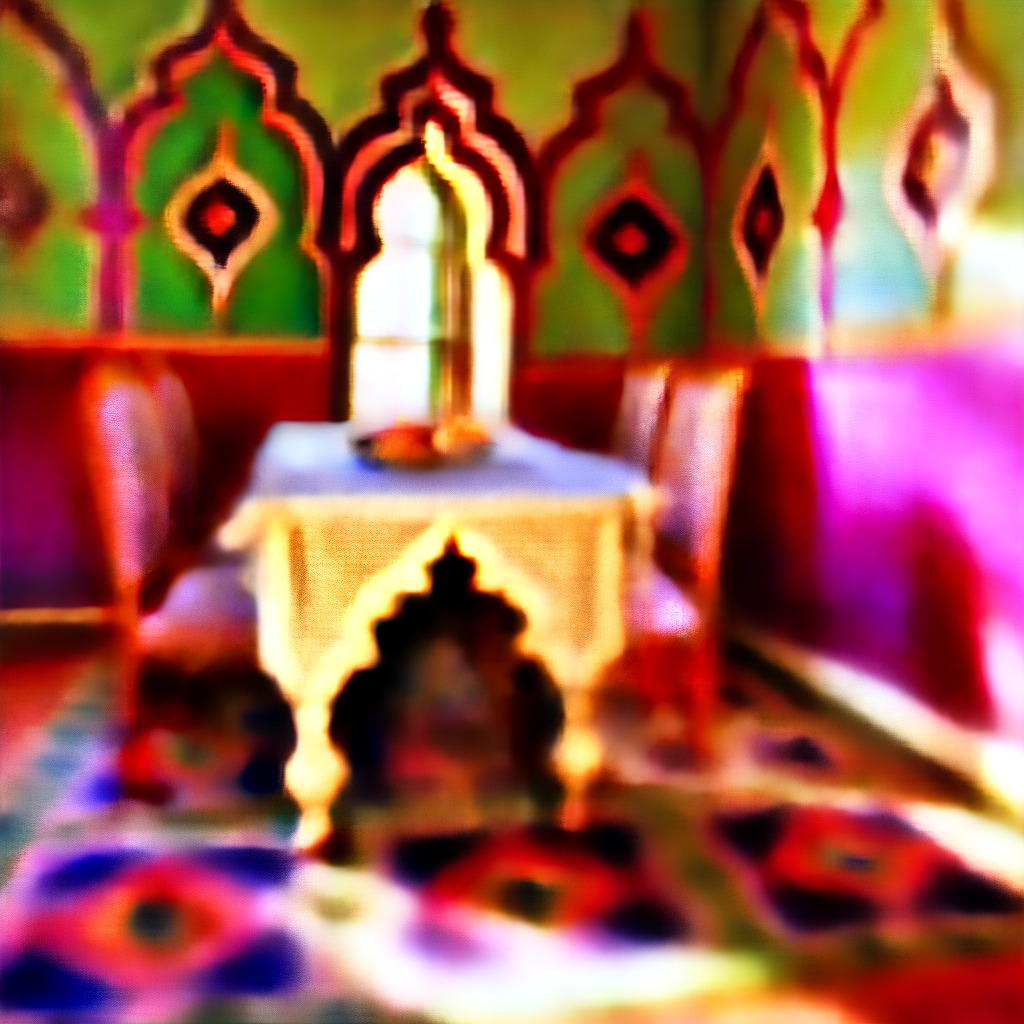} \\&
     \multicolumn{2}{c}{``A baroque dining room''} & \multicolumn{2}{c}{``A futuristic dining room''} &
     \multicolumn{2}{c}{``A Moroccan dining room''} 
    \end{tabular}
    \vspace{-0.1cm}
    \caption{{\bf Set-the-Scene results.} The same proxy setting can be used to create different styles of the same scene. The scene prompts are shown for each generated scene, and corresponding prompts are used for each object. For example ``a kid bedroom style wardrobe, closed doors'' or ``a baroque chair''.} 
    \vspace{-0.3cm}
    \label{fig:different styles}
\end{figure*}

 \paragraph{Object Proxies} 
When rendering multiple NeRFs in the same scene, one first has to define their respective locations in the scene.
We do so through \textit{object proxies}, where each proxy is associated with a NeRF model, offset position, orientation, and size, which define its settings in the scene.
We designed our method so that a single NeRF can be associated with multiple proxies. This allows rendering scenes with numerous instances of the same object, which in turn can be optimized together, as discussed in Section~\ref{subsec:training}.
\paragraph{Multi-Proxy NeRF Rendering}
When rendering multiple objects together, we divide the sampled points from each ray between the different objects, such that each object gets a fraction of the points. 
In order to integrate a single ray over different proxies, each set of points has to undergo a rigid transformation from the scene coordinate system to the proxy coordinate system, such that the computed colors and opacity values are calculated at the object level.
This allows sharing information between different proxies associated with a single NeRF model, and between the Local-Global training phases described in Section~\ref{subsec:training}.

We generalize the treatment in Equation~\ref{eq:rendering} by  summing over different NeRFs and transforming their input points according to their proxy parameters, i.e.,
\begin{equation}
    \label{eq:rendering2}
    \begin{split}
    \mathbf{\hat C}(\mathbf{r}) &= \sum_i  T_i \alpha_i^{(k)} \mathbf{c}^{(k)}_i, \\
    x_i^{(k)} & = R^{(k)}(x_i + {\rm loc}^{(k)}),
    \end{split}
\end{equation}
where $\alpha_i^{(k)}$  is the output of the $k$th NeRF, $R^{k}$  is a rigid matrix defined by the proxy's scale and orientation, and ${\rm loc^{k}}$ is a 3D coordinate defined by the proxy's location. The index $k$ is selected to be $k = i~{\rm mod}~N_{\rm obj}$, where $N_{\rm obj}$ is the total number of objects in the scene, i.e., we sample alternatively between each of the objects equitably.

\subsection{Global-Local Training}\label{subsec:training}
Given our composing mechanism, we now turn to describe our training losses and supervision technique.
\vspace{-0.3cm}
\paragraph{Score Distillation}
To optimize a NeRF using a text-prompt we follow the score distillation loss proposed in~\cite{poole2022dreamfusion}.
Score distillation turns a pretrained diffusion model $\mathcal{M}$ to a critique that is able to provide per-pixel gradients that measure the similarity of a given image to a target text.

\begin{equation}
    \nabla_x L_{sds} \sim \epsilon_{\theta} (x, \mathcal{T}) - \epsilon,
\end{equation}
where $\epsilon$ is a random noise purposely added to the image $x$, $\epsilon_{\theta}$ is the predicted noise by $\mathcal{M}$, and $\mathcal{T}$ is the target text prompt. This allows us to optimize a NeRF to gradually match a given text prompt. In practice, we apply the score-distillation directly to the latent representation and only later decode the results as proposed in~\cite{metzer2022latent}.
\vspace{-0.3cm}
\paragraph{Interleaved Training}

Our method involves an iterative process where we alternate between optimizing each object individually and optimizing the entire scene as a whole. This allows us to take advantage of our composable representation and create objects that harmonize well together but can still be rendered independently.
The object-level iterations are especially important for optimizing occluded areas, which might not be visible at all on the scene level.

In practice, for object-level iterations we choose one of the scene objects and render it in its canonical coordinate system, $\mathcal{O}_i$, such that it is located at the origin and we optimize it with a user-provided text prompt that describes it (\textit{e.g.} ``a wardrobe''). For scene-level iterations, we render all the objects together based on their proxies in the scene coordinate system $\mathcal{S}$ and use a text prompt describing the scene as a whole (\textit{e.g.} ``a baroque bedroom'').

\vspace{-0.3cm}
\paragraph{Defining Proxy Geometry}
A proxy object is always used to define a NeRF placement in the scene. For even higher levels of control, we 
adopt the shape loss from~\cite{metzer2022latent} and allow user-defined shape proxies for each individual object in the scene.
The shape constraints allow users to define proxy geometries in the form of 3D-like sketches and control the dimensions and structure of the generated object, resulting in a much more controllable process.

In practice, the geometry proxy constraint is imposed through an auxiliary loss function, applied both on the scene scale $\mathcal{S}$ and on each individual object scale $\mathcal{O}_i$ alongside the score distillation loss function $L_{sds}$:
\begin{equation*}
    L_{shape} = CE(\alpha_{NeRF}(p), \alpha_{GT}(p)) \cdot (1- e^{-\frac{d^2}{2\sigma_S}}),  
\end{equation*}
where $\alpha_{NeRF}$ is the NeRF's occupancy, $\alpha_{GT}$ is the occupancy of the specified proxy, $d$ is the distance to the proxy's surface, and $\sigma_S$ is a hyperparameter that controls the leniency of the constraint.

\subsection{Post-Training Editing}
Given a generated scene, one might wish to modify some aspects of it. We propose several different tools for refining and editing a generated scene. 

\vspace{-0.4cm}
\paragraph{Placement Editing} The composable formulation of Set-the-Scene inherently allows editing objects' placement in the scene. This is done by changing the proxy location and updating the rays accordingly during rendering. The same technique can also easily duplicate or remove objects.

\vspace{-0.4cm}
\paragraph{Shape Editing} To modify an object's geometry, we simply edit the proxy's geometry and then fine-tune the scene for more iterations. This allows defining shape edits without having to extract a mesh from the implicit NeRF representation. Only the weights of the relevant NeRF are updated in the fine-tuning, and we alternate between rendering it in its canonical coordinate system $\mathcal{O}_i$ and with the rest of the scene in $\mathcal{S}_i$. Scene-level iterations are key to ensuring that the object remains consistent with the scene when edited.

\vspace{-0.4cm}
\paragraph{Color Editing} Finally, we show that one may also edit the color scheme of a generated object. This is done by using an architecture where the density and albedo predictions are separable, and the albedo is predicted using a set of additional fully-connected layers. During fine-tuning we can then optimize the albedo layers to guarantee that the generated shape will not change while modifying its color.

\section{Experiments}
We now turn to a set of experiments that validate and highlight the generation capabilities of Set-the-Scene.

\vspace{-0.3cm}
\paragraph{Implementation Details}
Our method uses the Stable-Diffusion 2.0 model~\cite{rombach2021highresolution} implemented in \textit{Diffusers}~\cite{von-platen-etal-2022-diffusers}. For score-distillation, we use~\cite{metzer2022latent}. During training we iteratively pass over the objects one after the other and apply 10 training steps for each object, global iterations are interleaved between the objects. We train for about 15K iterations, with about 5K of them being global iterations.

\begin{figure}
    \centering
    \setlength{\tabcolsep}{0pt}
    {\small
    \begin{tabular}{c c c c}
        \includegraphics[width=0.24\linewidth]{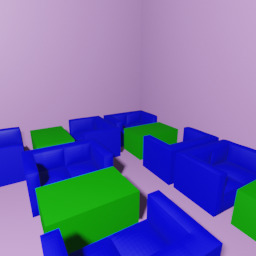} & 
        \includegraphics[width=0.24\linewidth]{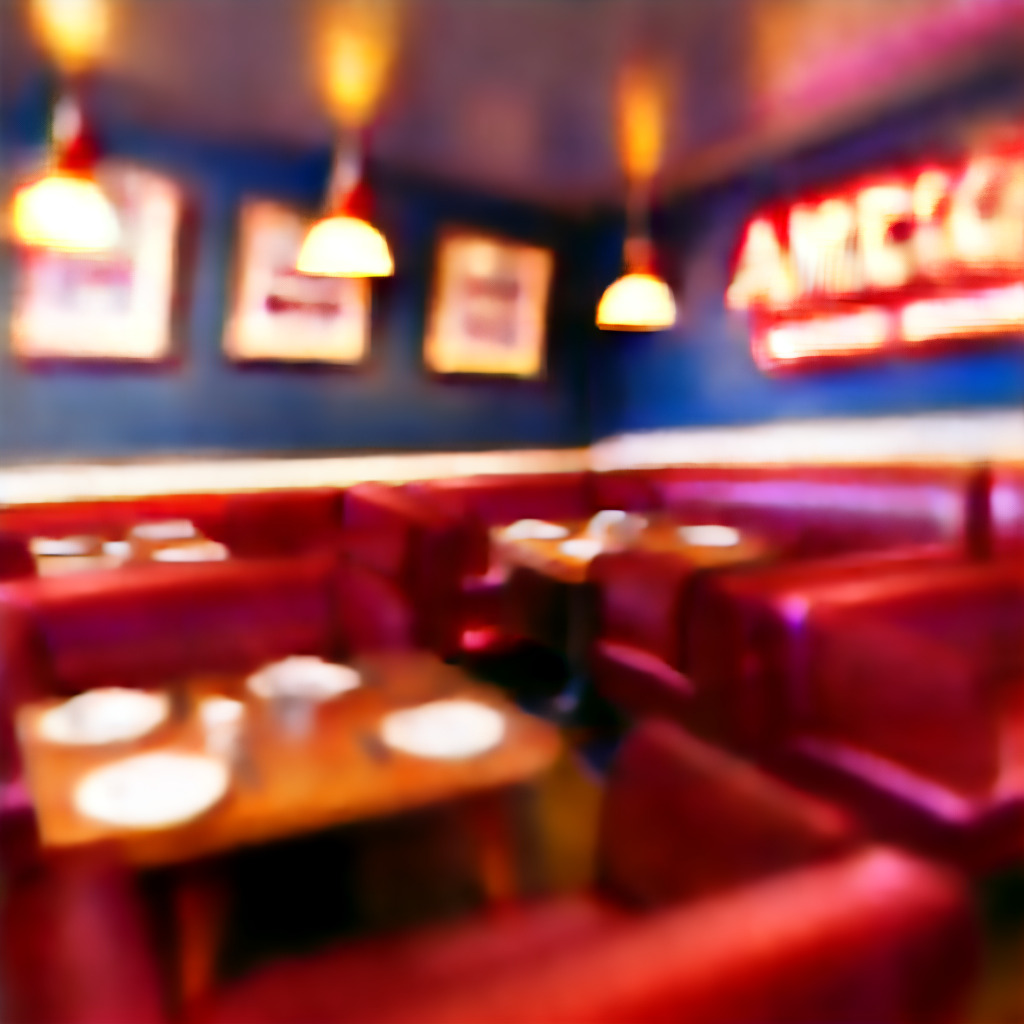} & 
        \includegraphics[width=0.24\linewidth]{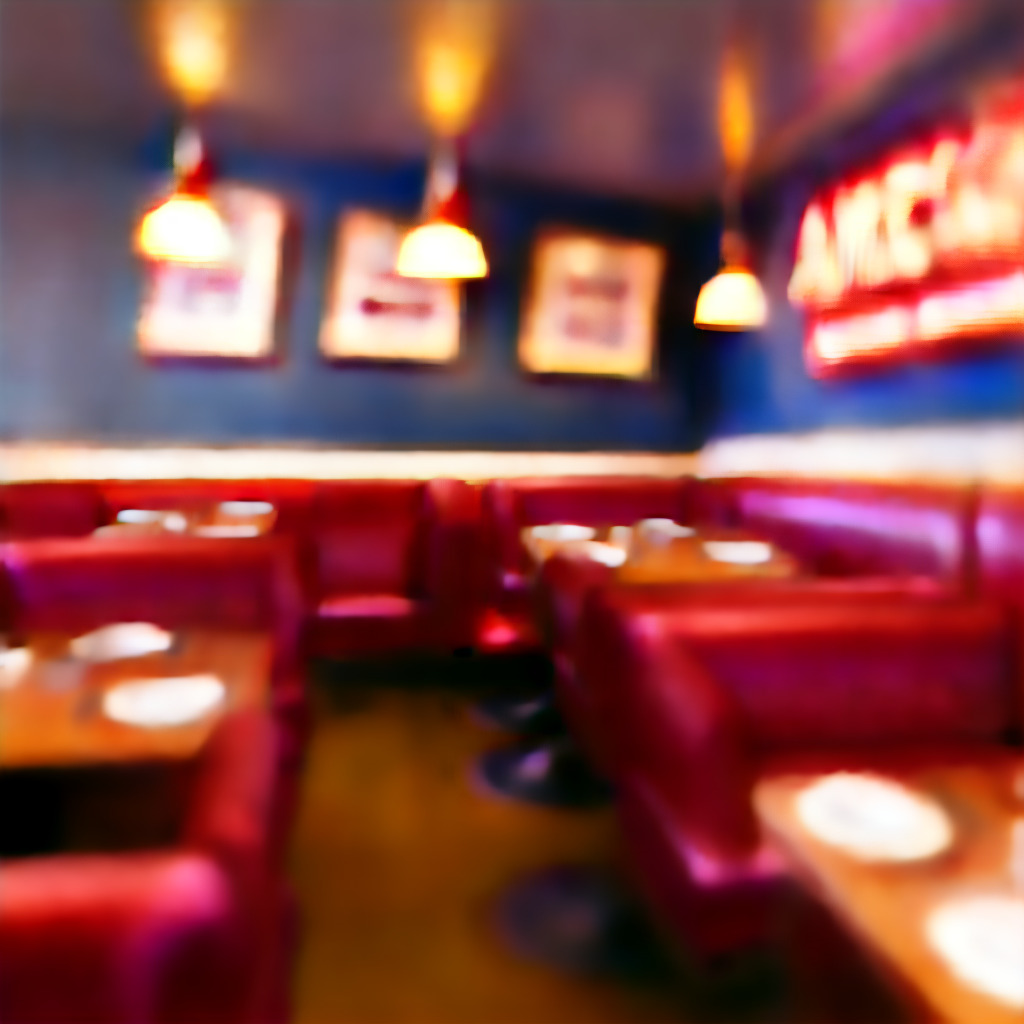} & 
        \includegraphics[width=0.24\linewidth]{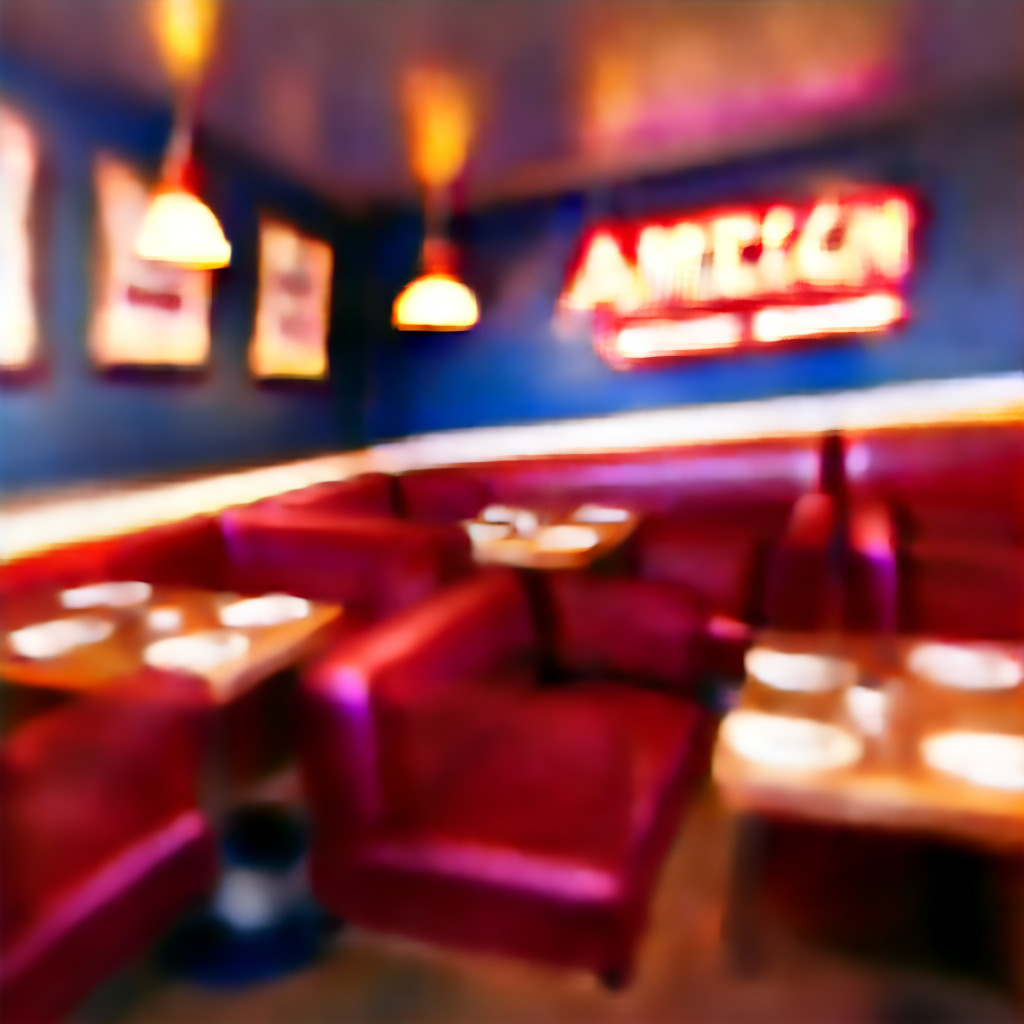}  \\
        & \multicolumn{3}{c}{``An American diner''} \\
        \includegraphics[width=0.24\linewidth]{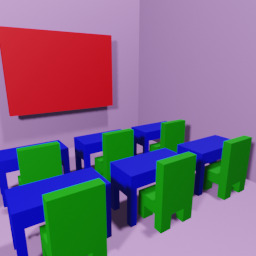} &
        \includegraphics[width=0.24\linewidth]{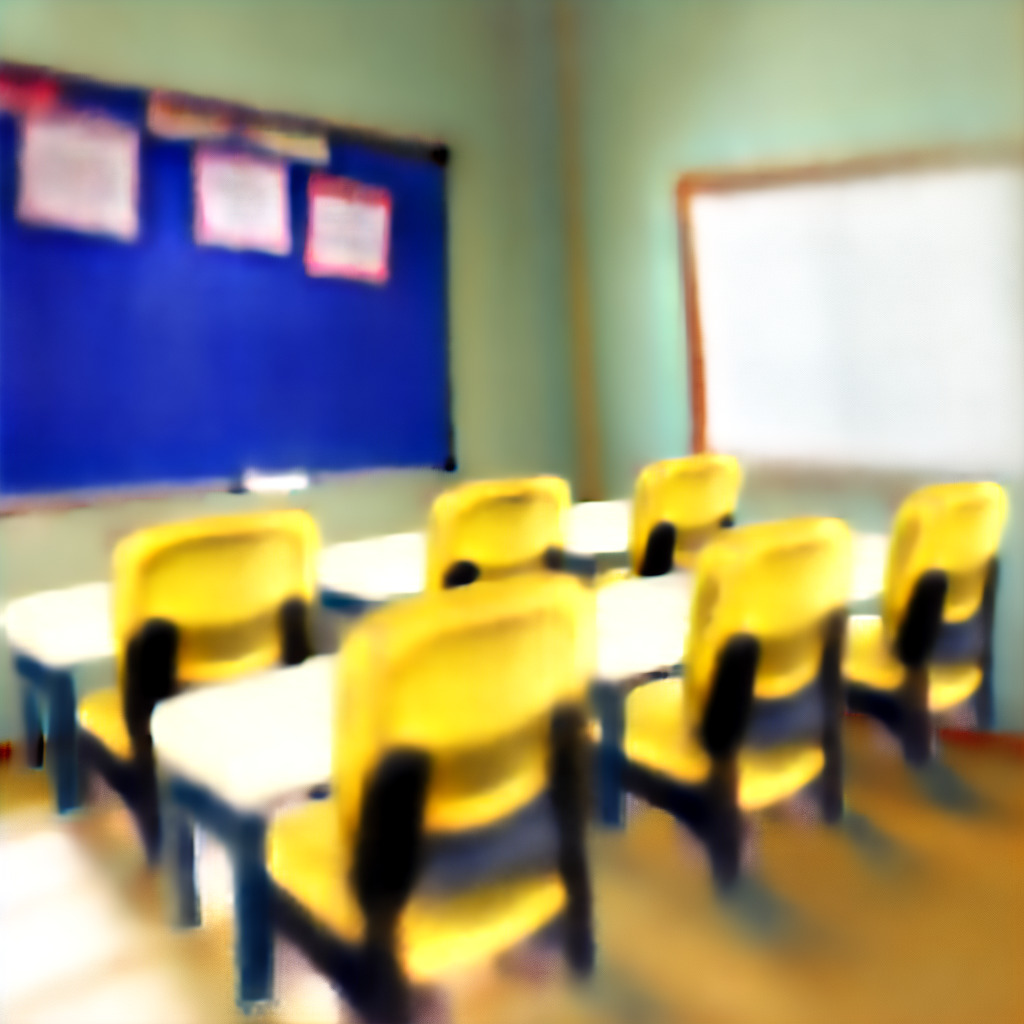} &
        \includegraphics[width=0.24\linewidth]{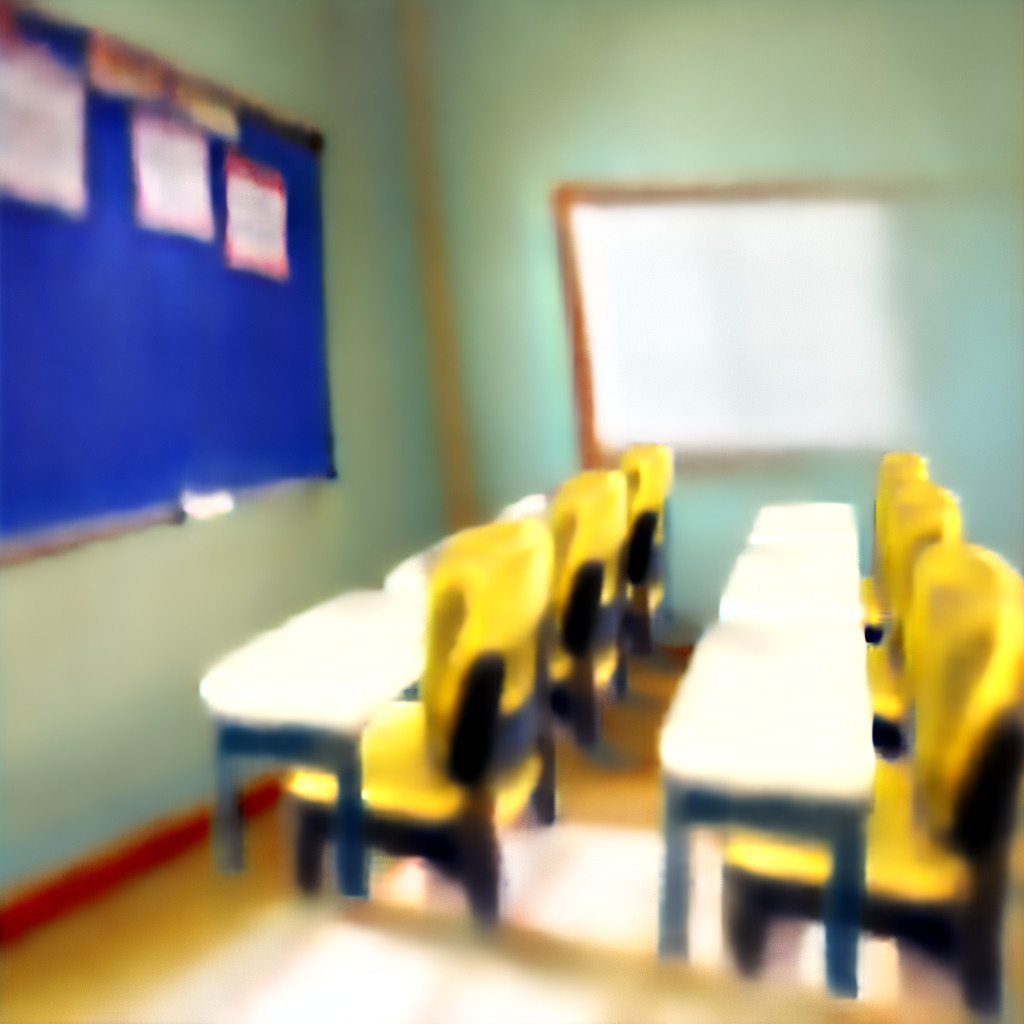} &
        \includegraphics[width=0.24\linewidth]{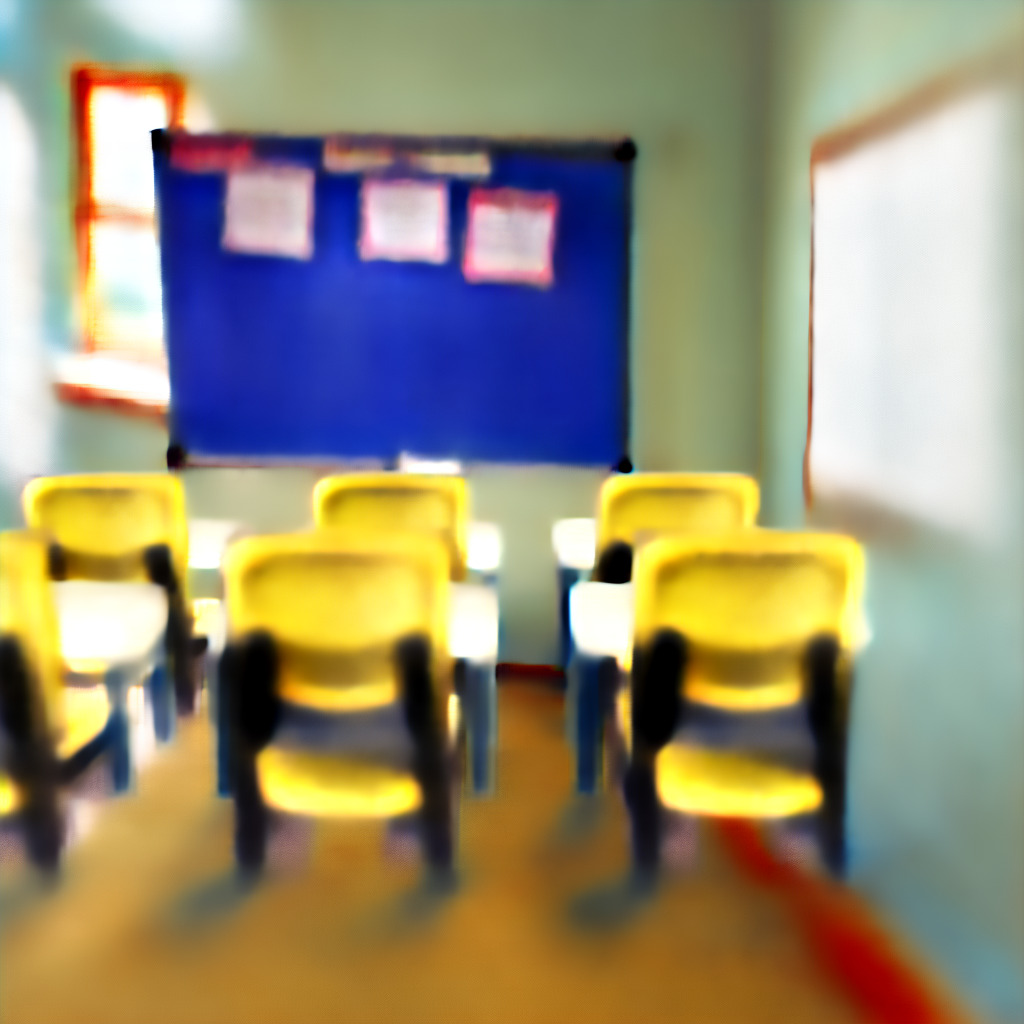} \\
        & \multicolumn{3}{c}{``A classroom''} 

    \end{tabular}}
    \caption{{\bf Scene generation results.} Our method is able to handle complex scenes with multiple repeating objects. 
    } 
    \vspace{-0.3cm}
    \label{fig:more_scenes}
\end{figure} 
\begin{figure}
    \centering
    \setlength{\tabcolsep}{0pt}
    {\small
    \begin{tabular}{c c c }
        \includegraphics[width=0.33\linewidth]{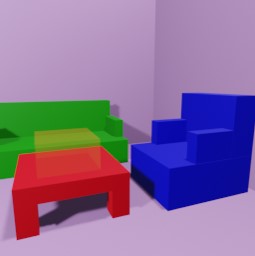} &
        \includegraphics[width=0.33\linewidth]{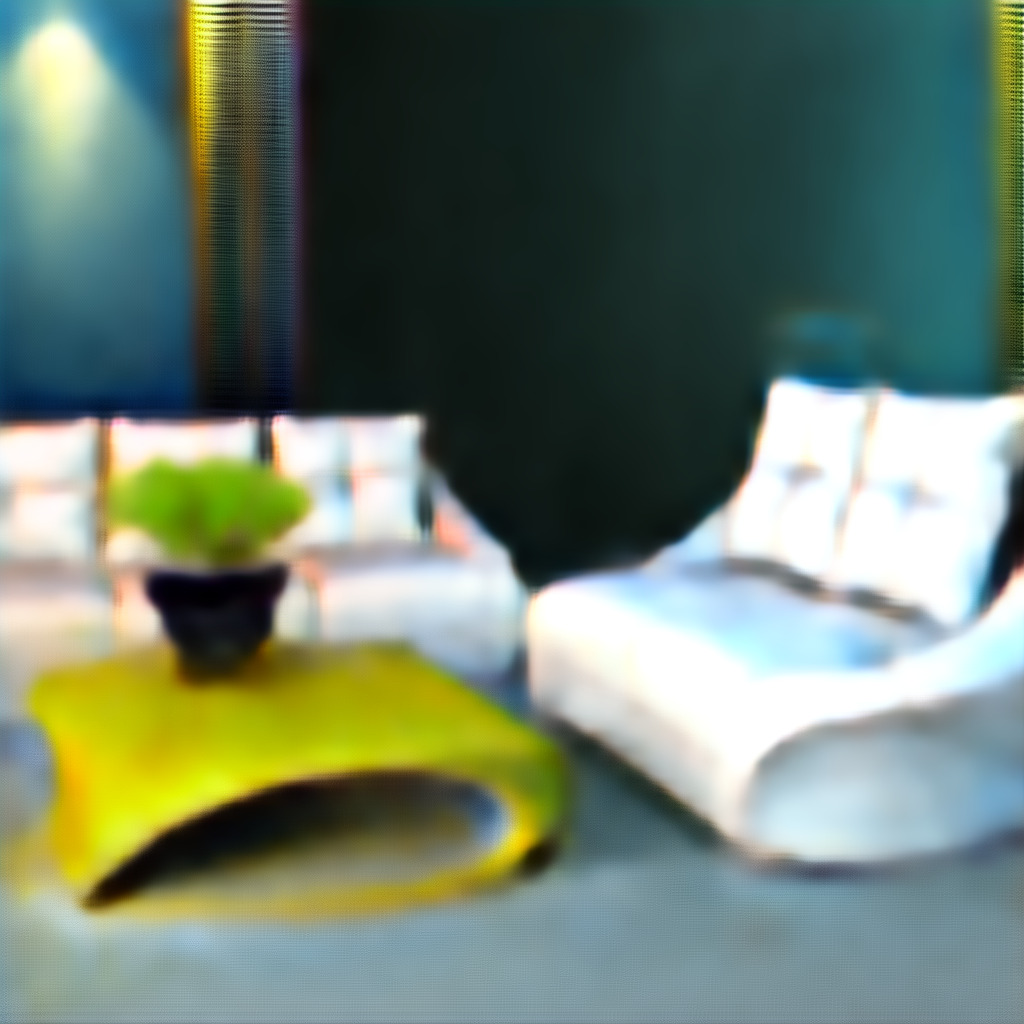} & 
        \includegraphics[width=0.33\linewidth]{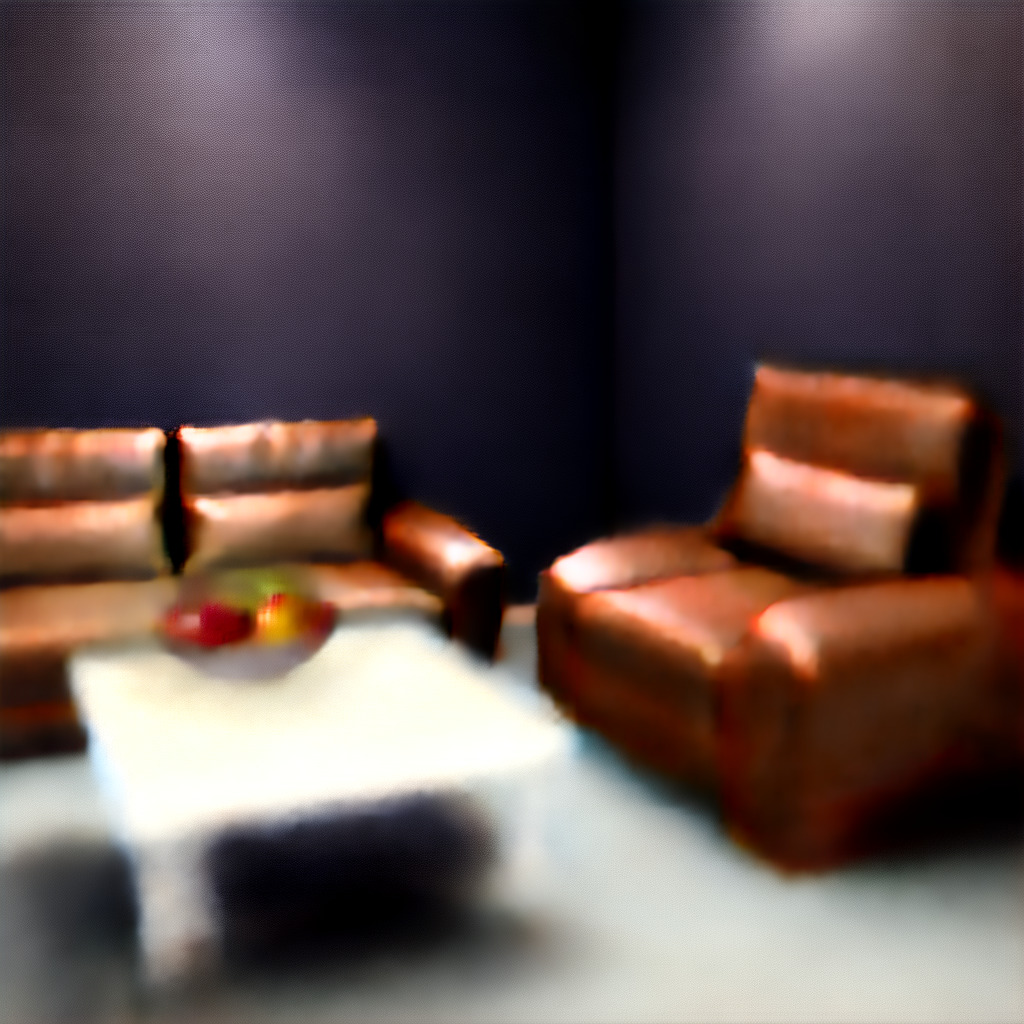}  \\
        Input Settings & ``a flowerpot'' & ``a fruit bowl''

    \end{tabular}}
    \caption{{\bf Objects with no shape priors.} For some objects it is beneficial to define only their location without explicitly stating their shape. Here we highlight how different objects can be generated on top of the table.}
    \label{fig:no_sketch_shape}
\end{figure}

\subsection{Scene Generation}
\paragraph{Qualitative Results} Figure~\ref{fig:different styles} shows results of generated scenes guided by different text prompts. 
One can see that our method closely follows the given proxies in terms of location and coarse shape. 
For example, observe how in the first row the wardrobe and bed are consistently placed in the scene even when the guiding text prompt changes.
Nevertheless, our method is still able to expressively alter the style of the generated shapes according to the guiding text prompt given the geometric constraints, thus offering both control and expressiveness.
In Figure~\ref{fig:more_scenes}, the objects are used multiple times within a single scene.
This is done by defining a set of object proxies that share the same object NeRF, explicitly enforcing similarity between a set of duplicated objects in a scene, which is difficult to achieve when generating a scene directly without object proxies.
This design choice also enables aggregating information about how an object is viewed from different locations and viewing angles into a single NeRF model, instead of training each model separately. 
Figure~\ref{fig:no_sketch_shape} shows results where an object in the scene is generated without any specific geometry constraint, but rather only with its respective location in the scene and a guiding text prompt, showing the flexibility of our controls.
Finally, Figure~\ref{fig:convergence} presents the exact text prompts used for a specific scene along with the convergence of each object and the scene as a whole during the optimization process.

\vspace{-0.3cm}
\paragraph{Comparisons} Recent text-to-3D methods generate the scene as a whole and do not utilize a composable representation. This makes it harder to control and can cause it to fail on complex scenes. 
To highlight this issue, Figure~\ref{fig:comparison} shows a result of Latent-NeRF~\cite{metzer2022latent} on our scene text prompt. Observe how Latent-NeRF struggles with generating the complex scene.
We note that although methods like Dreamfusion~\cite{poole2022dreamfusion} might be able to generate better scenes due to their larger diffusion model~\cite{imagen}, they still would not provide explicit control.
Furthermore, as Latent-NeRF is the basis of our local optimization process, comparison to it highlights the improvements gained using our scheme.

\newcommand{\imheight}{0.2\linewidth} 
\begin{figure}
    \centering
    \vspace{-0.5cm}
    \setlength{\tabcolsep}{0pt}
    {\small
    \begin{tabular}{c c c c c}
        \includegraphics[height=\imheight]{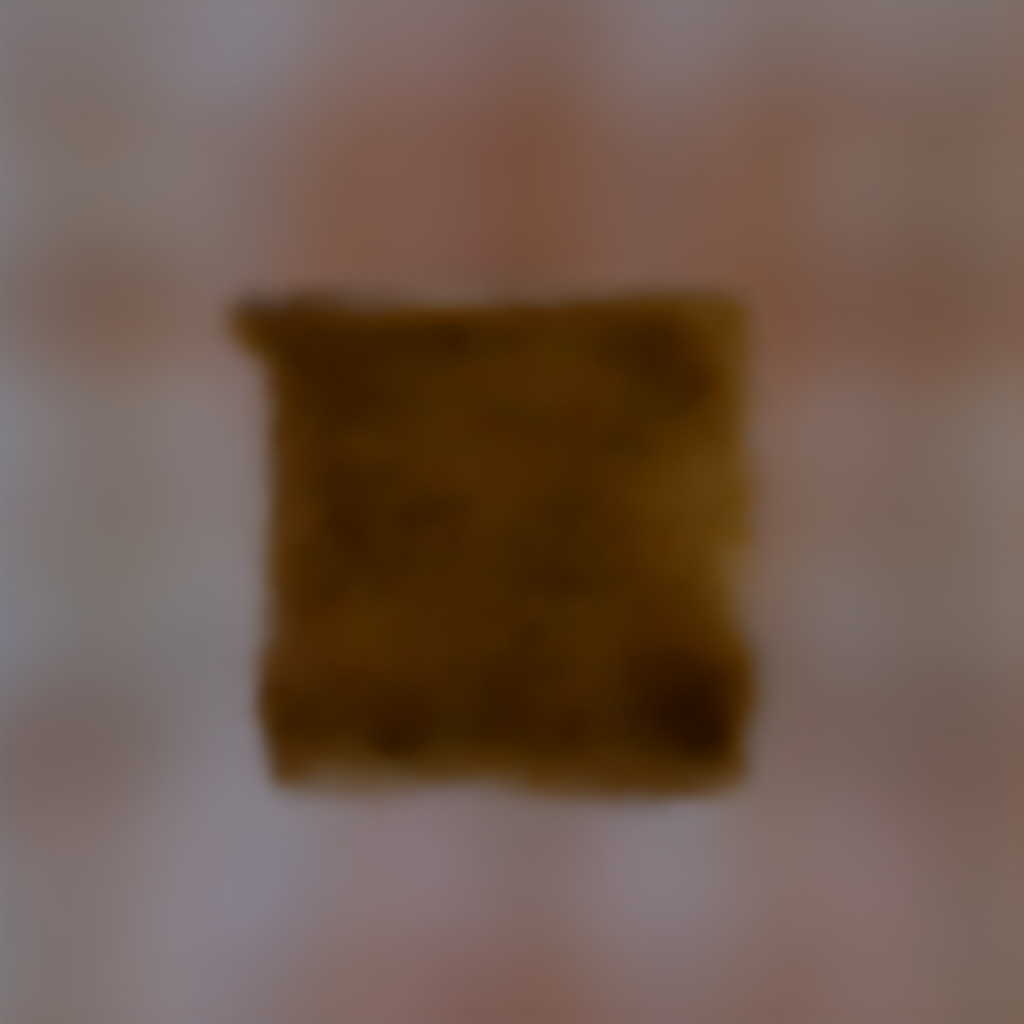} & 
        \includegraphics[height=\imheight]{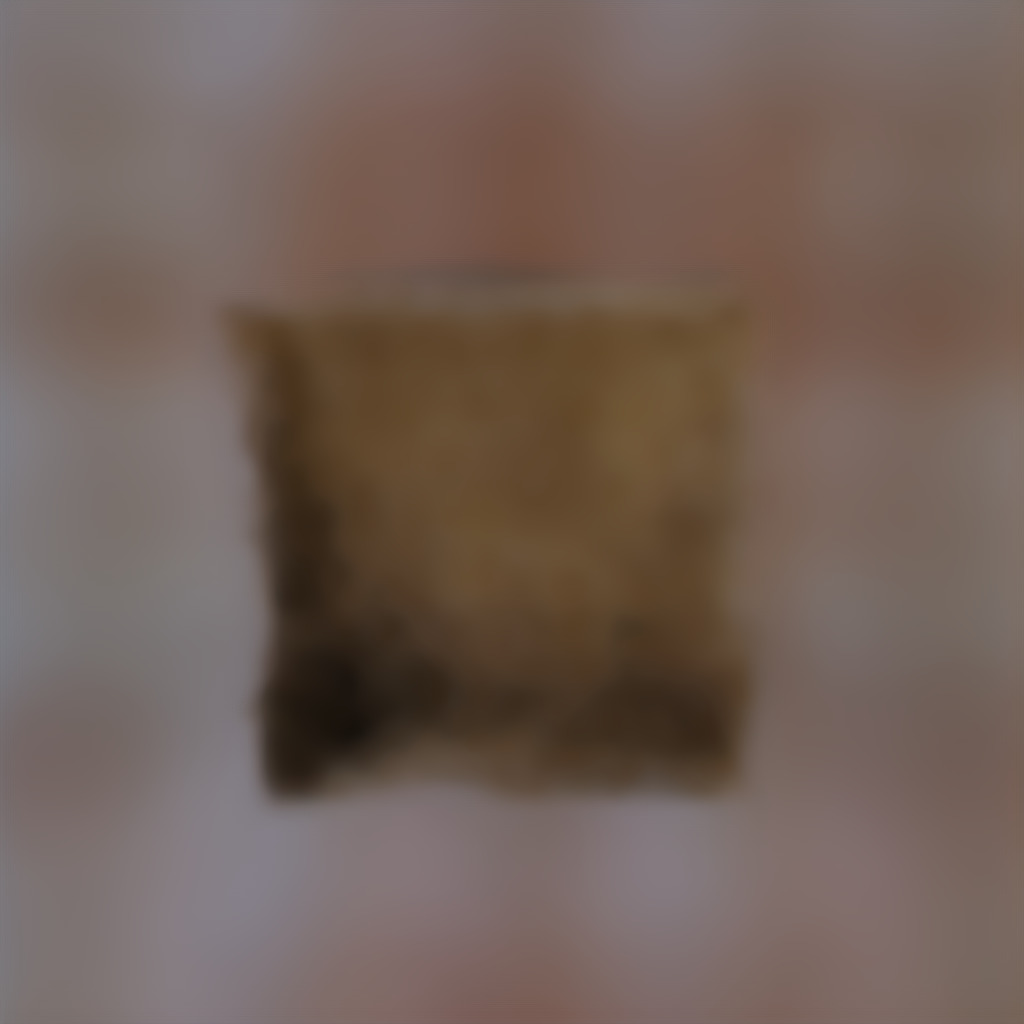} & 
        \includegraphics[height=\imheight]{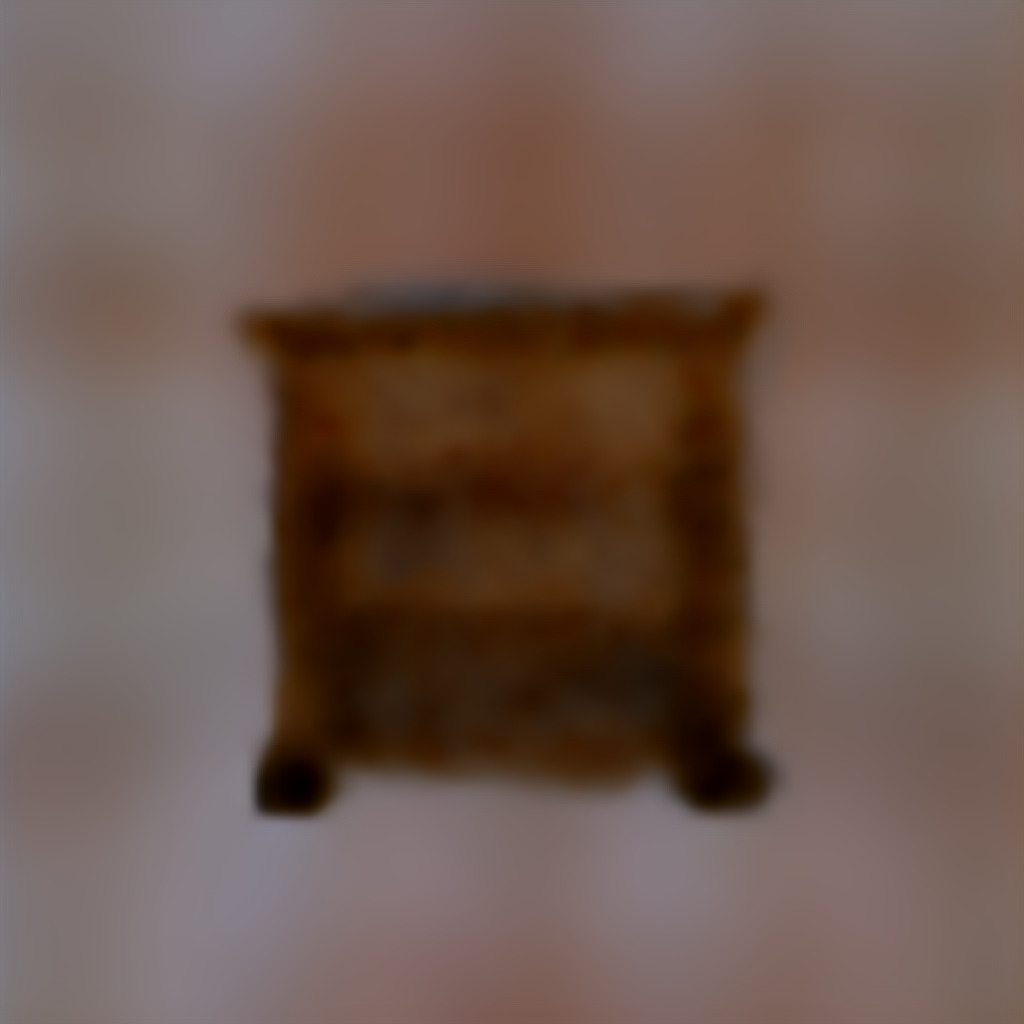} & 
        \includegraphics[height=\imheight]{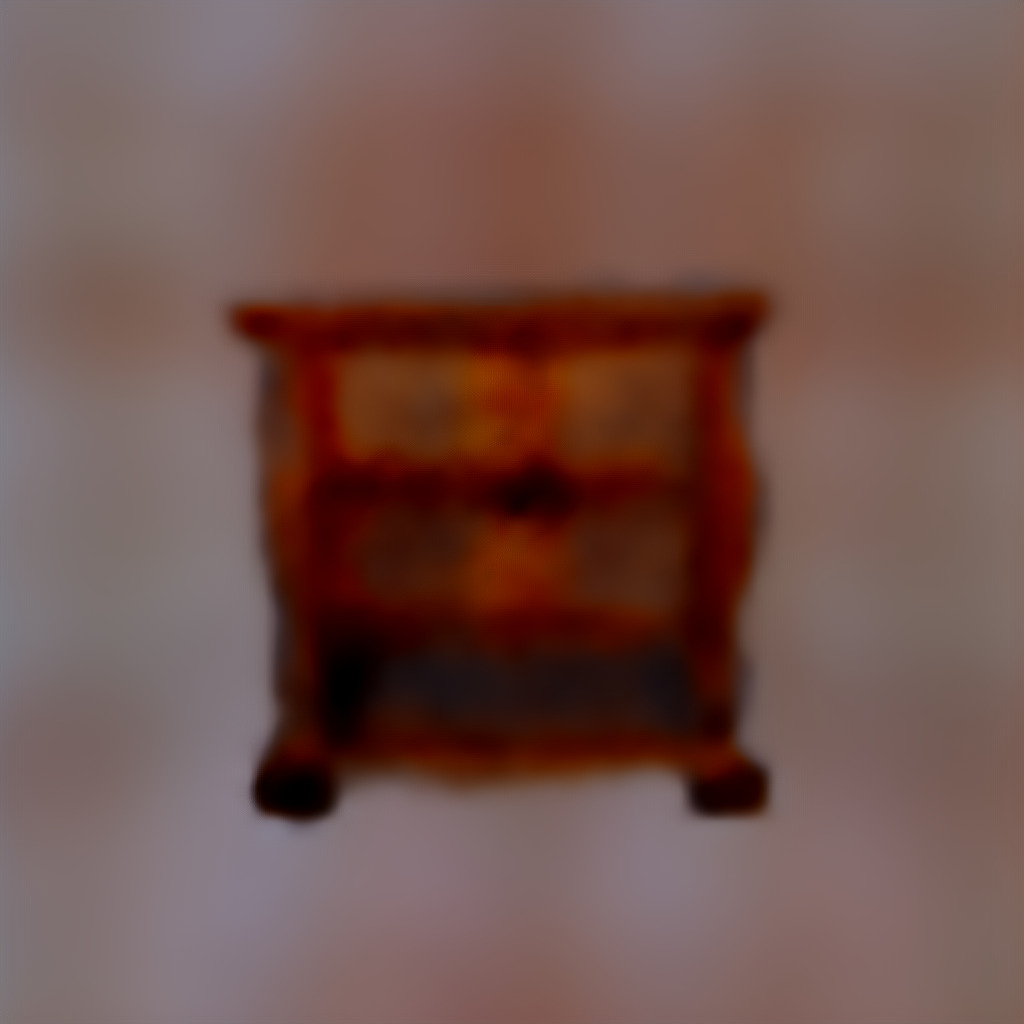} & 
        \includegraphics[height=\imheight]{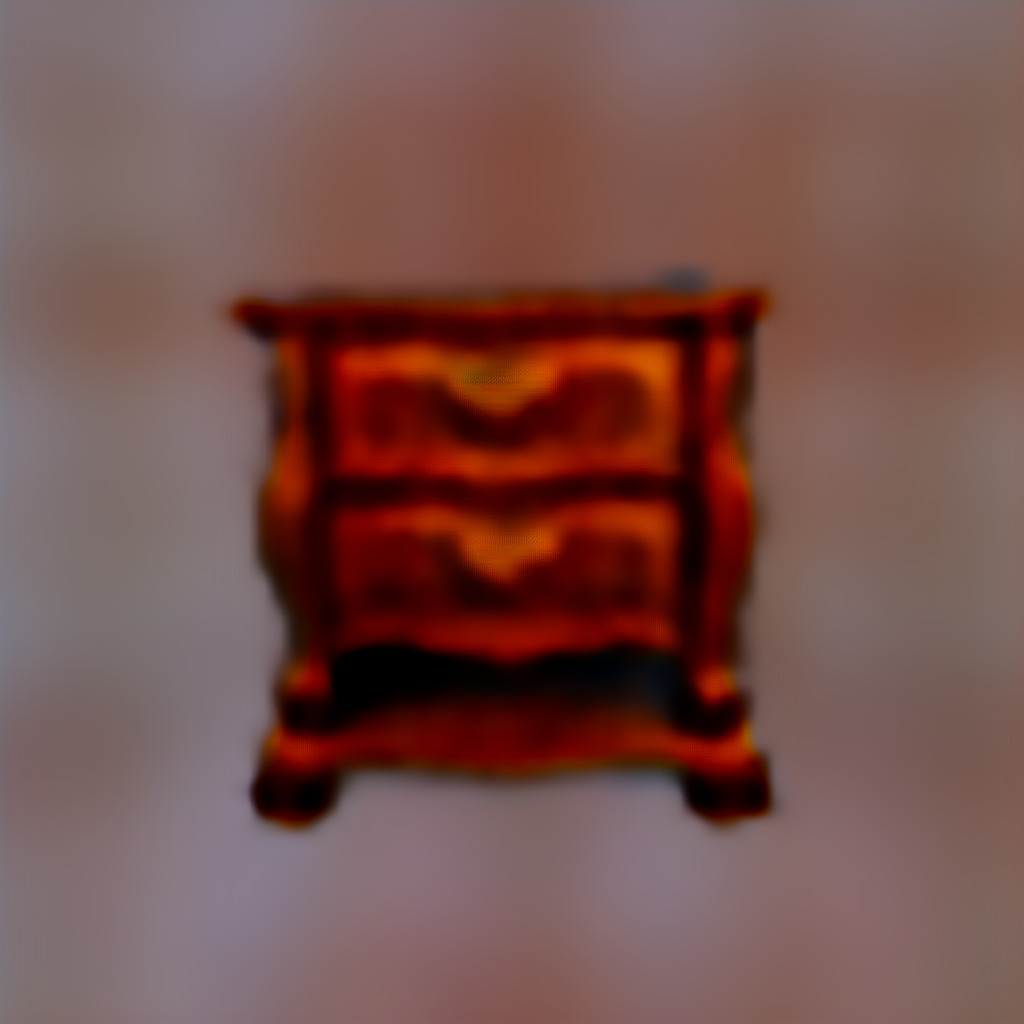} \\
         \multicolumn{5}{c}{``a baroque nightstand''} \\
        \includegraphics[height=\imheight]{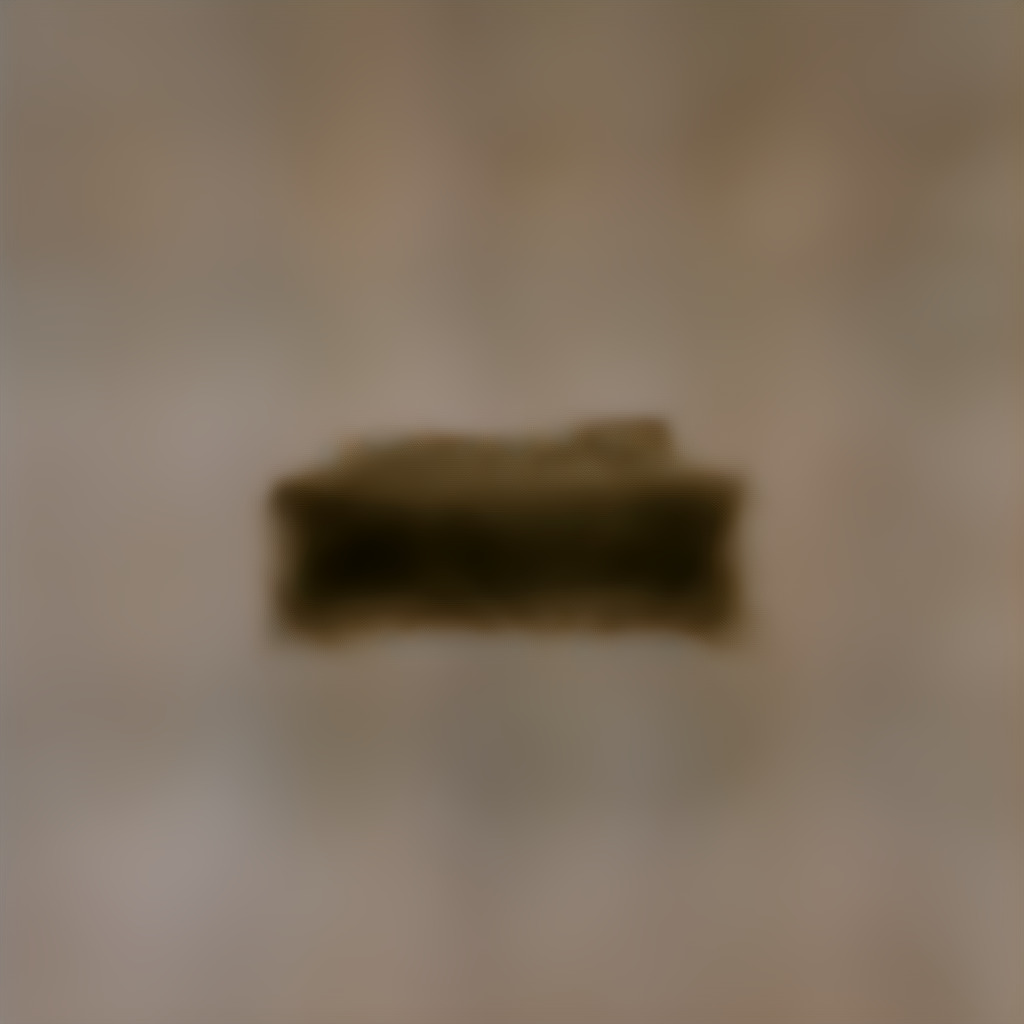} & 
        \includegraphics[height=\imheight]{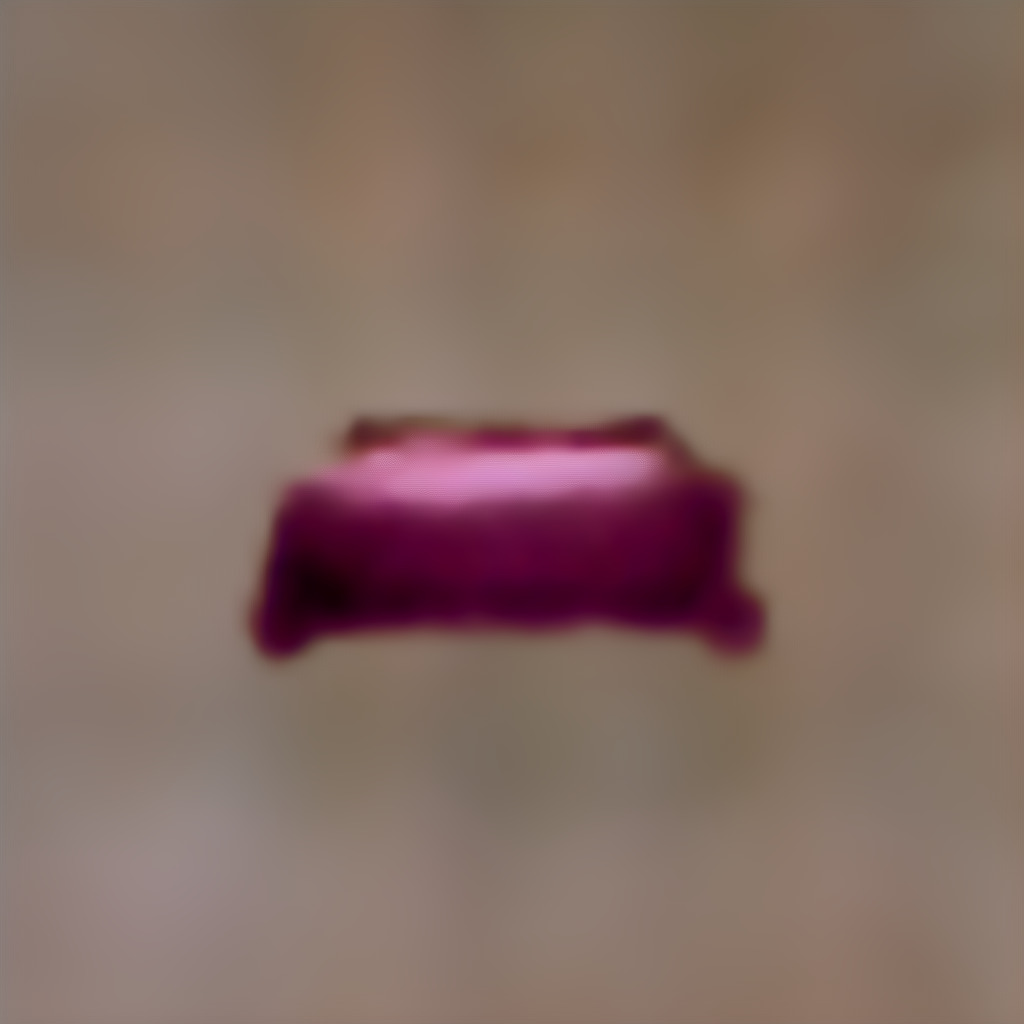} & 
        \includegraphics[height=\imheight]{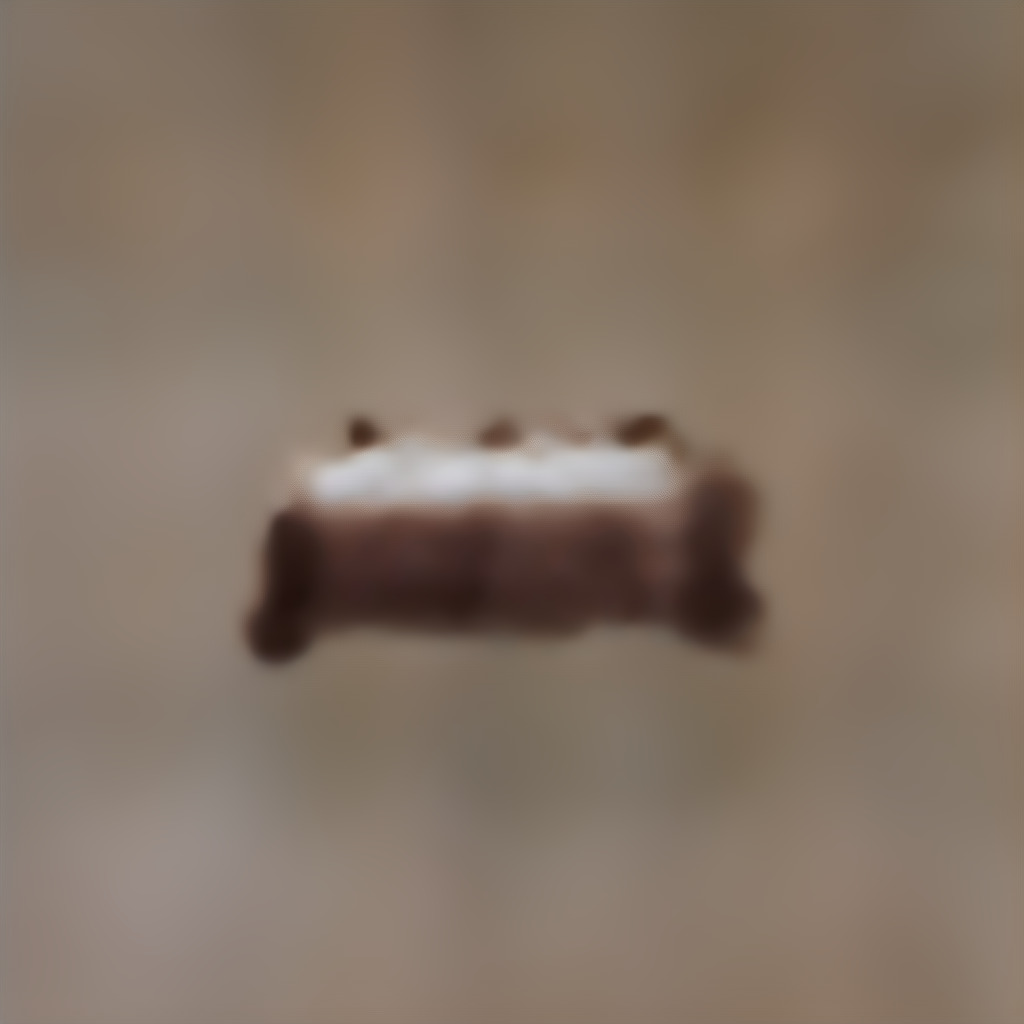} & 
        \includegraphics[height=\imheight]{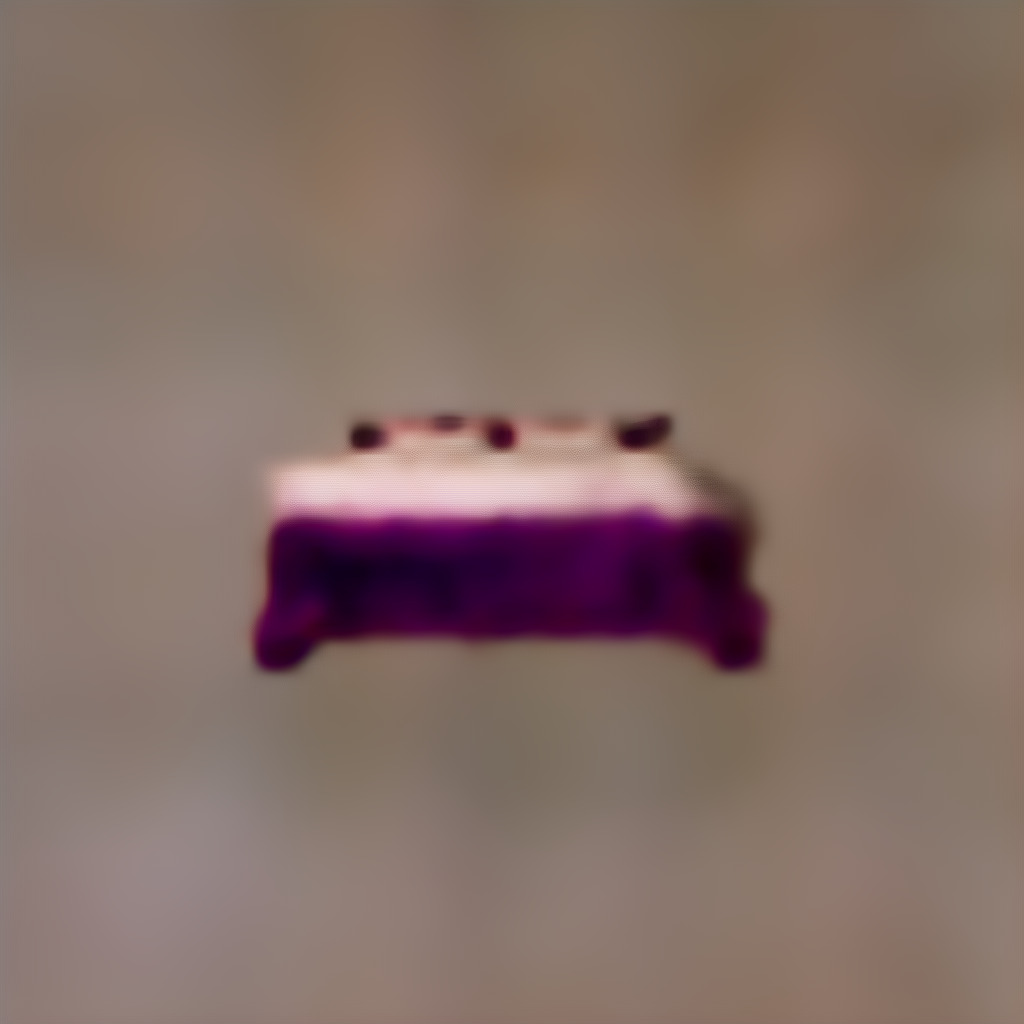} &
        \includegraphics[height=\imheight]{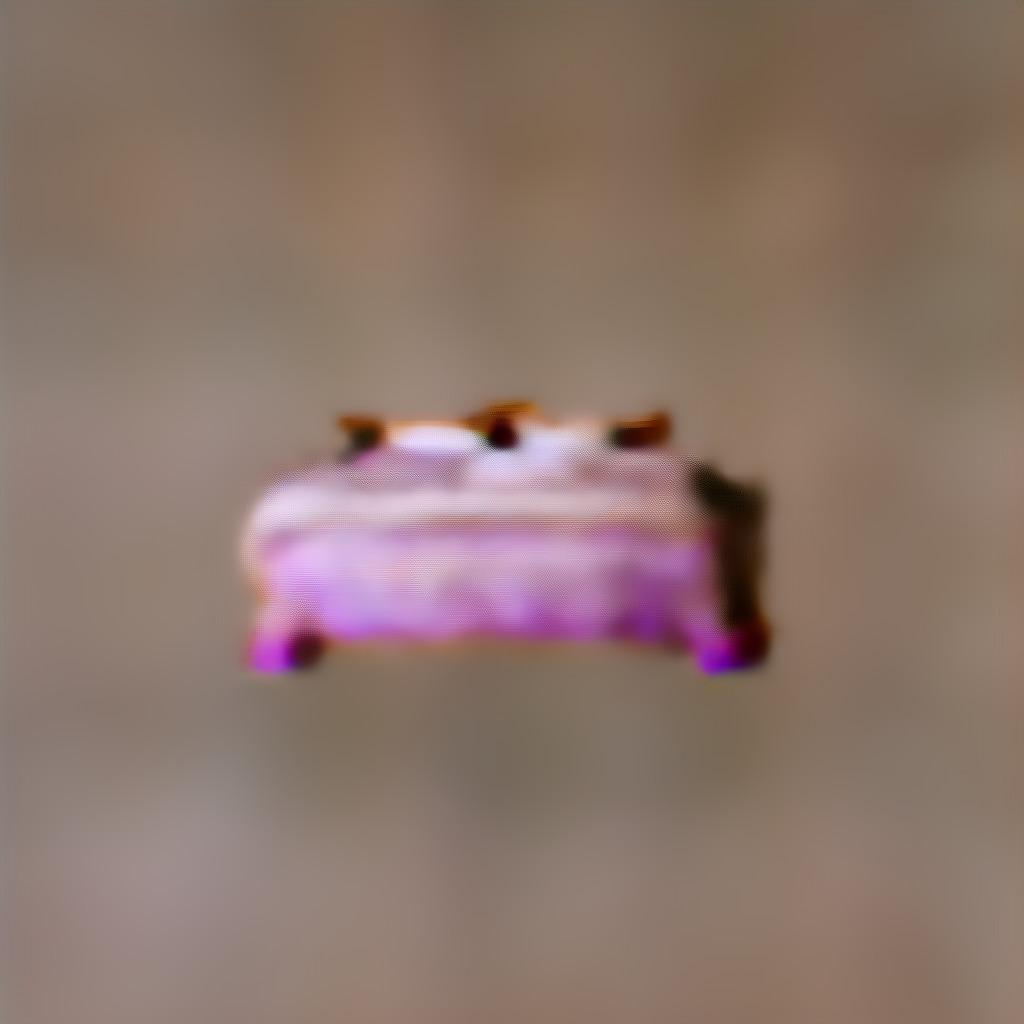} \\
        \multicolumn{5}{c}{``a baroque bed''} \\
        \includegraphics[height=\imheight]{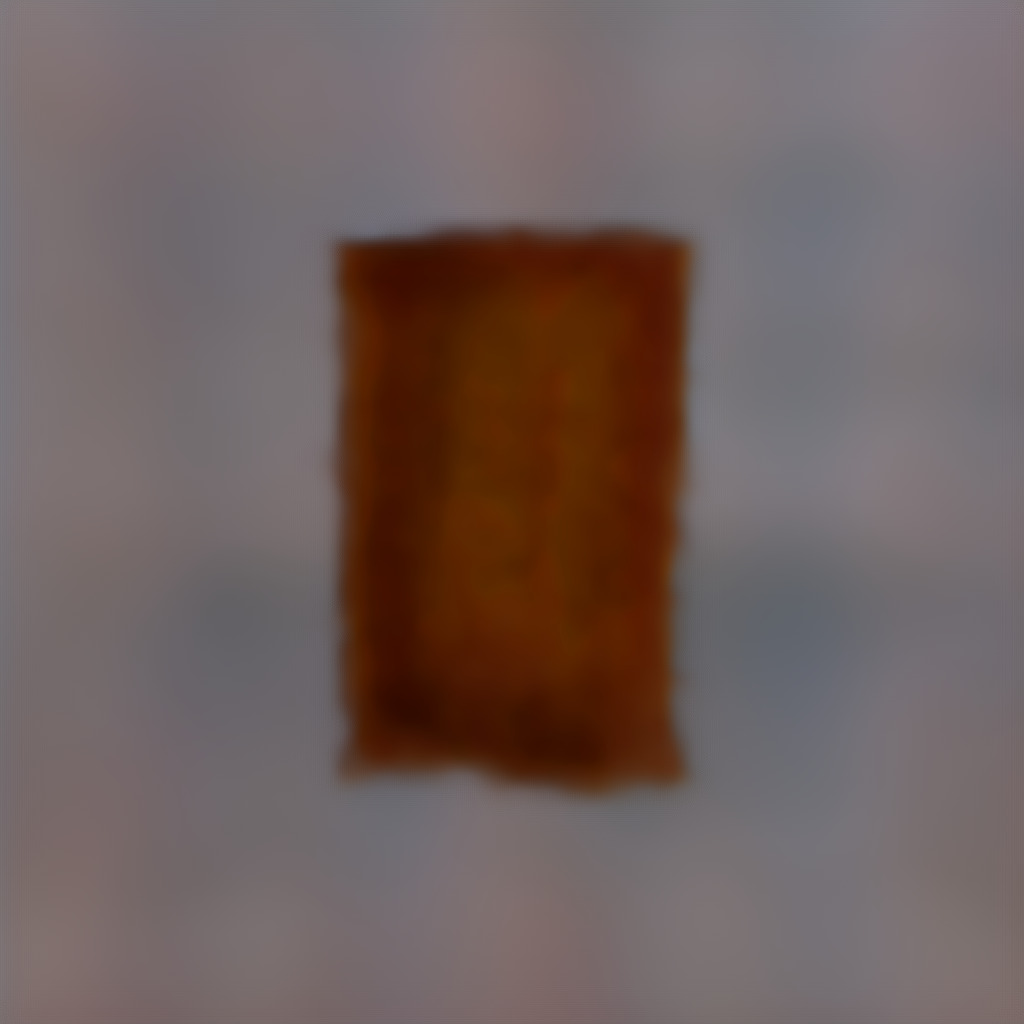} & 
        \includegraphics[height=\imheight]{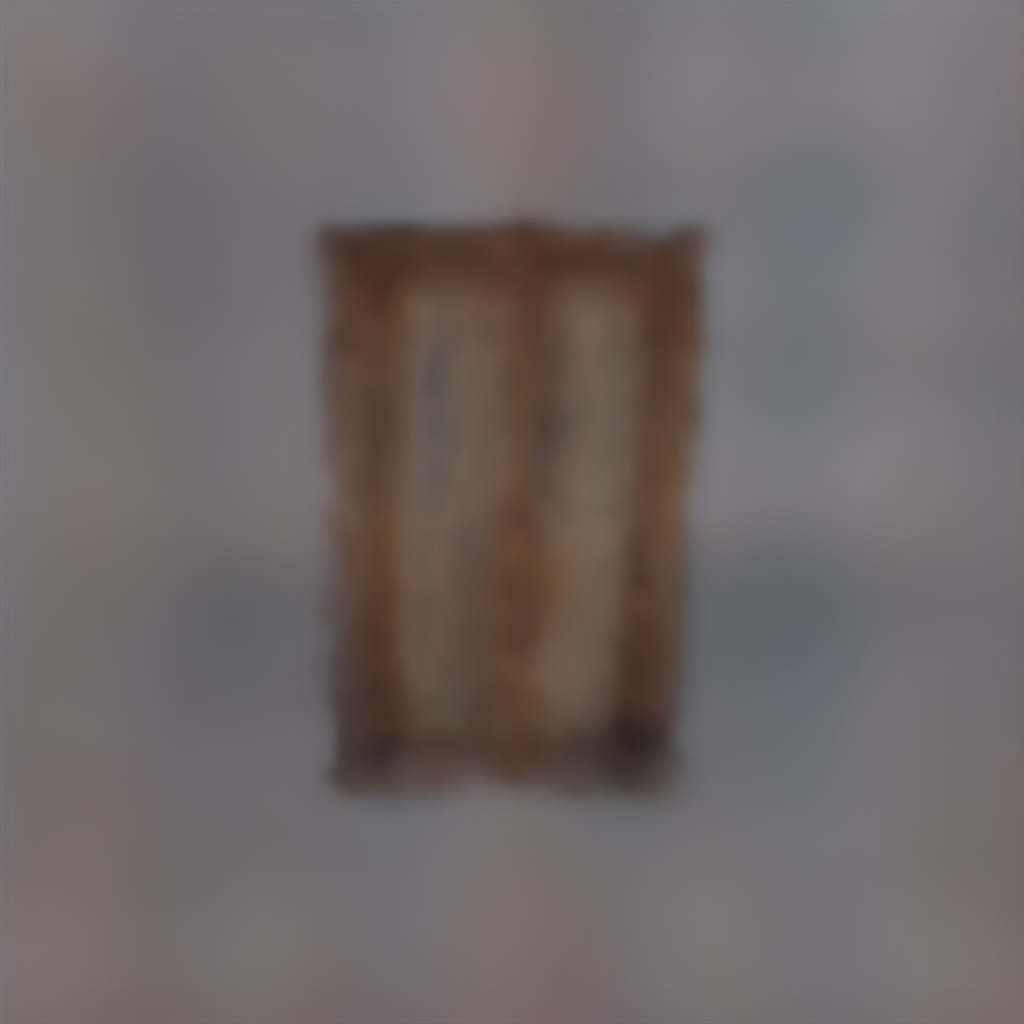} & 
        \includegraphics[height=\imheight]{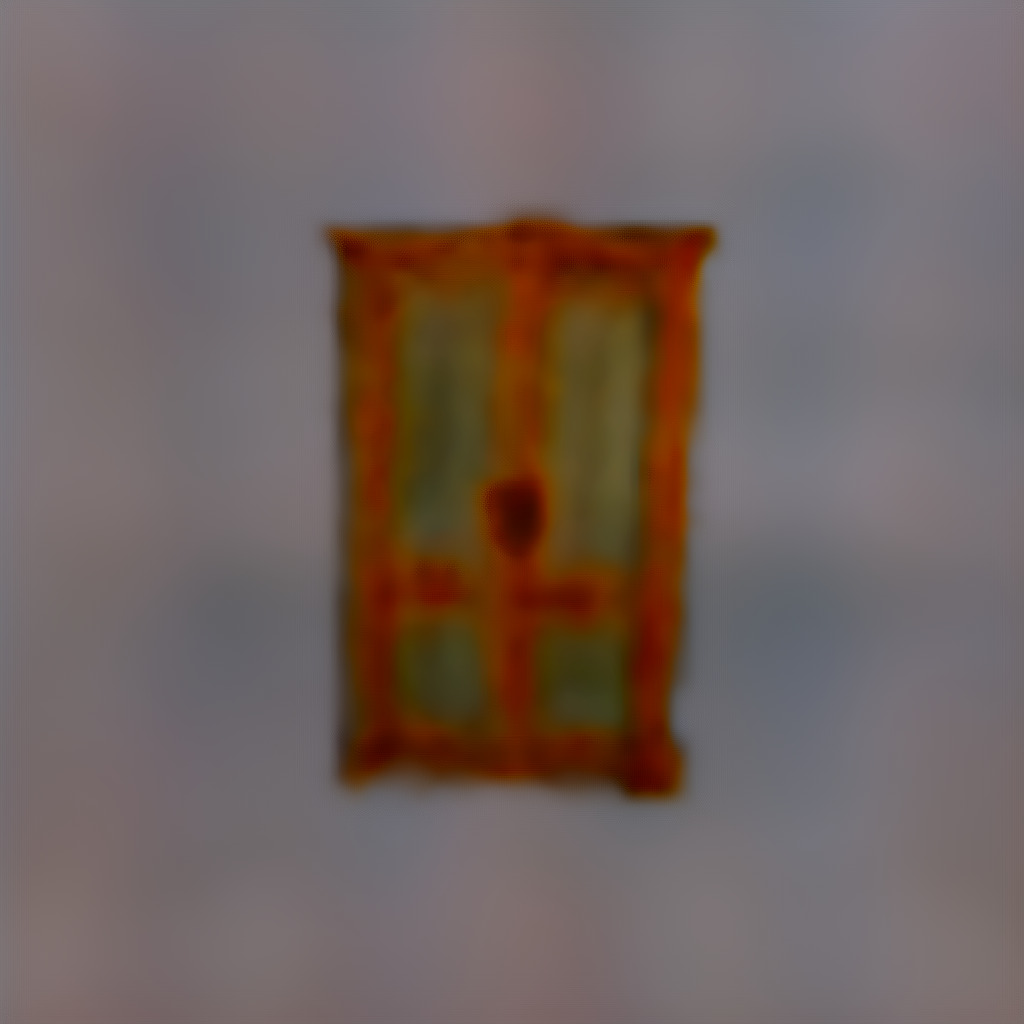} & 
        \includegraphics[height=\imheight]{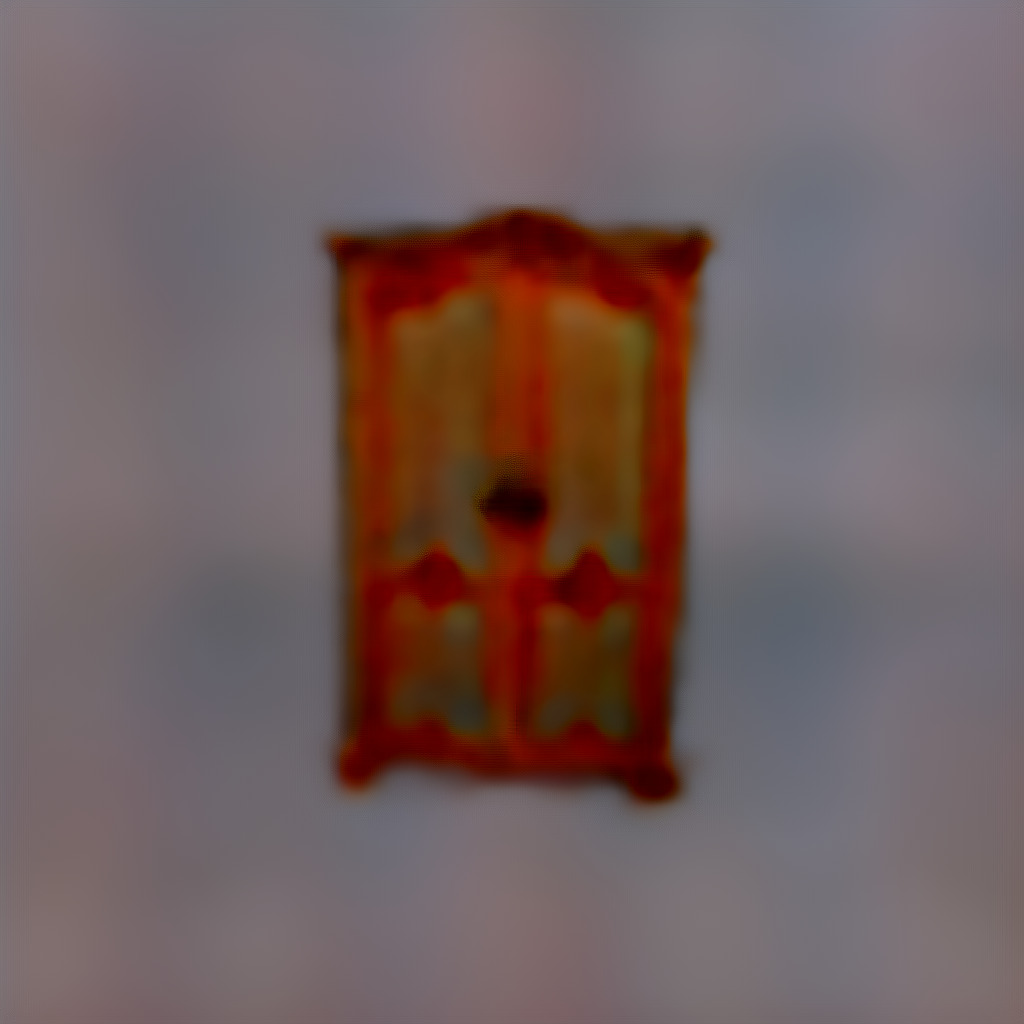} &
        \includegraphics[height=\imheight]{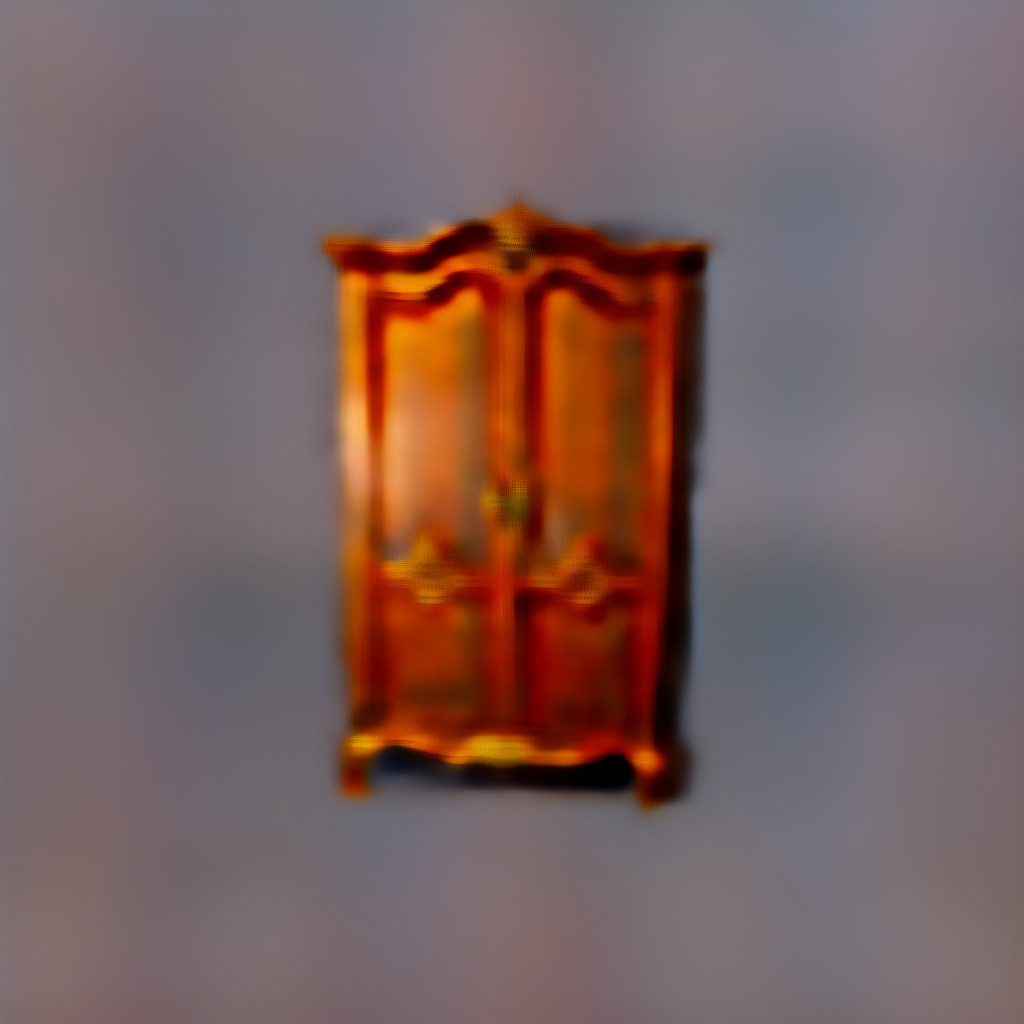} \\ 
        \multicolumn{5}{c}{``a baroque style wardrobe, closed doors''} \\
        \includegraphics[height=\imheight]{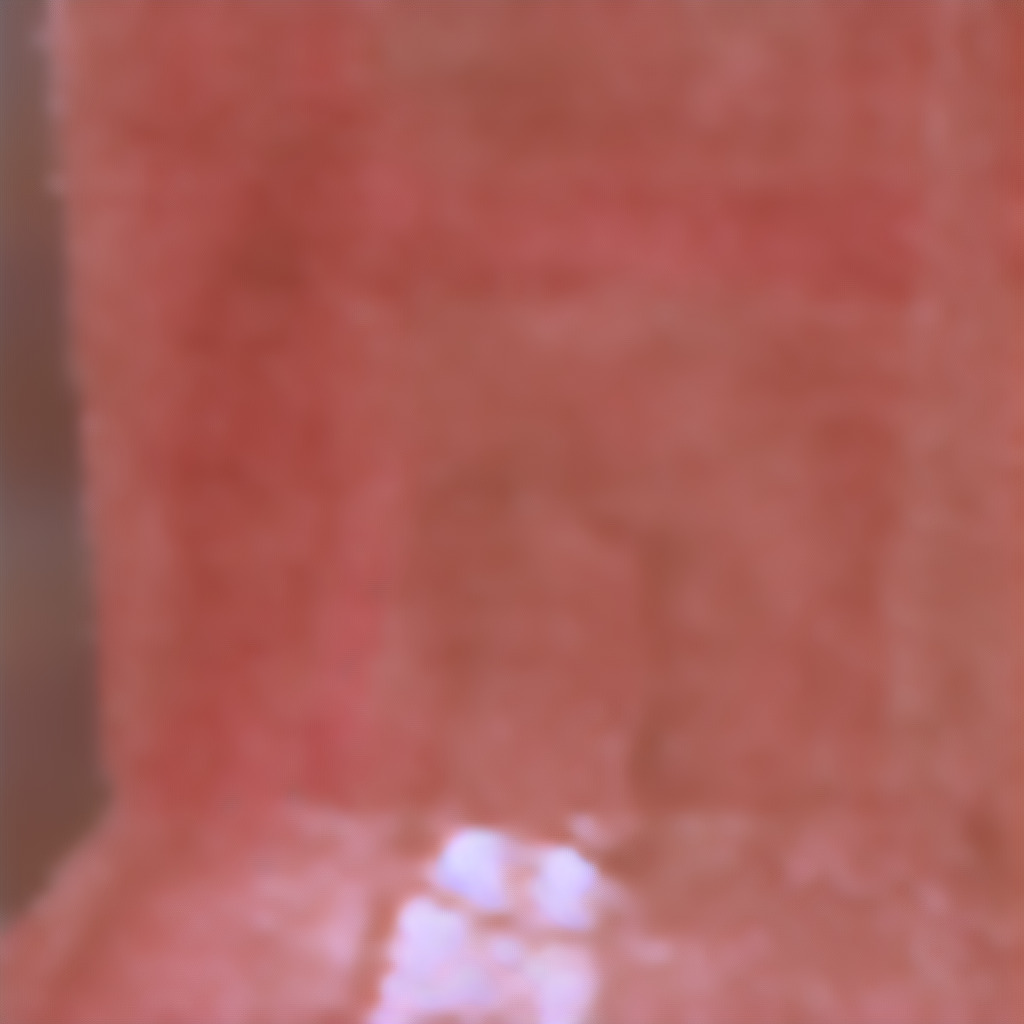} & 
        \includegraphics[height=\imheight]{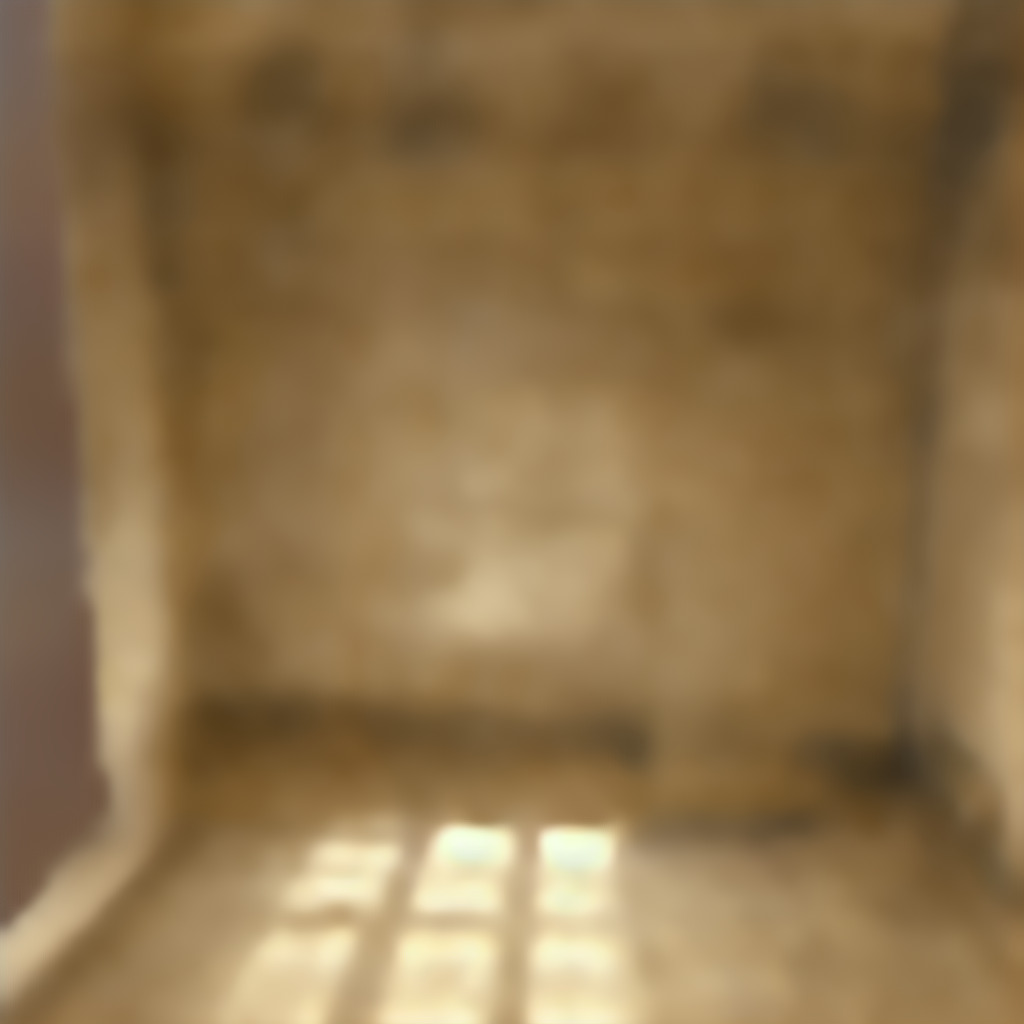} & 
        \includegraphics[height=\imheight]{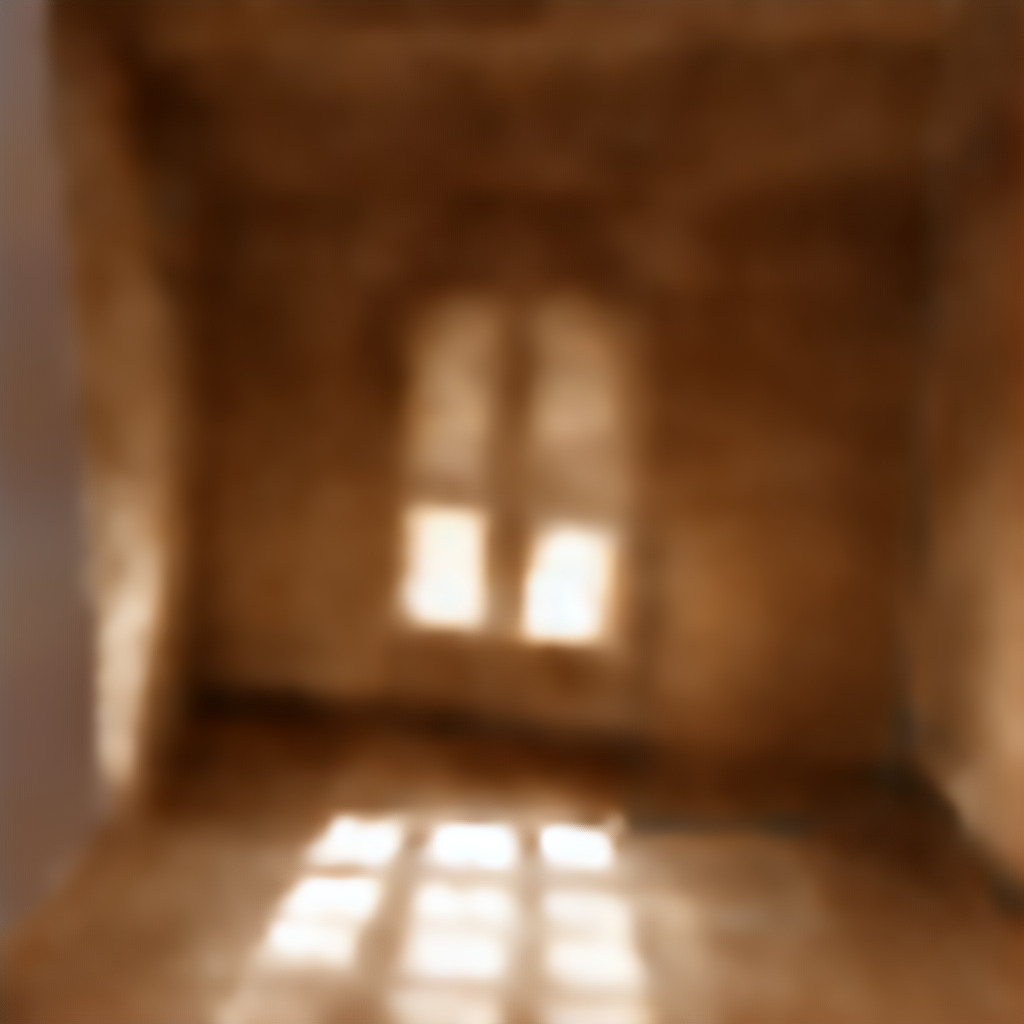} & 
        \includegraphics[height=\imheight]{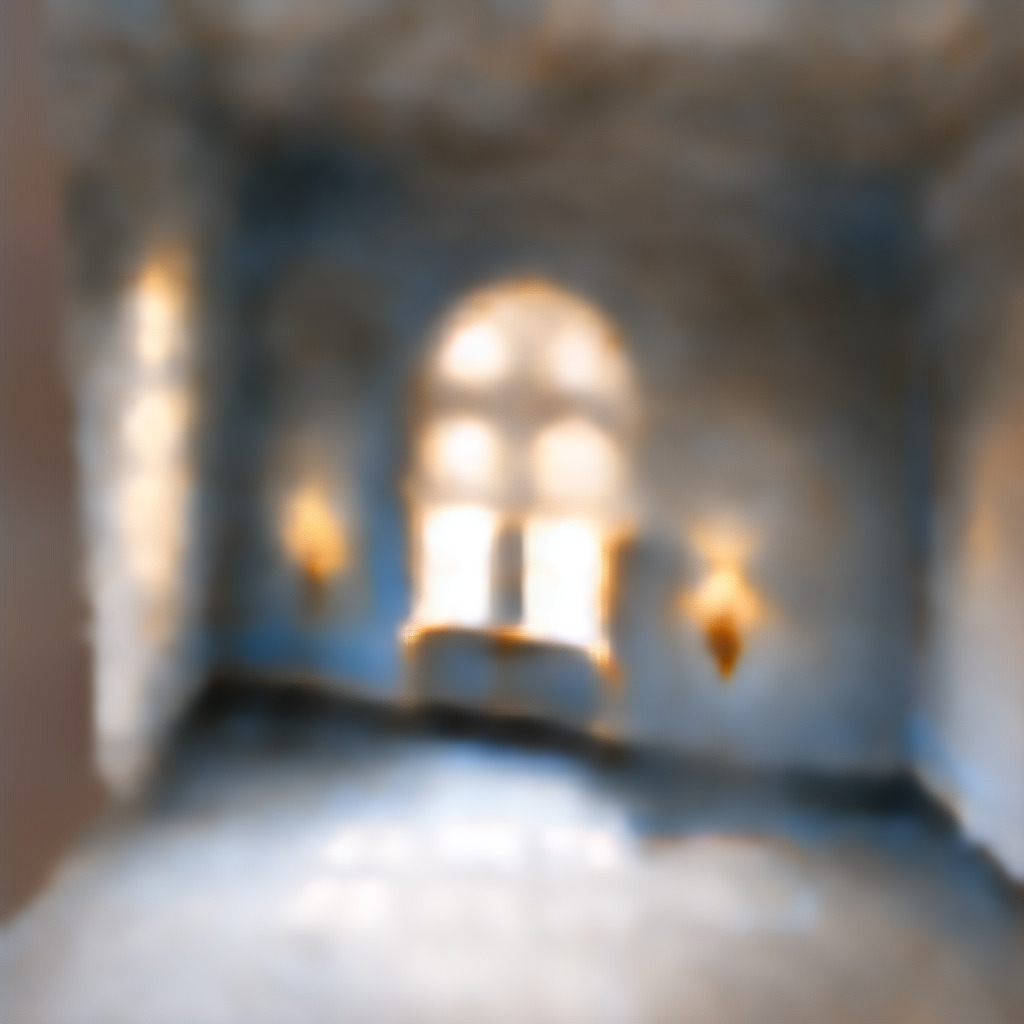} &
        \includegraphics[height=\imheight]{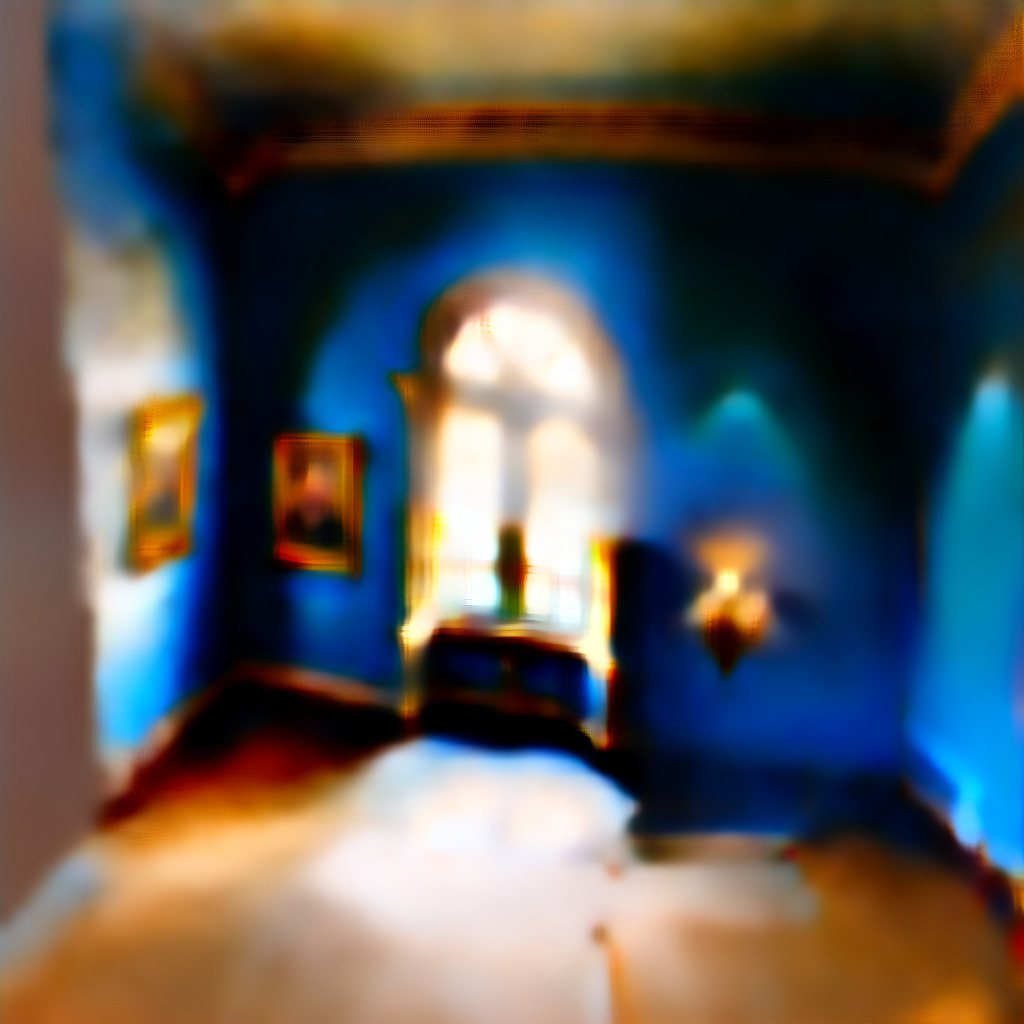} \\
        \multicolumn{5}{c}{``an empty baroque style room with windows''} \\
        \includegraphics[height=\imheight]{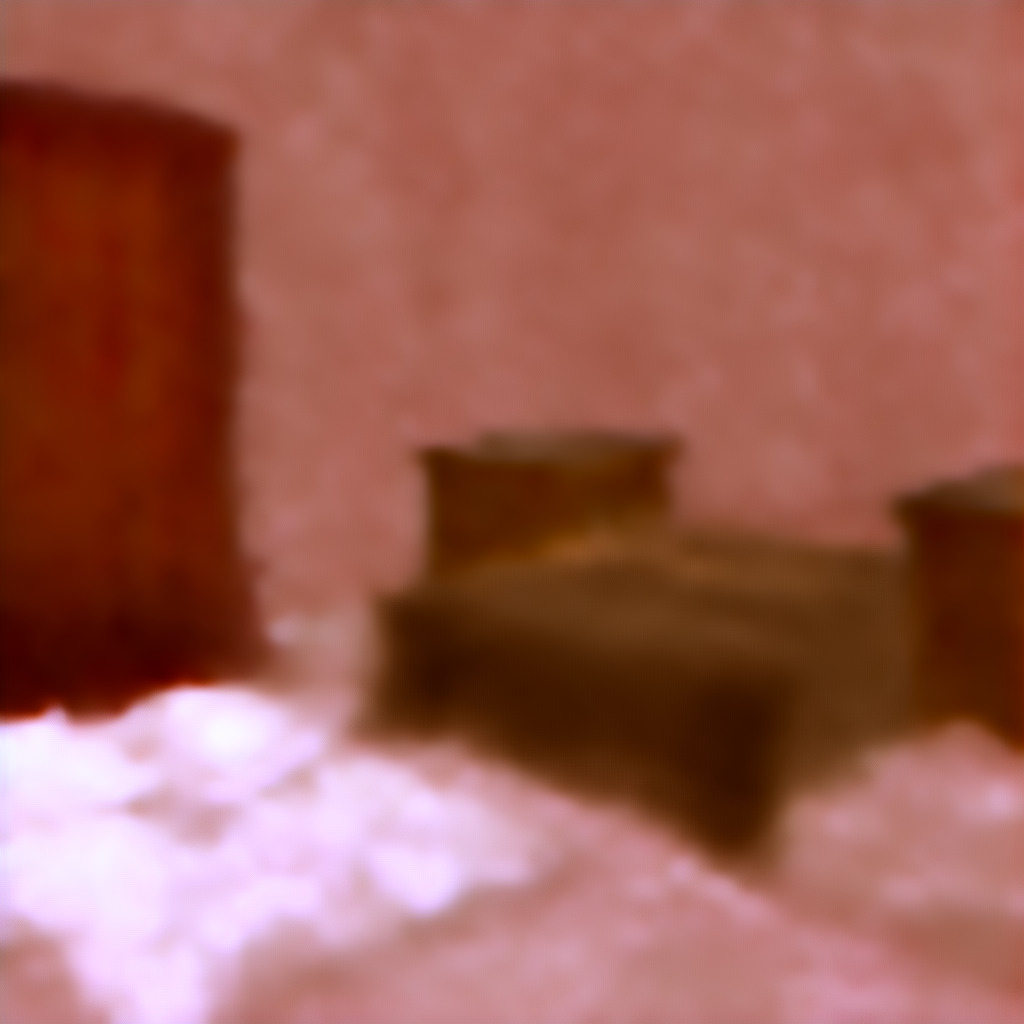} & 
        \includegraphics[height=\imheight]{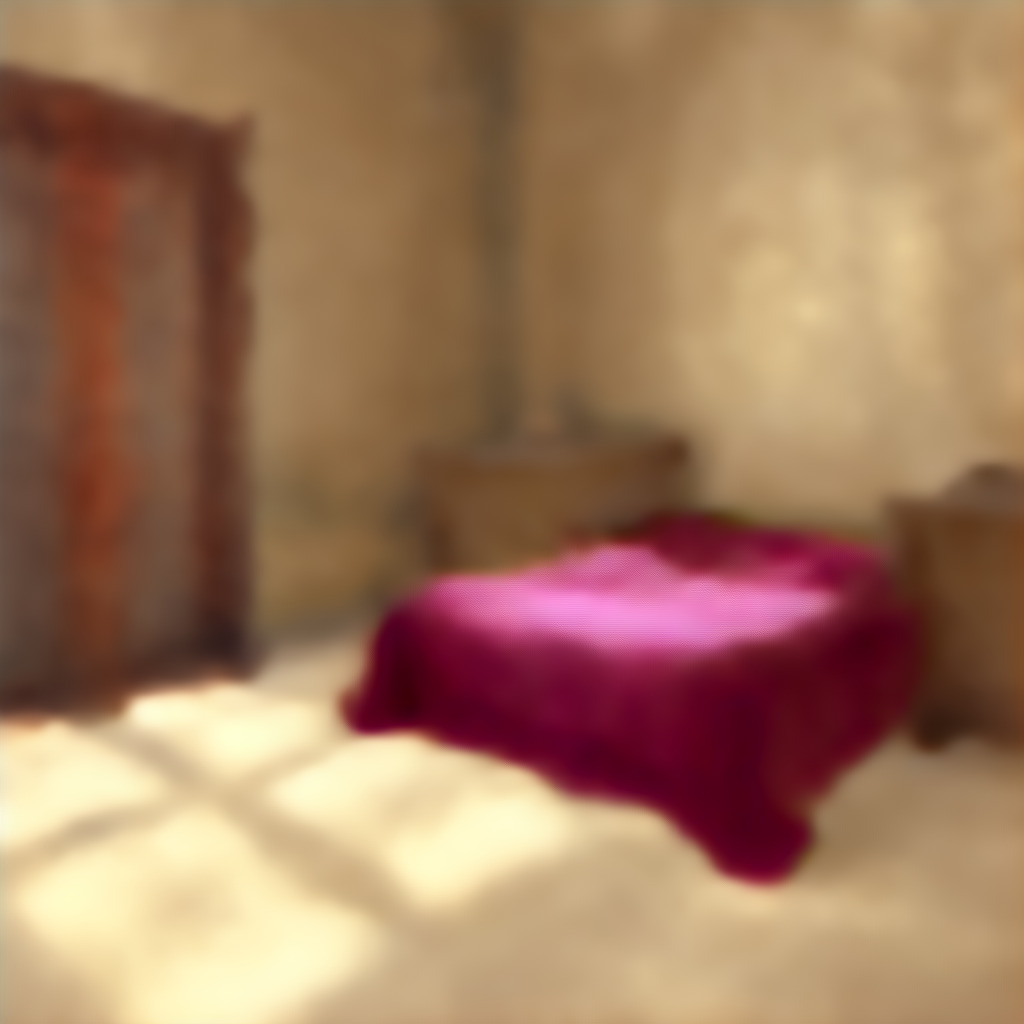} & 
        \includegraphics[height=\imheight]{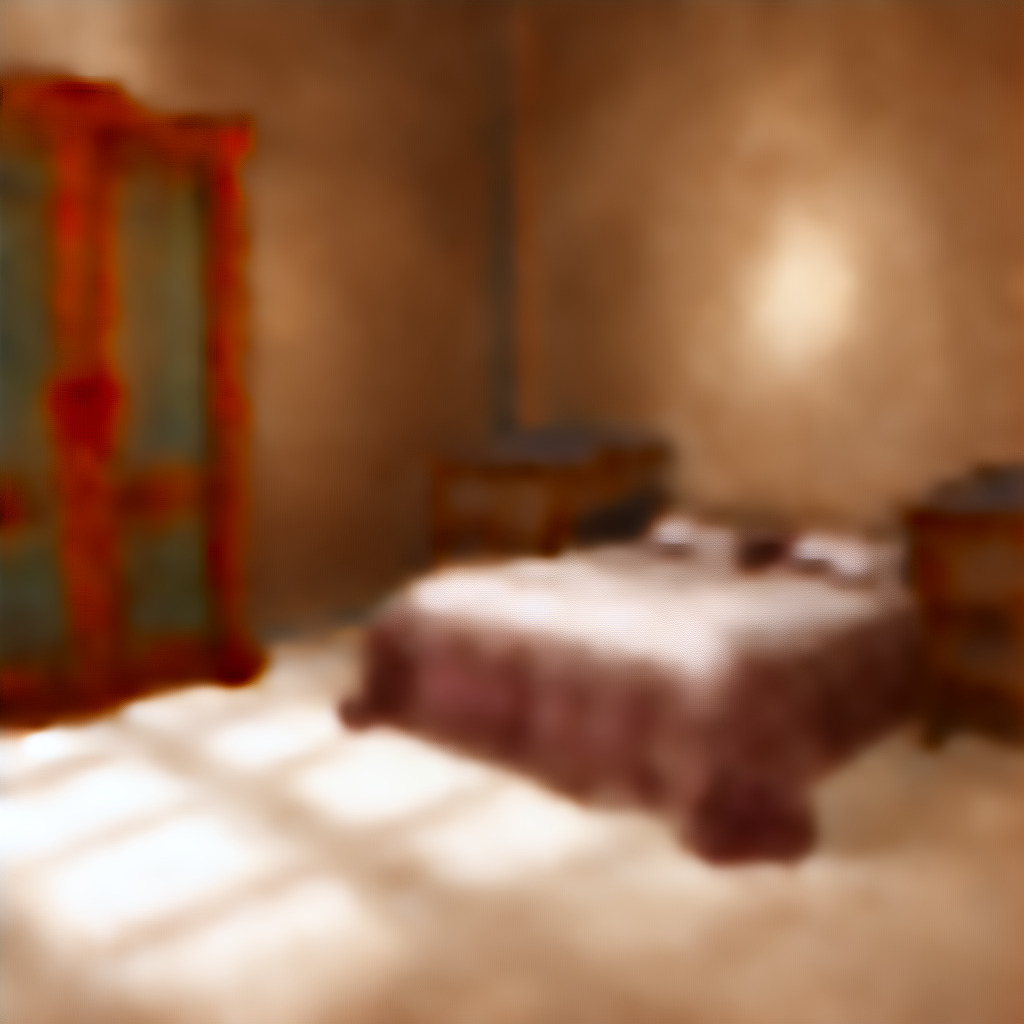} & 
        \includegraphics[height=\imheight]{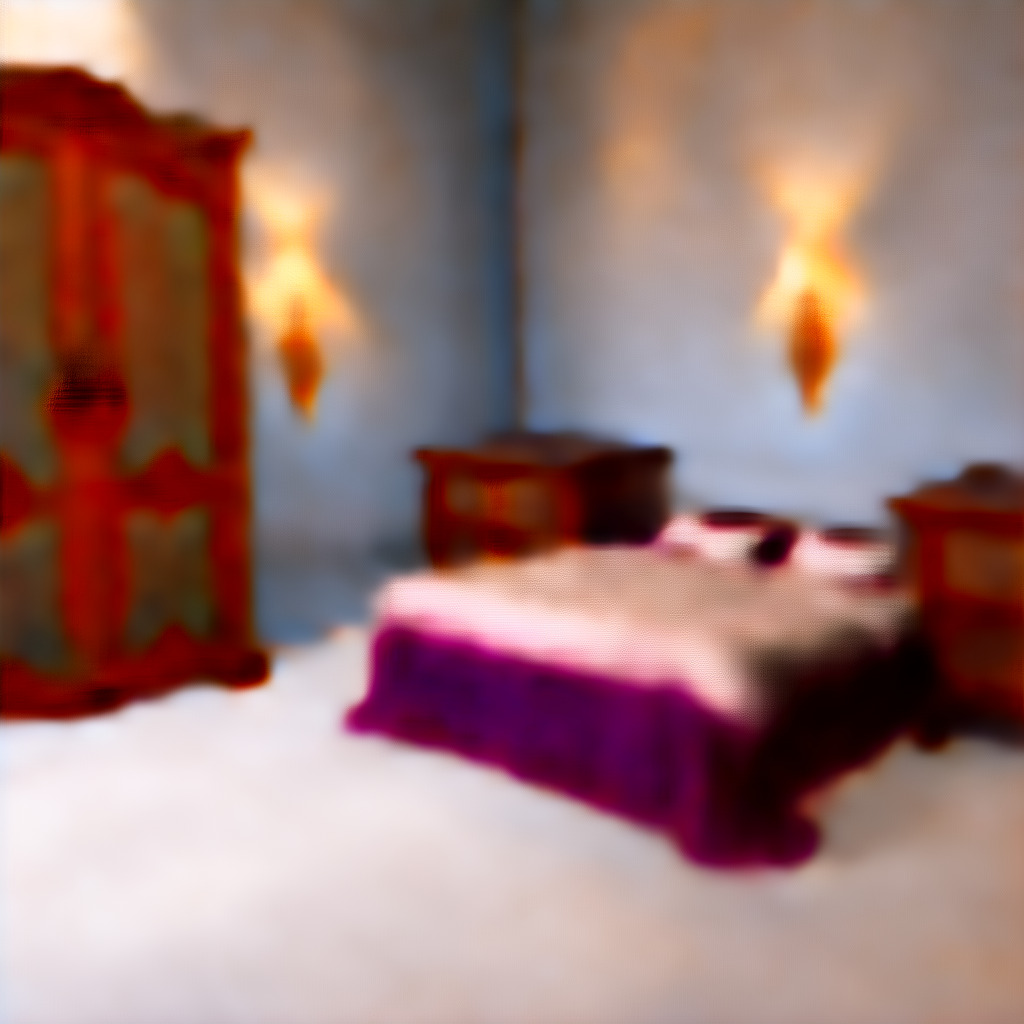} &
        \includegraphics[height=\imheight]{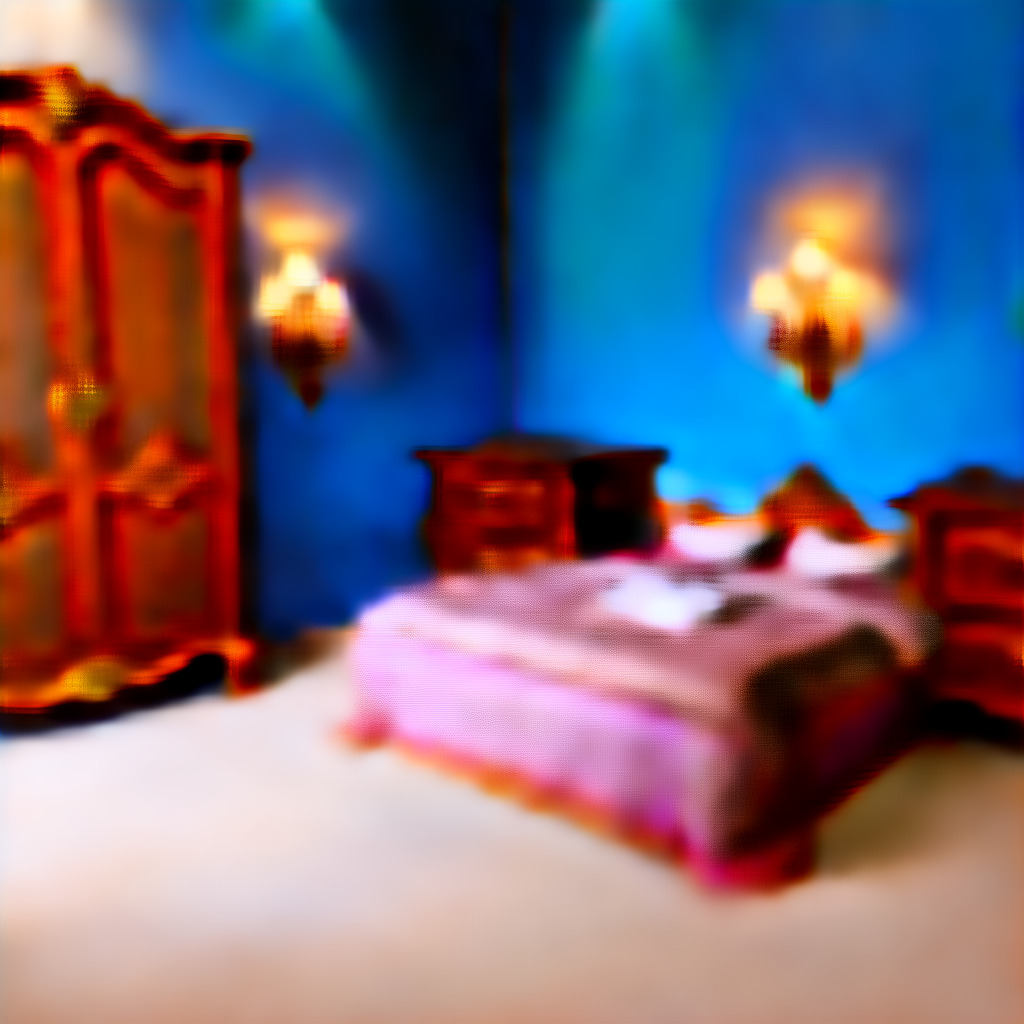} \\
        \multicolumn{5}{c}{``a baroque bedroom''} 
    \end{tabular}}
    \caption{{\bf Convergence process of Set-the-Scene} for different NeRFs composing the scene. The bottom row shows the entire scene. 
    A specific text prompt is given for each object, specified below each row of images.}
    \label{fig:convergence}
\end{figure}

\newcommand{\compimheight}{0.31\linewidth} 
\begin{figure}
    \centering
    \setlength{\tabcolsep}{1pt}
    {\small
    \begin{tabular}{c c c c }
    \raisebox{0.06\textwidth}{\rotatebox[origin=t]{90}{Latent-NeRF~\cite{metzer2022latent}}} & 
        \includegraphics[width=\compimheight]{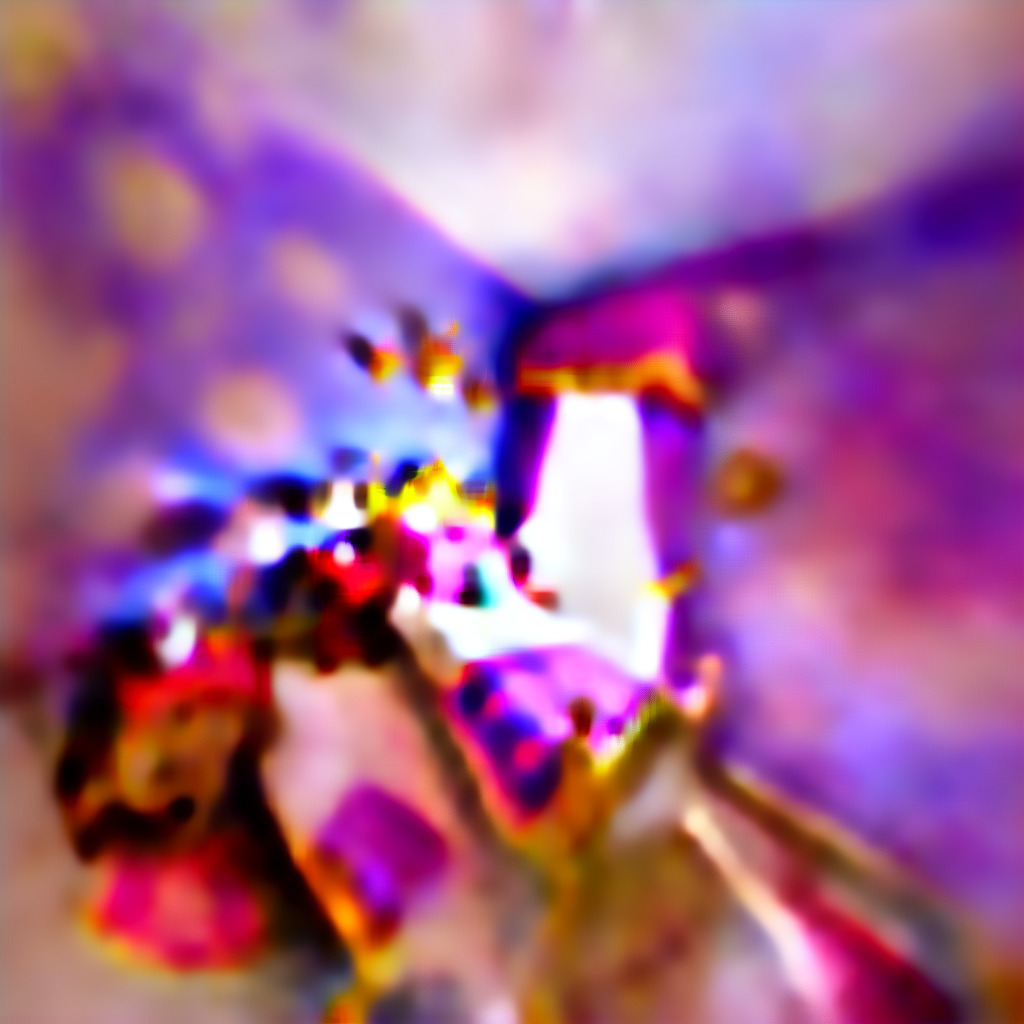} &
        \includegraphics[width=\compimheight]{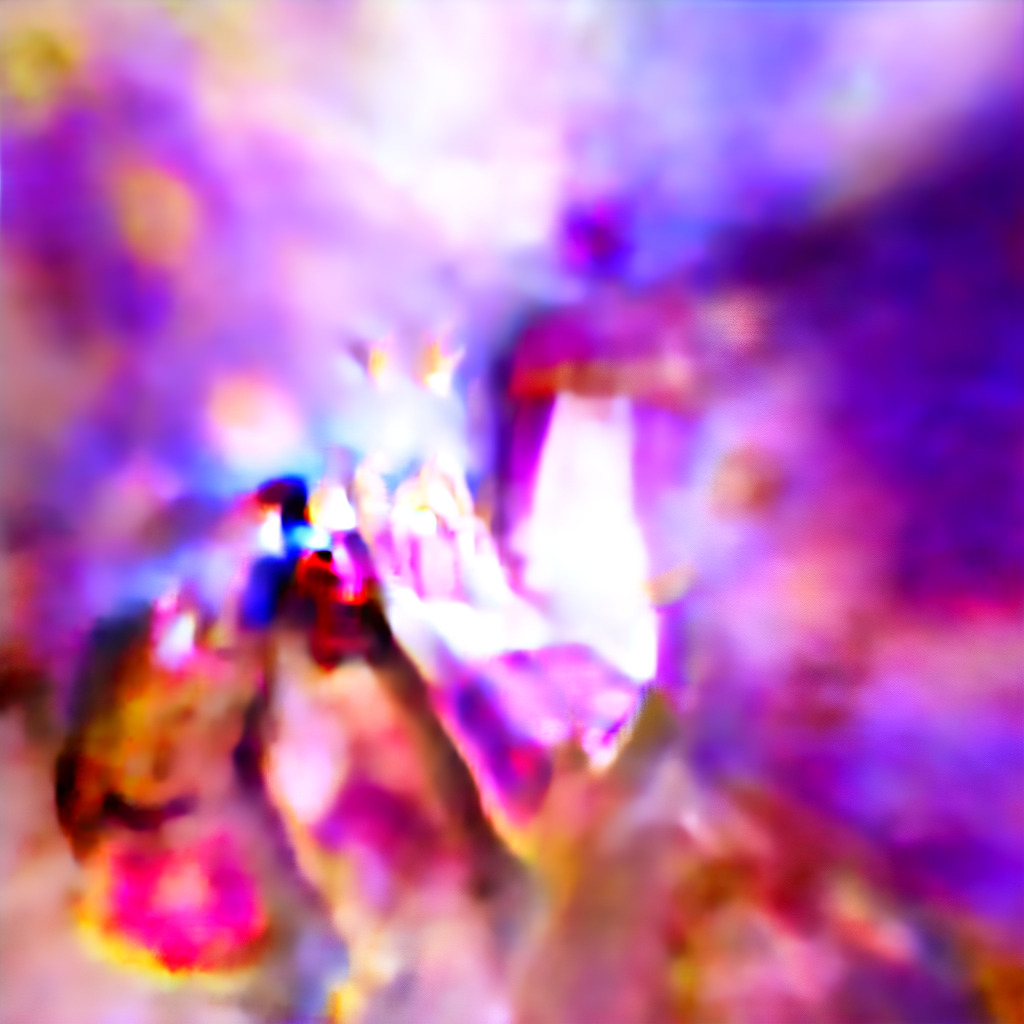} & 
        \includegraphics[width=\compimheight]{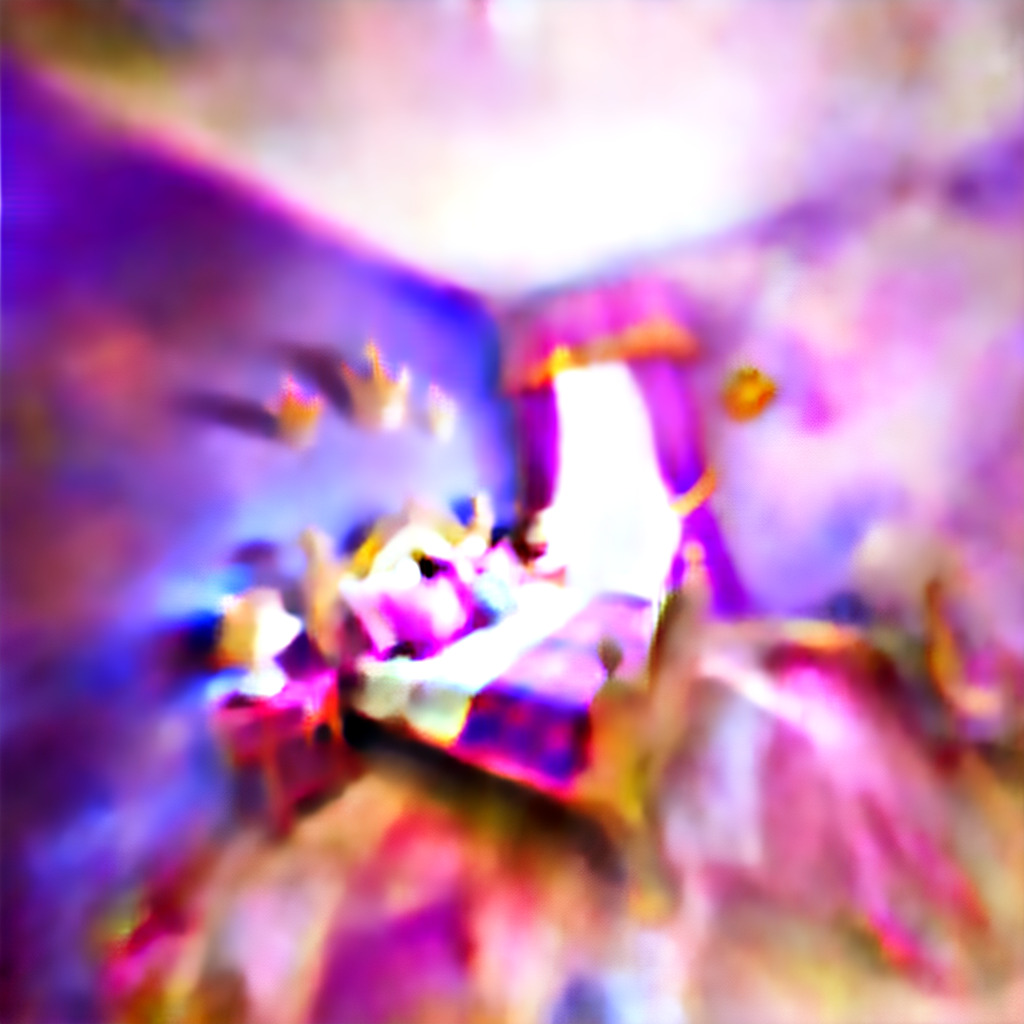} \\
        \raisebox{0.07\textwidth}{\rotatebox[origin=t]{90}{Set-the-Scene}} & 
        \includegraphics[width=\compimheight]{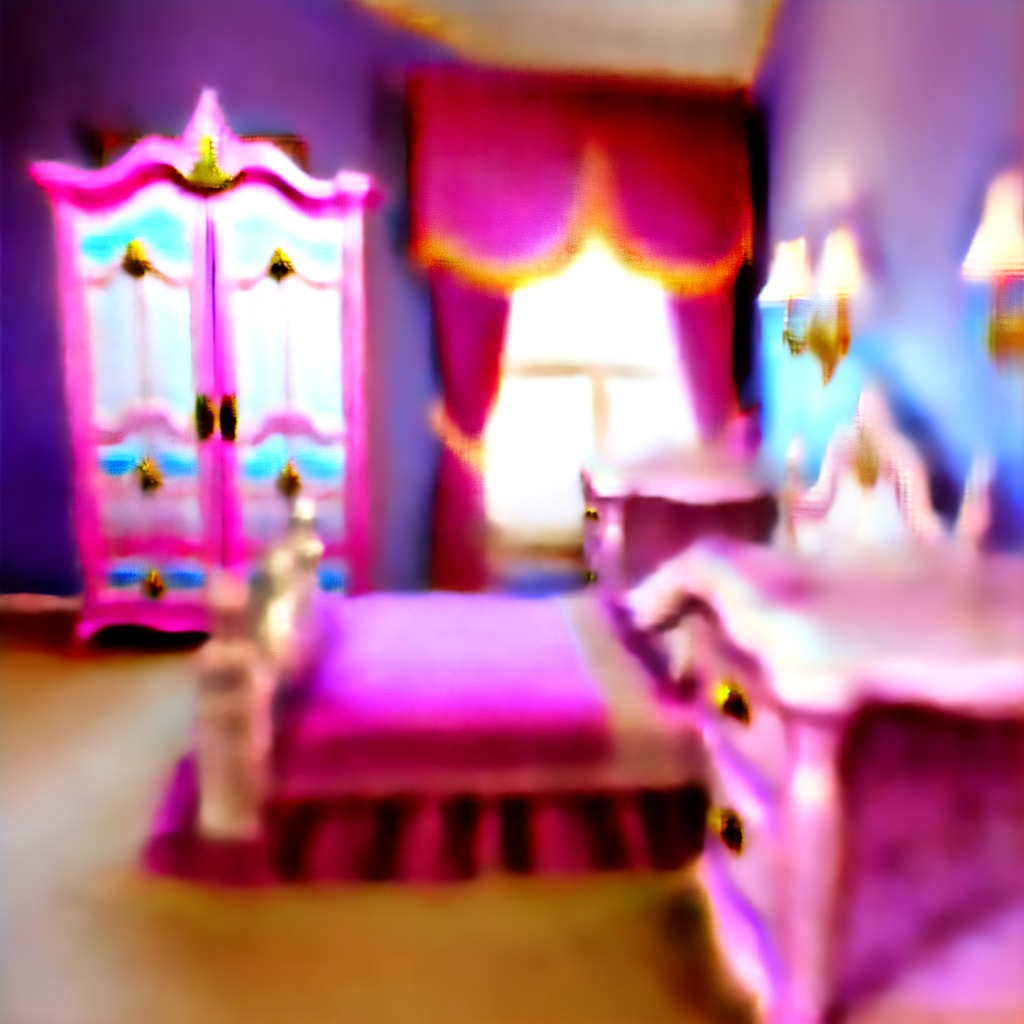} &
        \includegraphics[width=\compimheight]{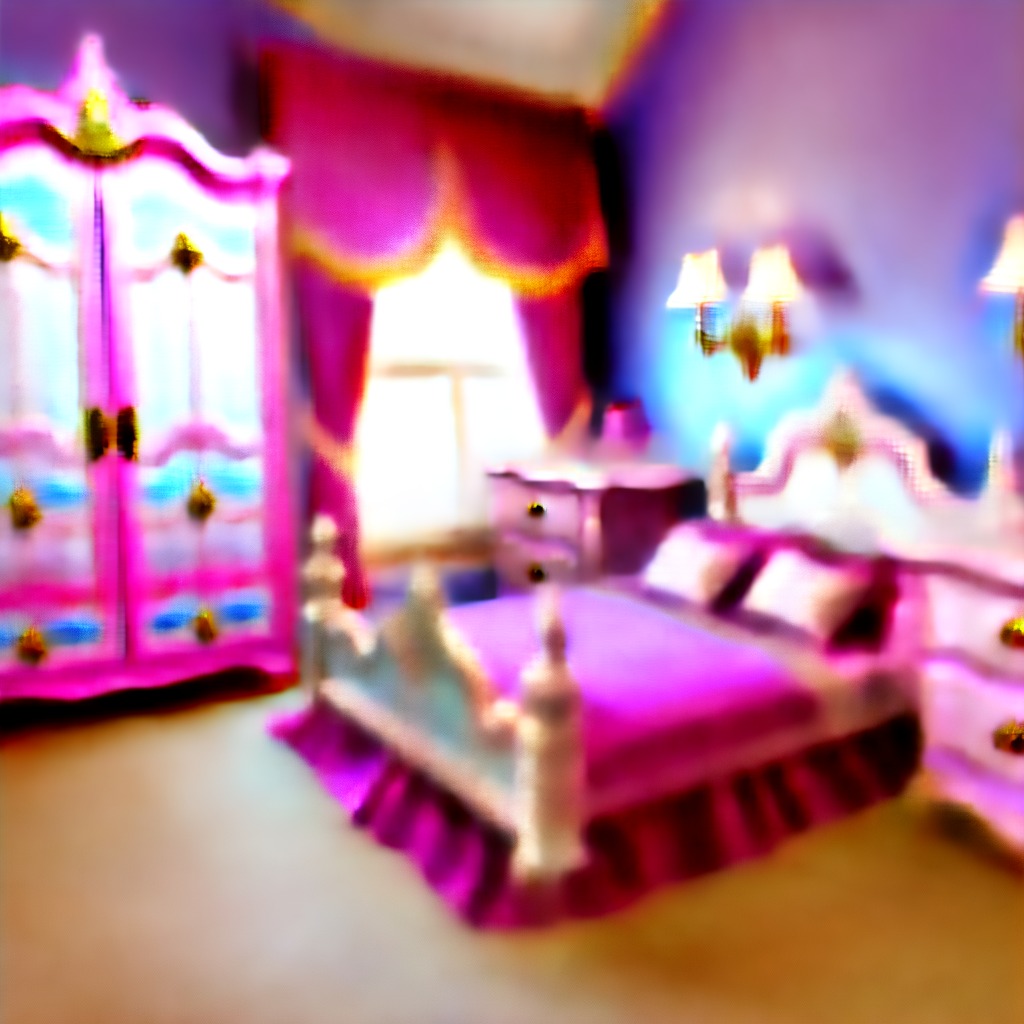} & 
        \includegraphics[width=\compimheight]{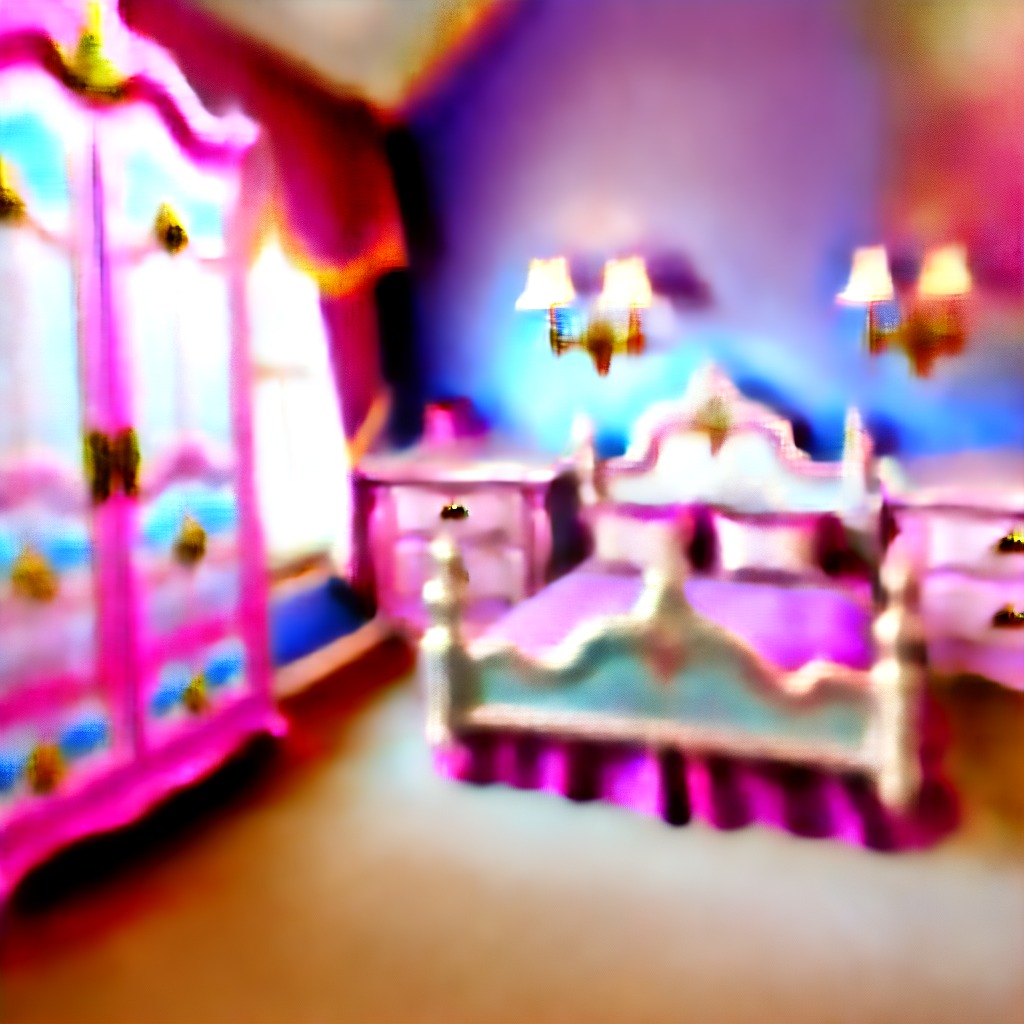} \\

    \end{tabular}}
    
    \caption{ {\bf Comparison with Latent-NeRF.} When generating ``A princess bedroom`` Latent-NeRF struggles with generating the complex scene. }
    \vspace{-0.3cm}
    \label{fig:comparison}
\end{figure} 

\begin{figure}
    \centering
    \setlength{\tabcolsep}{0pt}
    \renewcommand{\arraystretch}{1}
    {\small
    \begin{tabular}{c c c}
        \includegraphics[width=0.33\linewidth]{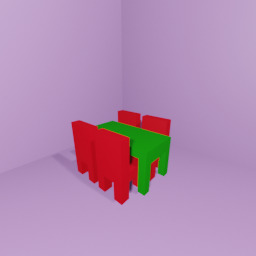} & 
        \includegraphics[width=0.33\linewidth]{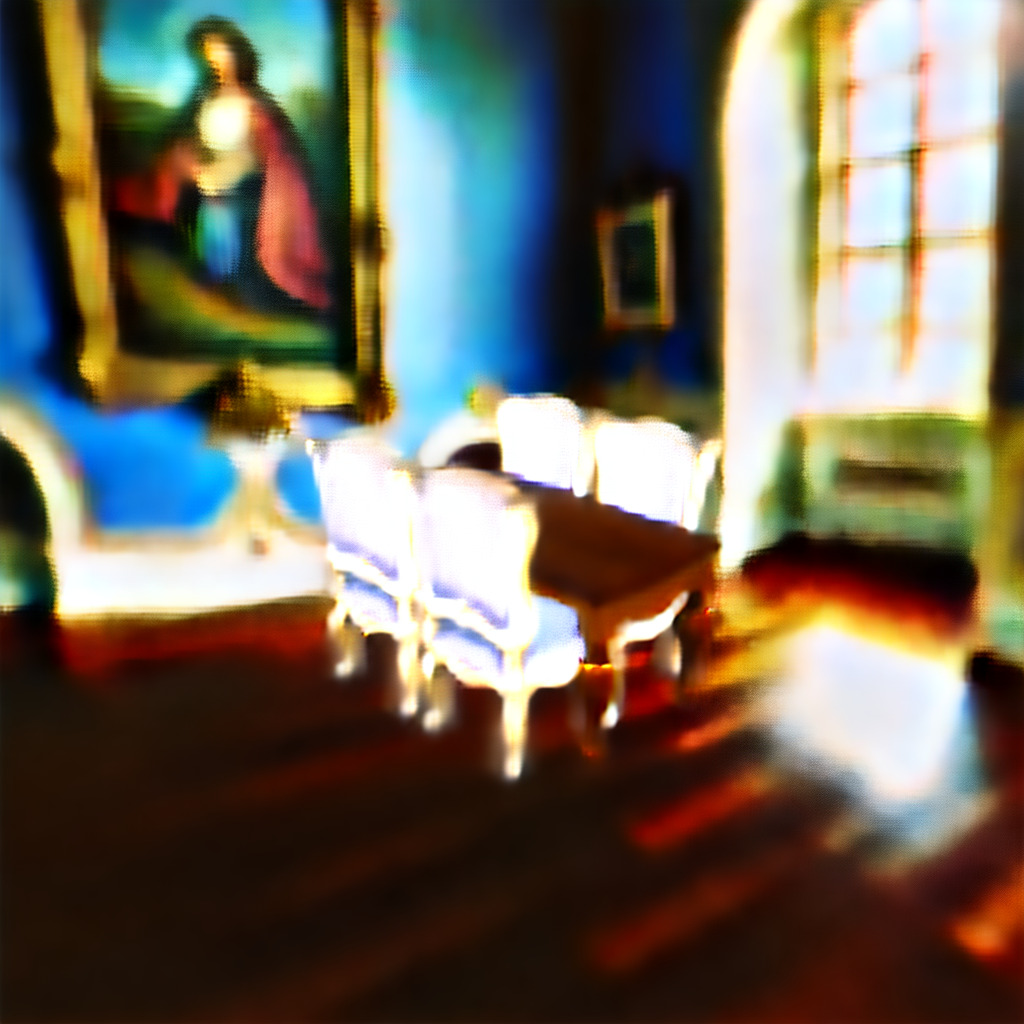} & 
        \includegraphics[width=0.33\linewidth]{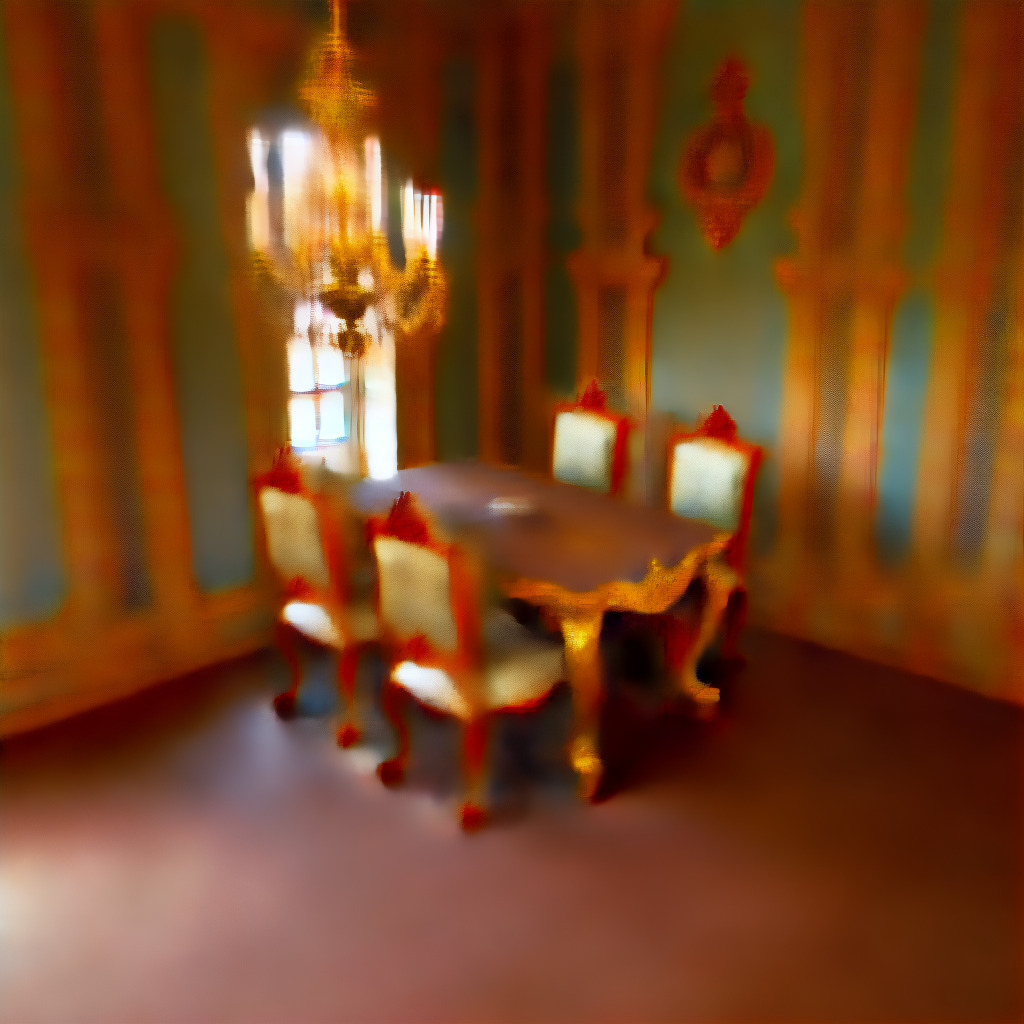} \\ 
        \includegraphics[width=0.33\linewidth]{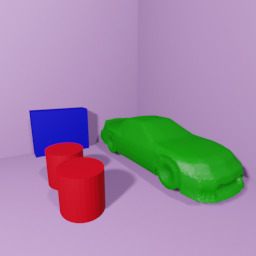} & 
        \includegraphics[width=0.33\linewidth]{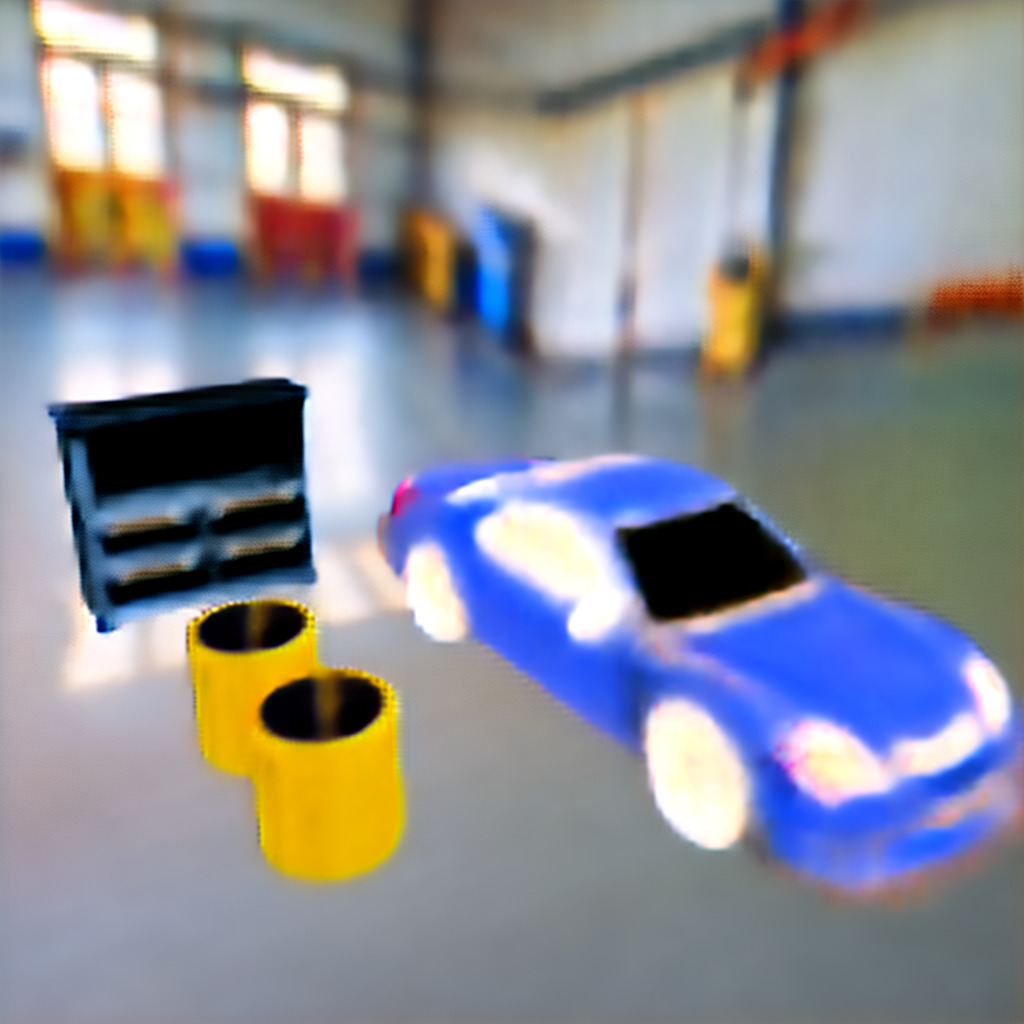} & 
        \includegraphics[width=0.33\linewidth]{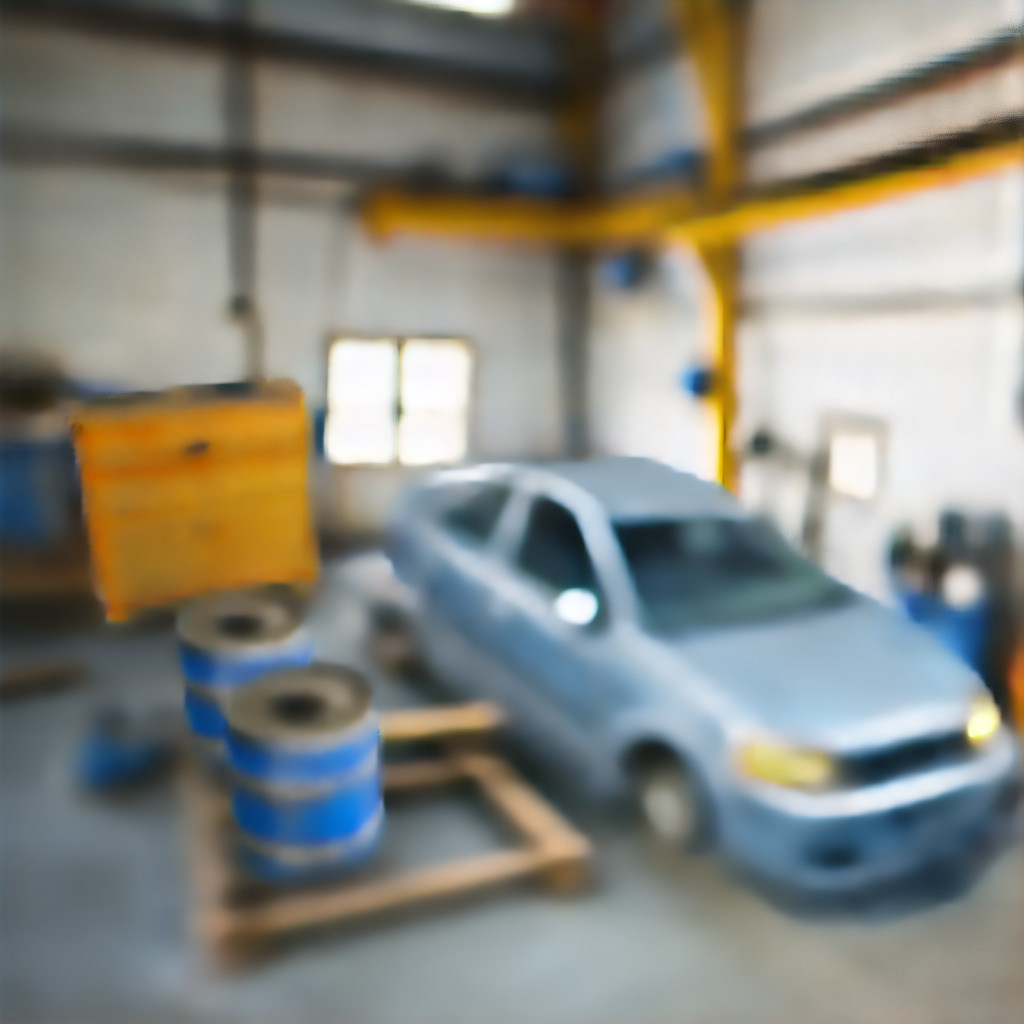}\\
        \includegraphics[width=0.33\linewidth]{figures/ablation/living_room_proxy.jpg} & 
        \includegraphics[width=0.33\linewidth]{figures/ablation/living_no_scene.jpg} & 
        \includegraphics[width=0.33\linewidth]{figures/ablation/living_with_scene.jpg}  \\
        & Local & Global-Local \\
        & Training & Training

    \end{tabular}}
    \caption{{\bf Impact of Global-Local training.}
    Here we show the importance of global training for three different scenes. Observe how our Global-Local training policy results in more coherent scenes, compared to scenes where each object is optimized independently that are inconsistent.} 
    \label{fig:ablation}
\end{figure}

\vspace{-0.3cm}
\paragraph{Ablation Study} Next, we perform an ablation study over our proposed Global-Local training scheme.
Figure~\ref{fig:ablation} shows results for the same proxy configuration with and without the global optimization phase, which makes several important aspects of joint optimization noticeable.
First, observe how color schemes between objects do not match well without joint optimization steps, for example, the sofa set which is generated in two different colors.
This supports the claim that without global optimization steps the model will not be able to match the objects well.
Moreover, the results generated without applying global optimization steps do not blend well within the scene and tend to look like they have been pasted over, which is expected for objects that were trained separately.
By contrast, our method generates globally consistent scenes with realistic shadows generated on the background object, resulting in more natural renderings.

\vspace{-0.3cm}
\paragraph{User Study} We conducted a user study to quantitatively assess the effectiveness of our training scheme. We chose 10 scenes and applied both our Global-Local and Local-Only training schemes. We then gave 40 participants two sets of tasks. In the ranking task, participants are presented with two results side-by-side (in random order) and are asked to choose their preferred result with respect to two aspects: (a) realism; and (b) compatibility between the objects in the scene. In the second set, participants are presented with only a single result and are asked to rank it on a scale of 1-5 with respect to (a) realism; (b) object compatibility; and (c) text fidelity.
All results are shown as short videos showing the moving scene to allow participants to better evaluate the generated results. Table~\ref{tb:user_study_rank} shows the outcome of our study. One can see that using our Global-Local method is superior to a Local-Only solution and results in generally higher scores for the generated images.
\begin{table}
\small
\centering
\setlength{\tabcolsep}{2pt}
\begin{tabular}{l c c} 
\toprule
&  \begin{tabular}{c} Local Training \end{tabular} & \begin{tabular}{c} Global-Local Training \end{tabular}  \\
\midrule
Realism Rank ($\uparrow$) & $19.2\%$ & $\mathbf{80.8\%}$ \\
Compatibility Rank ($\uparrow$) & $20.5\%$ & $\mathbf{79.5\%}$ \\
\midrule
Realism Score ($\uparrow$) & $1.92\pm0.28$ & $\mathbf{3.48\pm0.36}$ \\
Compatibility Score ($\uparrow$) & $2.91\pm0.40$ & $\mathbf{4.2\pm0.33}$ \\
Text Fidelity Score ($\uparrow$) & $2.69\pm0.34$ & $\mathbf{3.94\pm0.35}$ \\
\bottomrule
\end{tabular}
\caption{
{\bf User study results.} Each respondent is asked to rank both Local-Only and Global-Local training with respect to various aspects.
} 
\vspace{-0.2cm}
\label{tb:user_study_rank}
\end{table}

\begin{figure}[b]
    \centering
    \setlength{\tabcolsep}{0pt}
    {\small
    \begin{tabular}{c c c c}
        \multicolumn{2}{c}{Training setting} & 
        \multicolumn{2}{c}{Post-Training editing} \\
        \includegraphics[width=0.25\linewidth]{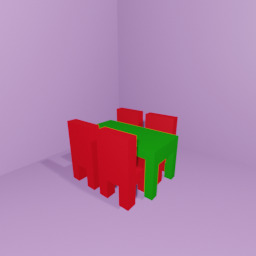} & 
        \includegraphics[width=0.25\linewidth]{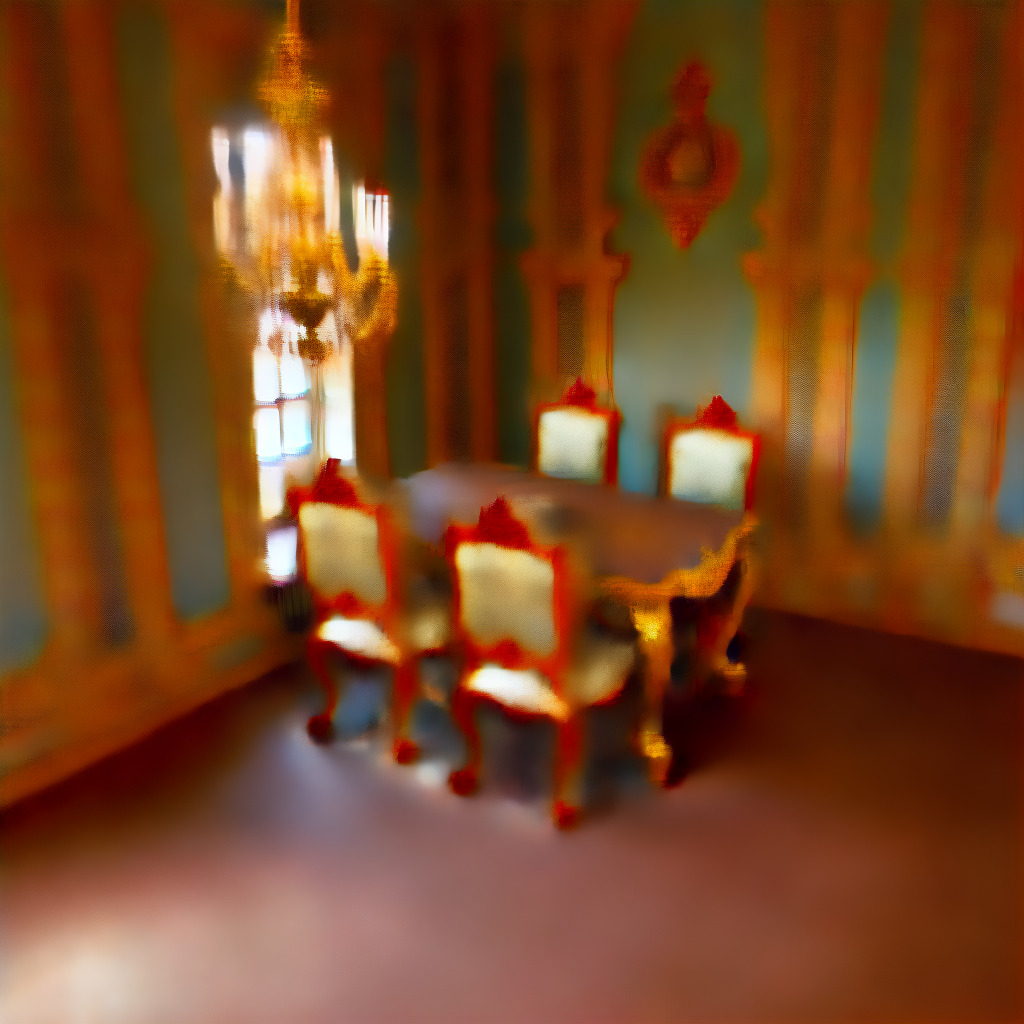} & 
        \includegraphics[width=0.25\linewidth]{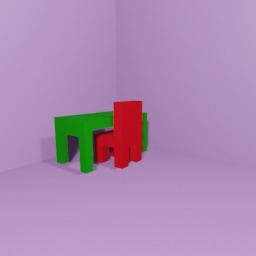} &
        \includegraphics[width=0.25\linewidth]{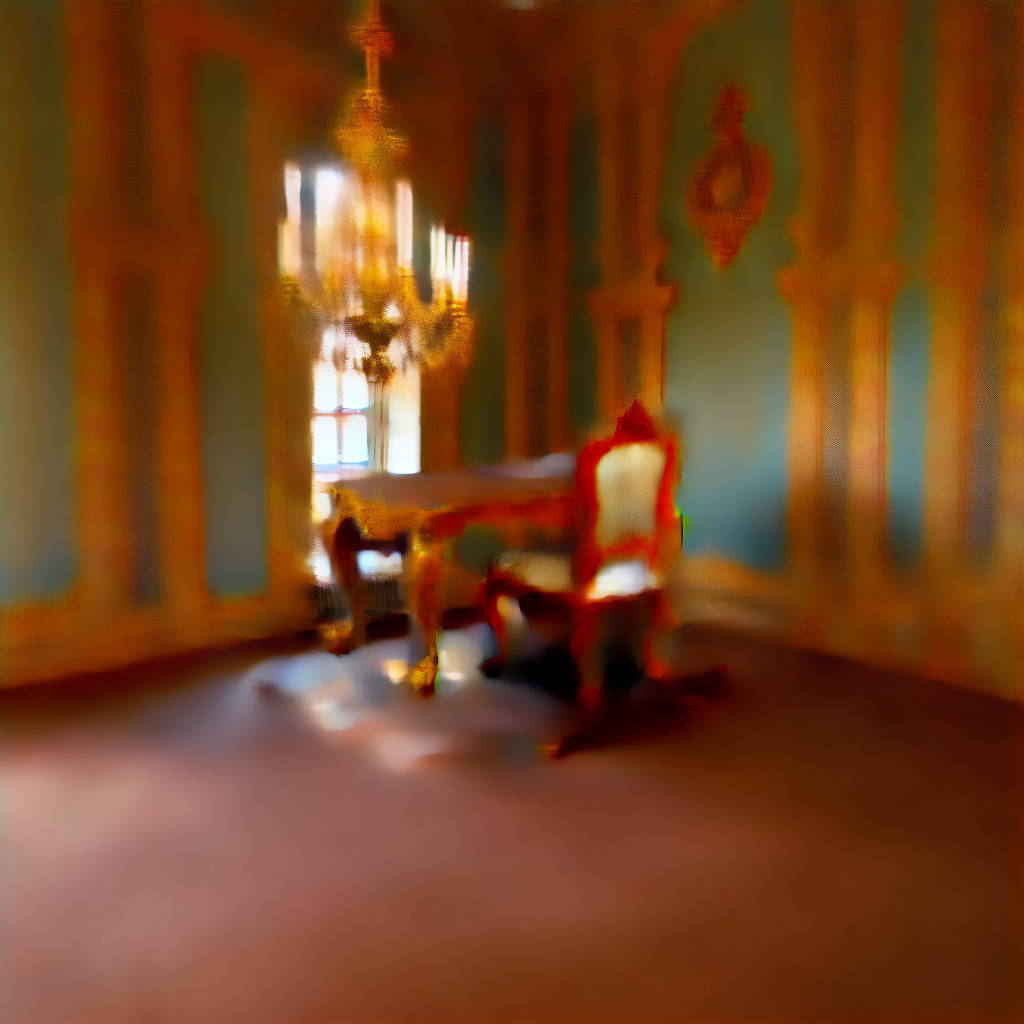}\\
        \multicolumn{4}{c}{``a Baroque dining room''} \\
        \includegraphics[width=0.25\linewidth]{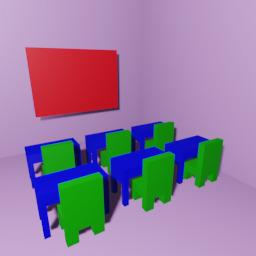} & 
        \includegraphics[width=0.25\linewidth]{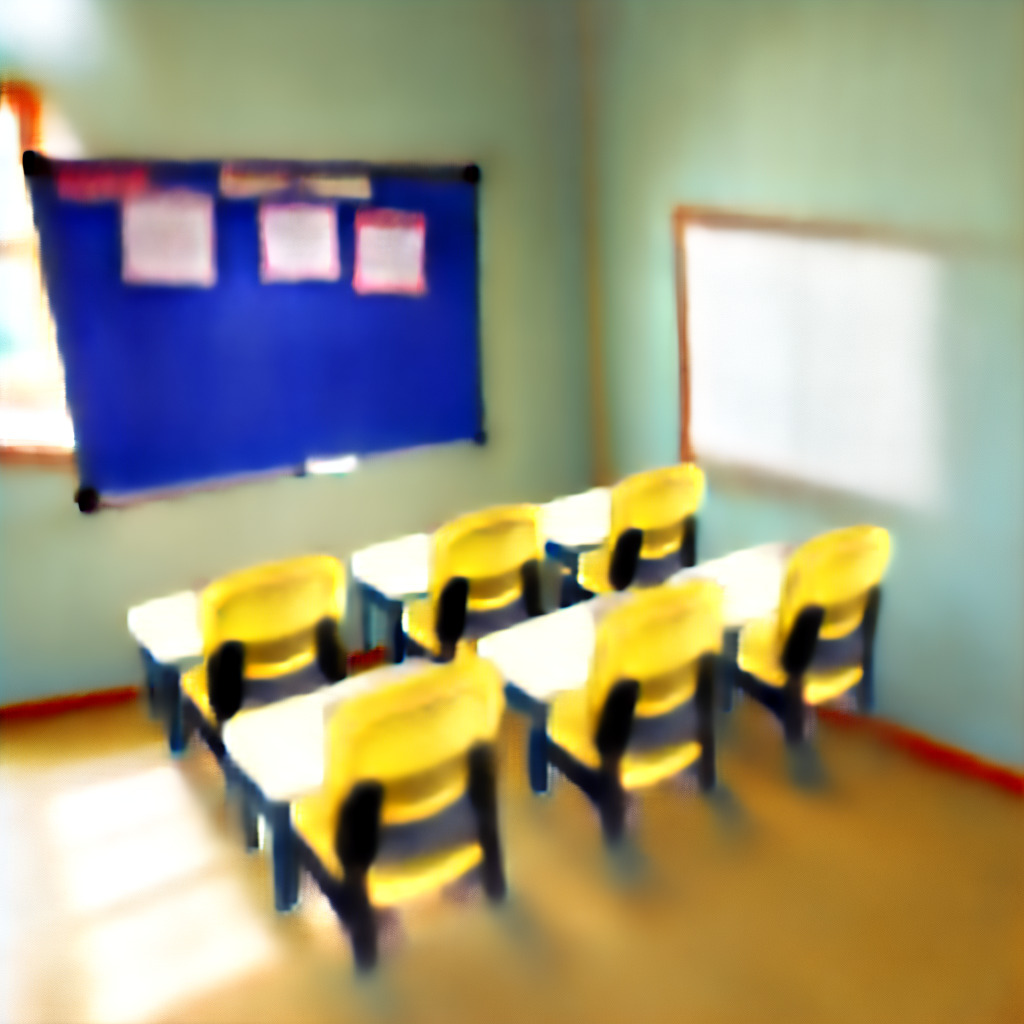} & 
        \includegraphics[width=0.25\linewidth]{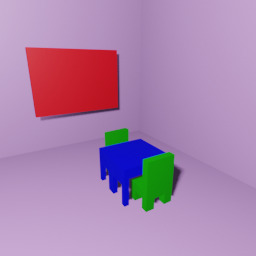} & 
        \includegraphics[width=0.25\linewidth]{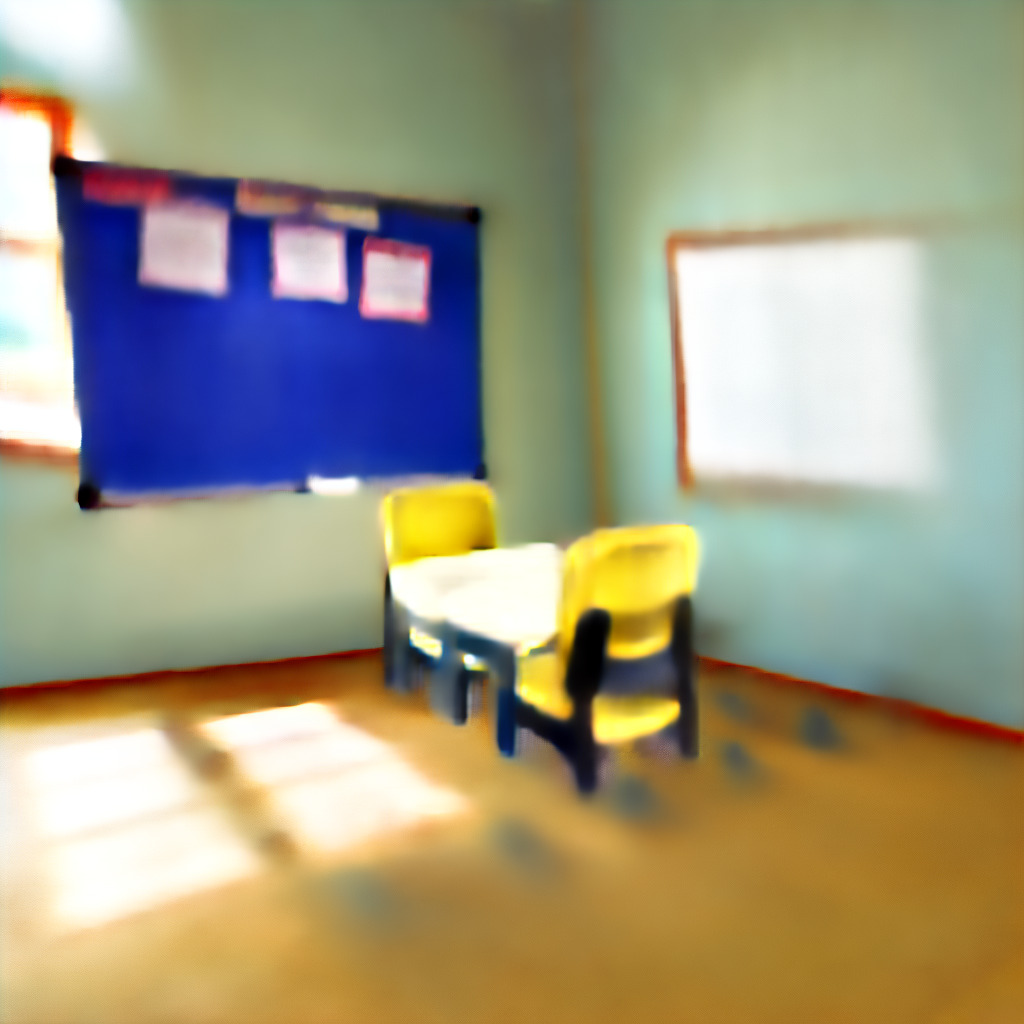} \\
        \multicolumn{4}{c}{``A classroom''} \\[0.1pt]

    \end{tabular}}
    \caption{{\bf Location Editing.}
    Here we show how the composable nature of our representation allows removing  or moving objects around. Given an optimized scene, we can simply move or remove each proxy, and still produce high-fidelity results of the same scene.}
    \label{fig:edit_placment}
\end{figure} 

\subsection{Scene Editing}
Having shown our scene generation capabilities, we now turn to evaluate our ability to edit already-generated shapes.
\paragraph{Placement Editing} Figure~\ref{fig:edit_placment} shows some examples of placement editing.
Due to our composable representation, Set-the-Scene inherently allows for editing of object placement after training by editing the object proxies, adding new proxies, or even removing existing ones. Note that this can be done without any additional fine-tuning steps. One can additionally apply some fine-tuning steps to further improve the result with respect to the new placement.

\begin{figure}
    \centering
    \setlength{\tabcolsep}{0pt}
    {\small
    \begin{tabular}{c c c c}
        \multicolumn{2}{c}{Training setting} & 
        \multicolumn{2}{c}{Post-Training editing} \\
        \includegraphics[width=0.25\linewidth]{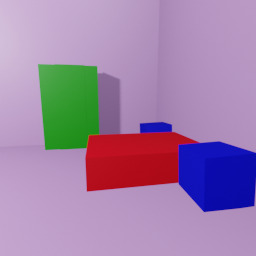} & 
        \includegraphics[width=0.25\linewidth]{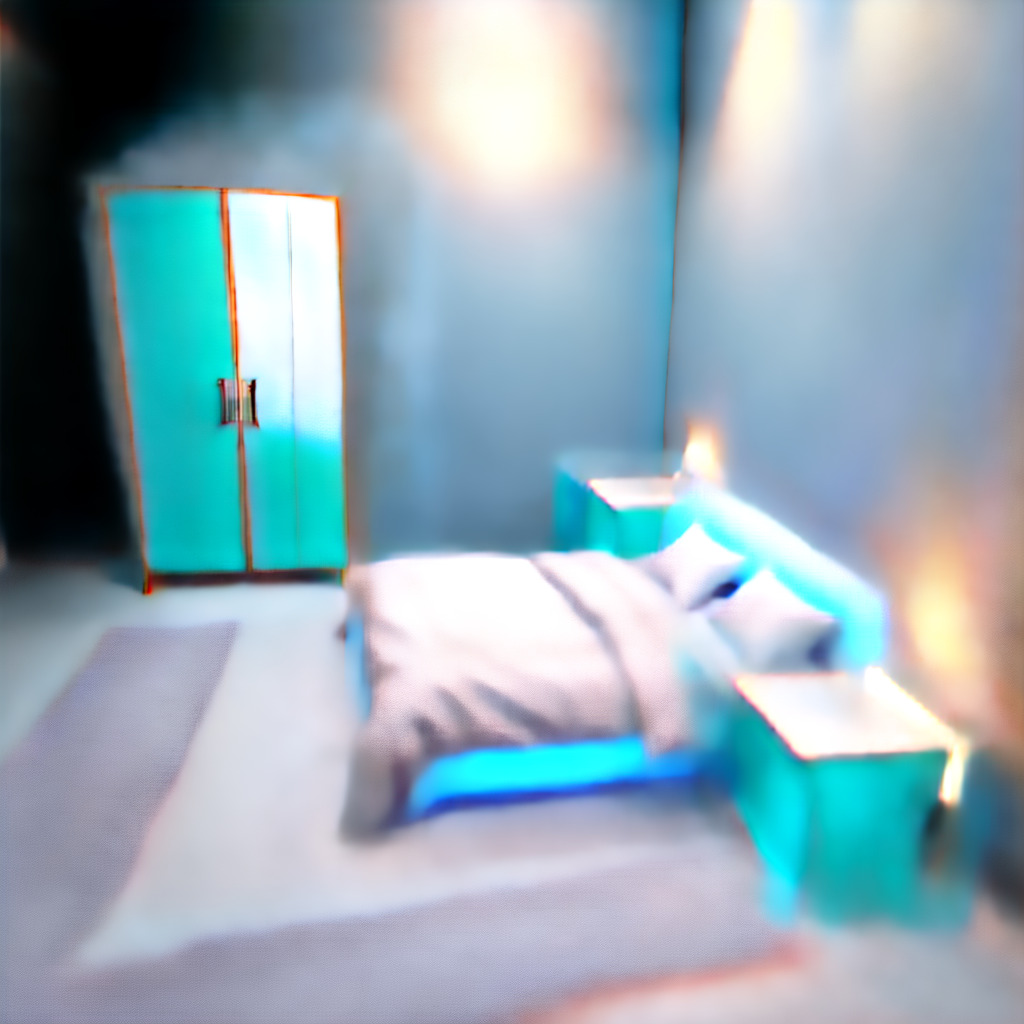} & 
        \includegraphics[width=0.25\linewidth]{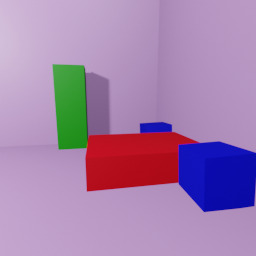} & 
        \includegraphics[width=0.25\linewidth]{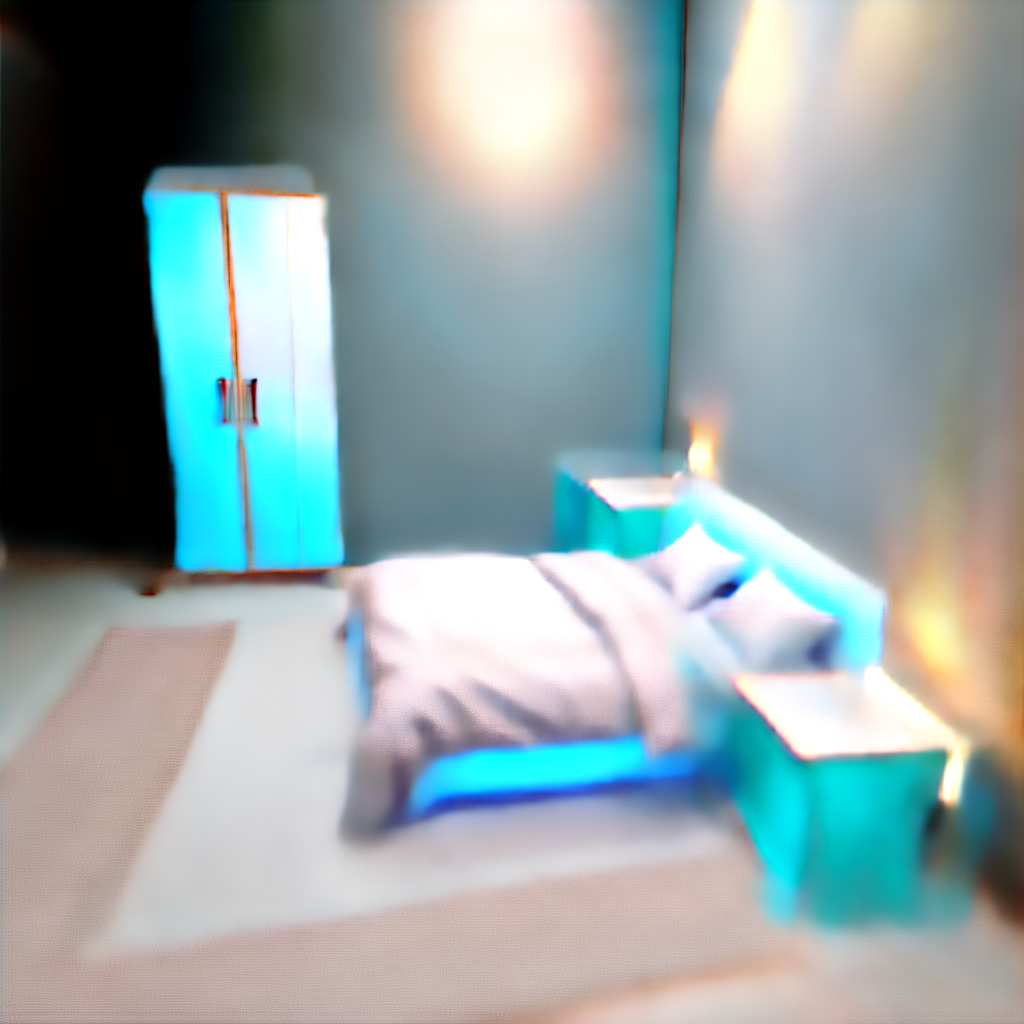} \\
 \\[0.1pt]
    \end{tabular}}
    \caption{{\bf Geometric Editing}. 
    (left) Input scene, (right) Edited scene.
    Observe how narrowing the closet proxy shapes results in a similar effect on the generated closet, while keeping the rest of the scene content intact. }
    \label{fig:edit_geometry}
\end{figure}

\paragraph{Geometry Editing} Figure~\ref{fig:edit_geometry} shows how we can edit the geometry of a specific NeRF within the scene by changing its corresponding object, which allows for easy and intuitive control over the implicit NeRF.
As the proxy geometry is based on a user-defined mesh, the corresponding mesh can be easily edited with existing tools such as MeshLab~\cite{meshlab} or Blender~\cite{blender}. One can see that after fine-tuning the edited result remains faithful to the original scene.

\begin{figure}
    \centering
    \setlength{\tabcolsep}{0pt}
    {\small
    \begin{tabular}{c@{\hskip 0.1cm}c c c}
    \raisebox{0.07\textwidth}{\rotatebox[origin=t]{90}{Global-Local}} & 
        \includegraphics[width=0.31\linewidth]{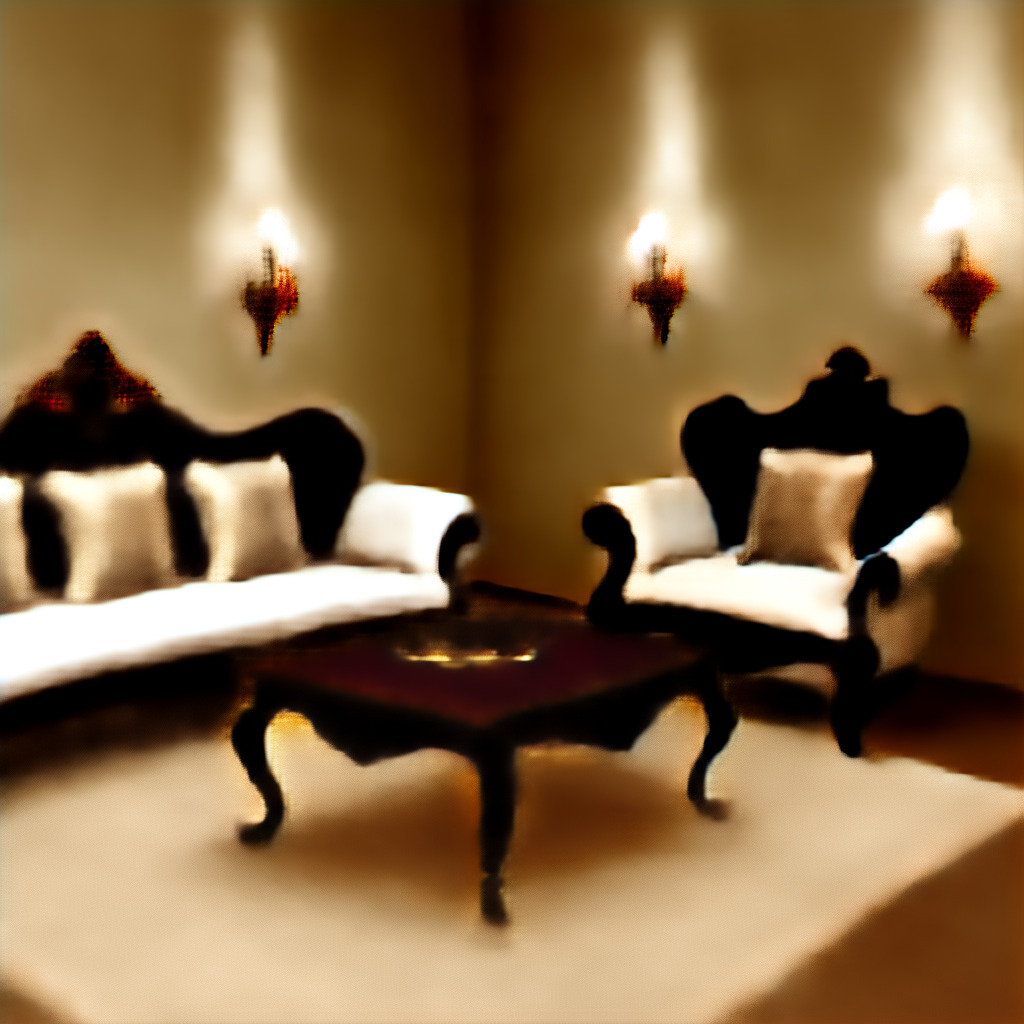} & 
        \includegraphics[width=0.31\linewidth]{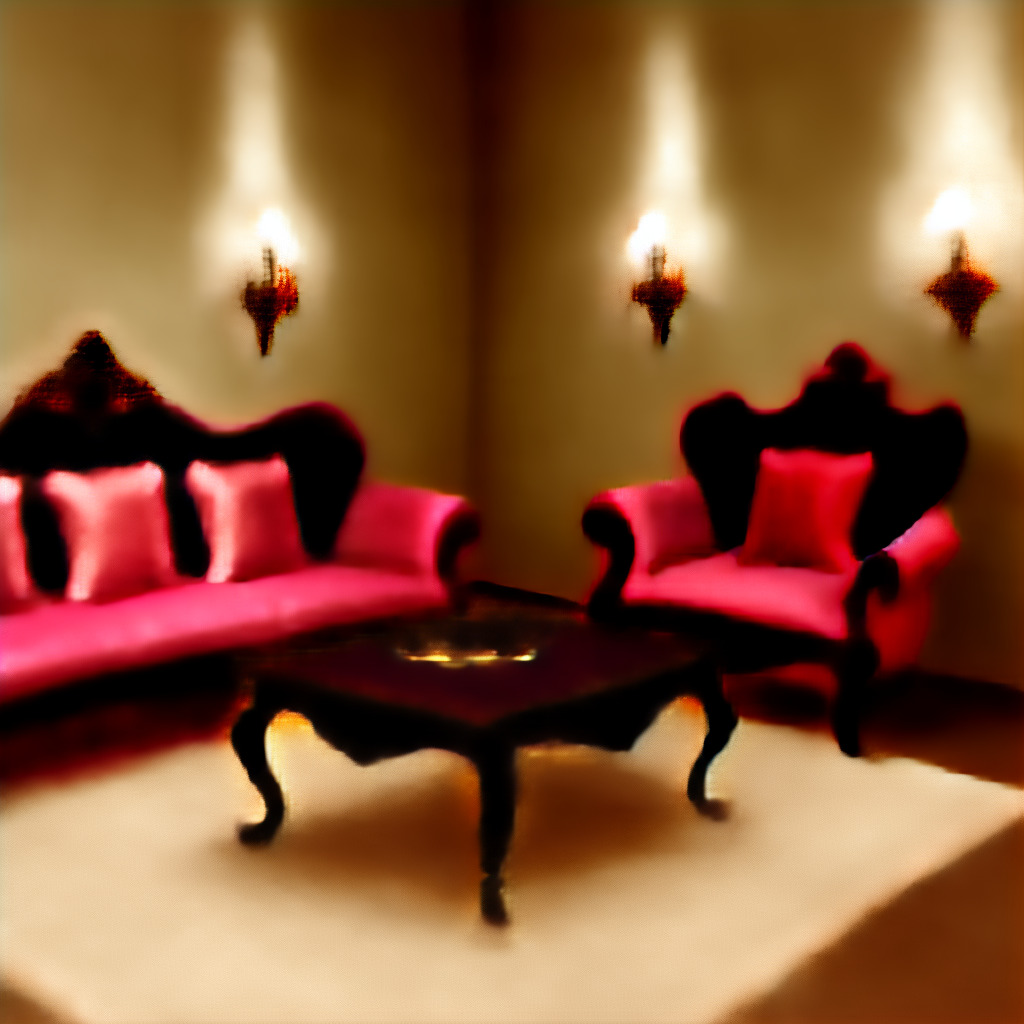} & 
        \includegraphics[width=0.31\linewidth]{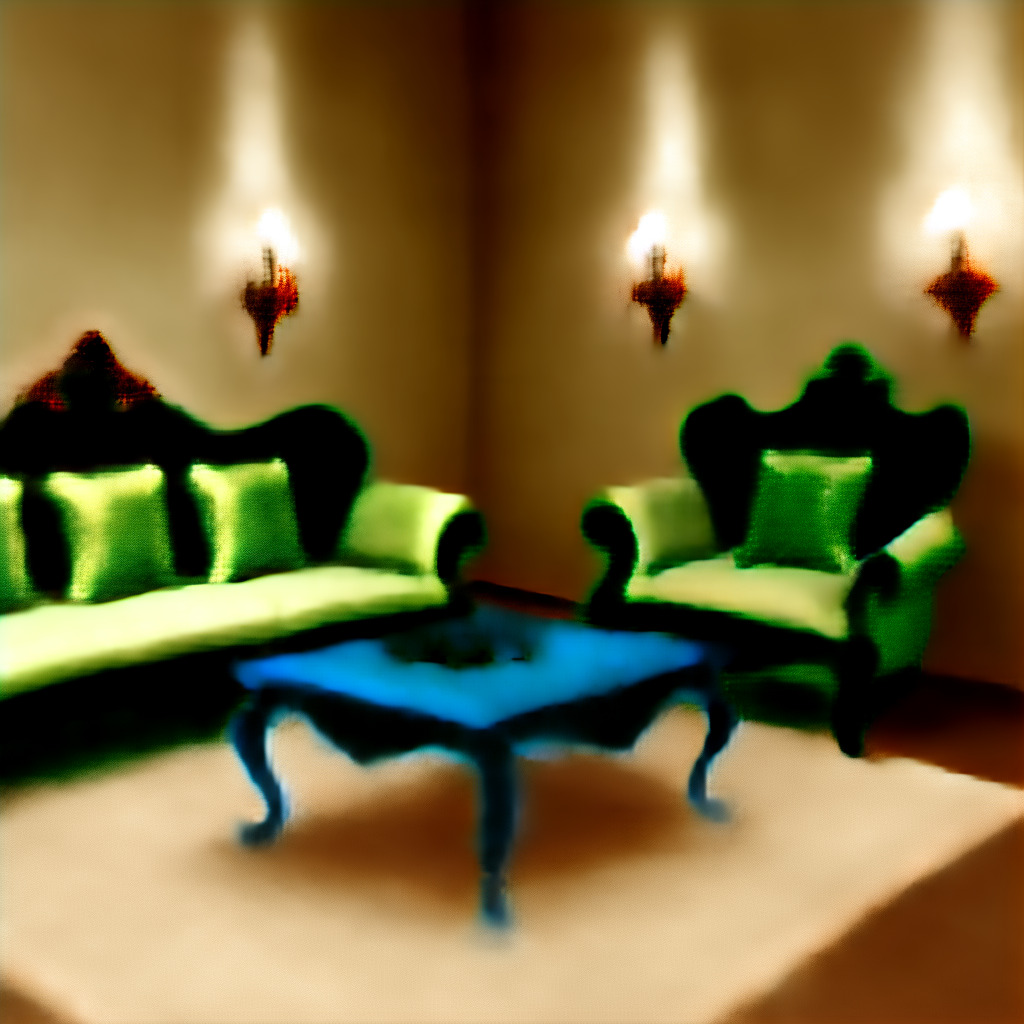} \\ 
        \raisebox{0.07\textwidth}{\rotatebox[origin=t]{90}{Local Only}} & 
        \includegraphics[width=0.31\linewidth]{figures/color/images/living_source.jpg} &
        \includegraphics[width=0.31\linewidth]{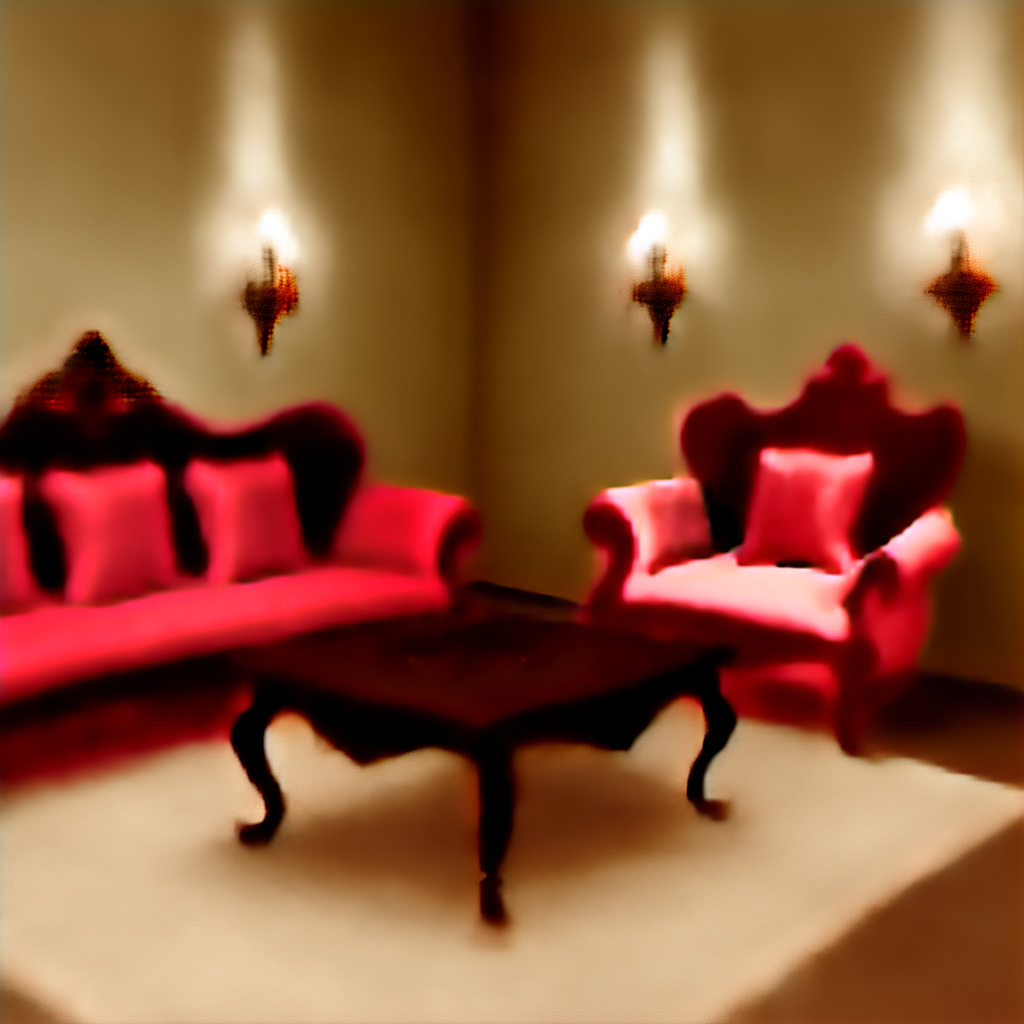} & 
        \includegraphics[width=0.31\linewidth]{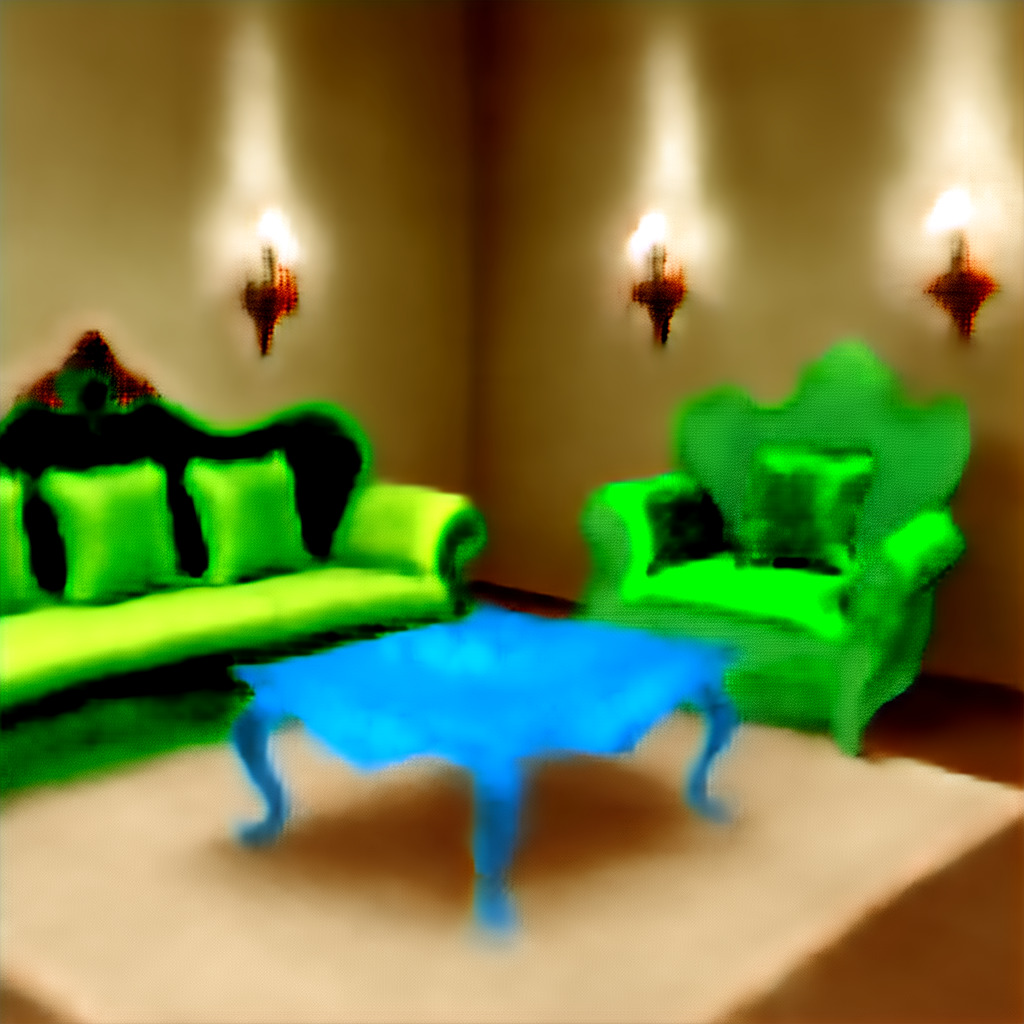} \\ 
        & Input Scene &  Edited Sofas  &  Edited Sofas+Table \\

    \end{tabular}}
    \caption{{\bf Color editing.} Using a dedicated albedo head, we can fine-tune the NeRFs to change the color scheme while keeping the geometry and other objects unchanged. For each edited object the object text prompt was changed according to the desired color, for example the prompt ``a baroque sofa`' was changed to '``a red baroque sofa''.}
    \label{fig:edit_color}
\end{figure}

\paragraph{Appearance Editing} Finally, Figure~\ref{fig:edit_color} demonstrates that our method can easily change the appearance of specified objects in the scene independently of other objects or the geometry of the edited object.
Our method is able to change the color scheme of both sofas while keeping them well-matched with one another. This is not the case when optimizing without global iterations, as also shown in the same Figure. Observe for example how the exact shade of red does not much well when using only local iterations.
\begin{figure}
    \centering
    \setlength{\tabcolsep}{1pt}
    {\small
    \begin{tabular}{c c c c}
        \includegraphics[height=0.24\columnwidth]{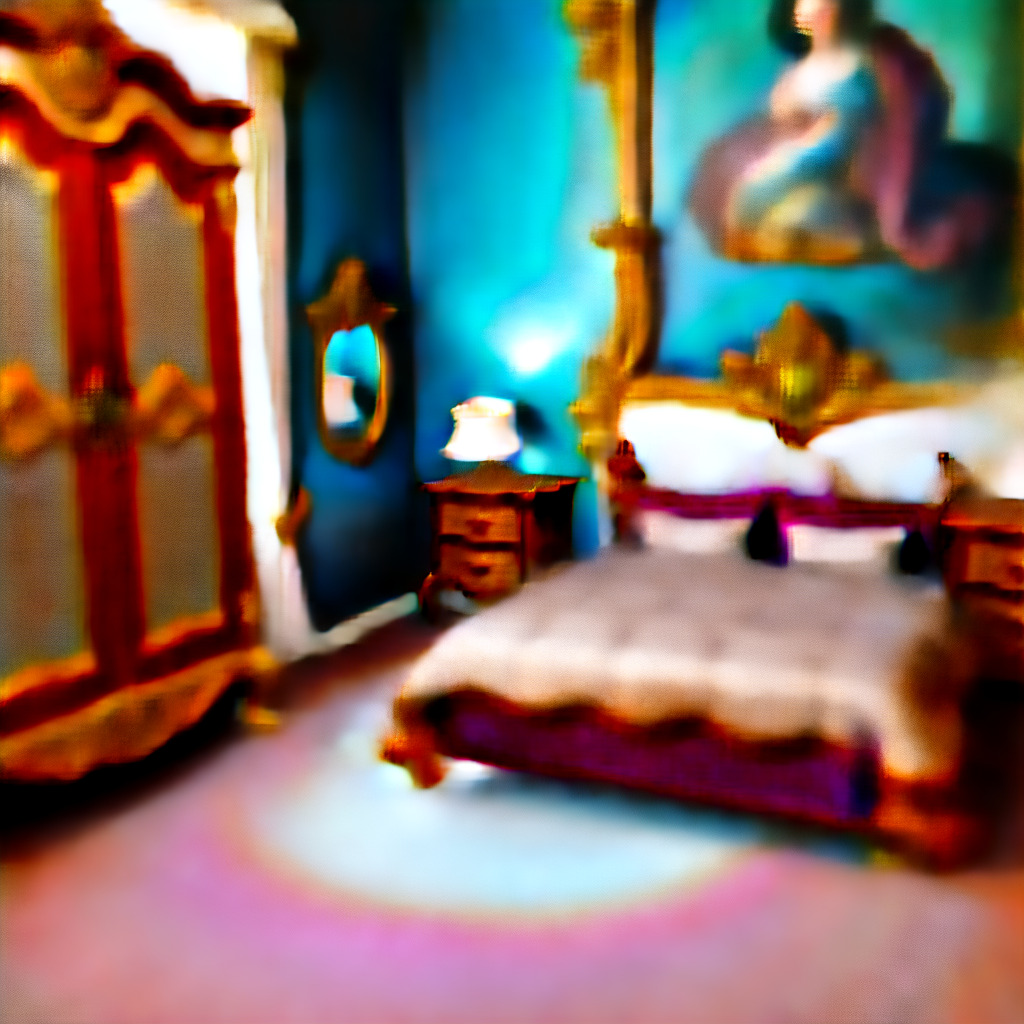} &
        \includegraphics[height=0.24\columnwidth]{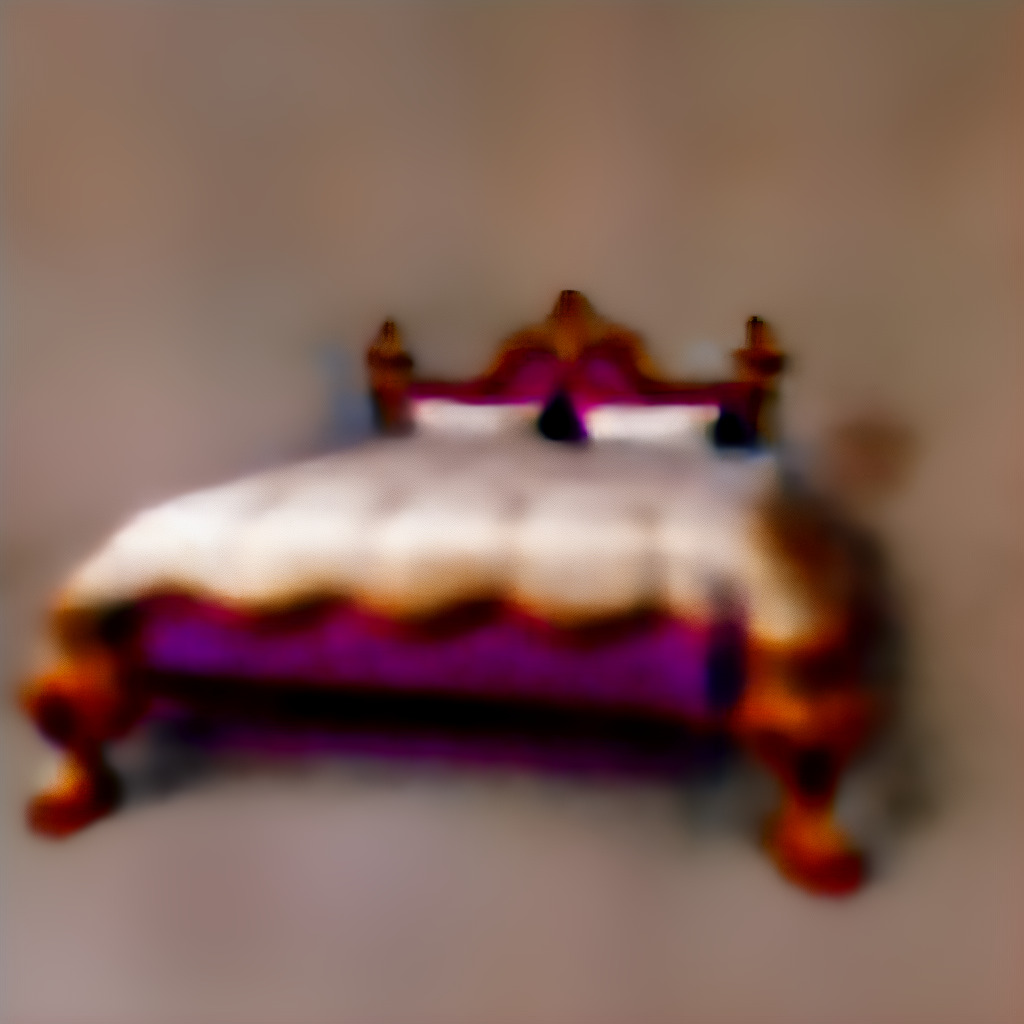} & 
        \includegraphics[height=0.24\columnwidth]{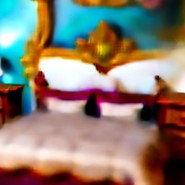} &
        \includegraphics[height=0.24\columnwidth]{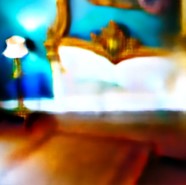} \\
        Generated Room & Generated Bed & Room close up & Walls close up

    \end{tabular}}
    \caption{{\bf Limitations.} In the scene view, the golden headboard is seen as a part of the bed. In the object view, the bed is seen as a complete object without it. }
    \label{fig:limitation}
\end{figure}

\paragraph{Limitations.} While our experiments show the capabilities of our approach, it is worthwhile to note that there are still some limitations. 
First of all, the quality of our results is governed by the local generation process which is based on score distillation with a latent diffusion model. This approach still generally lacks in terms of quality and resolution, and in turn, limits our generation's capabilities. 
Another limitation is that the objects in the scene may be generated in the background NeRF as textures, without a corresponding geometry, as shown in Figure~\ref{fig:limitation}.
We believe that this occurs due to the limited viewing angles available when optimizing an indoor scene.
Finally, our convergence time is also governed by the number of unique NeRFs in the scene, where adding more objects requires more optimization time, as each object requires its own local iterations.

\section{Conclusion}
In this paper we have presented Set-the-Scene, a method for generating controllable scenes using a composable NeRF representation.
Our method builds upon two main techniques.
First, a composable NeRF representation where each object in the scene is represented using a dedicated NeRF with its placement controlled using an object proxy.
Second, a Global-Local optimization process.
Building on the flexibility of the composable representation, each object is locally optimized based on its text prompt, while also getting optimized globally with the rest of the scene.
This allows us to create complete scenes, with objects that are consistent with one another.

Set-the-Scene is shown to be superior compared to existing text-to-3D solutions in terms of control and editing capabilities.
Additionally, our method can easily be applied on top of any existing optimization-based single-object text-to-3D solution, enhancing them with several forms of control over the object proxies, as well as easy-to-use editing techniques that require little to no fine-tuning.
We believe Set-the-Scene to be a useful step forward towards a more controllable text-to-3D future.

{\small
\bibliographystyle{ieee_fullname}
\bibliography{egbib}
}

\end{document}